\pdfoutput=1
\documentclass[11pt]{article}

\usepackage[final]{acl}

\usepackage{times}
\usepackage{latexsym}
\usepackage{multirow}

\usepackage[T1]{fontenc}
\usepackage[utf8]{inputenc}

\usepackage{microtype}

\usepackage{inconsolata}

\usepackage{graphicx}
\usepackage{comment}
\usepackage{wrapfig}
\usepackage{booktabs}
\usepackage{tabularx}
\usepackage{pifont} % For tick and cross marks
\usepackage{xcolor}   % For coloring table rows
\usepackage{colortbl}
\usepackage{graphicx} % Required for resizebox

\title{Evaluating Apple Intelligence’s Writing Tools for Privacy Against Large Language Model-Based Inference Attacks: Insights from Early Datasets}
\author{  Mohd. Farhan Israk Soumik, Syed Mhamudul Hasan, and  Abdur R. Shahid\\
  School of Computing\\
  Southern Illinois University\\
  Carbondale, IL, 62901 \\
  \texttt{mohdfarhanisrak.soumik@siu.edu, syedmhamudul.hasan@siu.edu, shahid@cs.siu.edu} 
  }

\begin{document}
\maketitle
\begin{abstract}

%ACL Industry Track

The misuse of Large Language Models (LLMs) to infer emotions from text for malicious purposes, known as emotion inference attacks, poses a significant threat to user privacy. In this paper, we investigate the potential of Apple Intelligence's writing tools, integrated across iPhone, iPad, and MacBook, to mitigate these risks through text modifications such as rewriting and tone adjustment. By developing early novel datasets specifically for this purpose, we empirically assess how different text modifications influence LLM-based detection. This capability suggests strong potential for Apple Intelligence's writing tools as privacy-preserving mechanisms. Our findings lay the groundwork for future adaptive rewriting systems capable of dynamically neutralizing sensitive emotional content to enhance user privacy. To the best of our knowledge, this research provides the first empirical analysis of Apple Intelligence's text-modification tools within a privacy-preservation context with the broader goal of developing on-device, user-centric privacy-preserving mechanisms to protect against LLMs-based advanced inference attacks on deployed systems.

\end{abstract}

\section{Introduction}\label{sec:intro}

In October 2024, Apple Inc. introduced \textit{Apple Intelligence}\footnote{https://www.apple.com/apple-intelligence/}, its generative artificial intelligence (GAI) system~\cite{gunter2024apple}, marking a significant step in the real-world deployment of LLMs for everybody users. In a widely circulated ad\footnote{https://youtu.be/deNzYrTvqCs?feature=shared}, Apple Inc. showcased Apple Intelligence's tone adjustment capabilities, where a user, initially  drafting an emotionally charged message about a stolen pudding, applies the ``friendly'' tone modification feature. The rewritten message transforms the interaction which ultimately helps in mitigating potential conflict, and improving communication~\cite{shu2024rewritelm}. This demonstration and our exploration of Apple Intelligence across various Apple products since its launch have sparked novel research question: \textit{Can we utilize Apple Intelligence's writing tools as a Privacy-Ehancing Technology?} 

To address this question, we investigate its potential as a tool for \textit{emotional privacy}, the ability to conceal or regulate emotional expression in digital communication. Specifically, we examine whether system-wide integration of such tone-adjustment features can enhance privacy, reduce unintended emotional leakage through adversarial detection~\cite{kqiku2022sentiment}, and encourage broader adoption of privacy settings. Our study specifically explores Apple Intelligence in the context of a broader challenge of balancing user control of emotional expression with the increasing use of AI-powered language technologies in daily communication.

We examine adversarial threats exploiting AI models to infer emotional states from text after it leaves a user's device (e.g., MacBook, iPhone). Specifically, we consider adversaries using LLM-based sentiment analysis~\cite{liu2024emollms, zhang2023sentiment, sun2023sentiment, yuan2025improving, zhao-etal-2025-simple, fan-etal-2025-aspect} and fine-tuning on emotionally labeled data~\cite{bucher2024fine, mao2022biases} to reveal latent emotions. Our objective is to investigate whether Apple's on-device AI can offer built-in, system-wide privacy protections against these adversaries. Our methodology contributes to this goal as follows. \textbf{(1) Apple Intelligence as a Privacy-Enhancing AI system:} To our knowledge, this is the first study to examine Apple Intelligence as a deployed AI system that protects user privacy by regulating emotional expressions in text. We analyze its capability to reduce unintended emotional leakage and discuss implications for emotional privacy. \textbf{(2) Early Datasets on Apple Intelligence:} Due to its recent introduction, no public dataset currently exists for evaluating the system’s tone transformation capabilities. We address this gap by introducing two early datasets comprising texts generated by Apple Intelligence’s tone adjustment feature across different tonal options. These datasets establish a foundation for future research on AI-driven emotional privacy and practical NLP applications. \textbf{(3) Evaluation of On-Device Emotion Privacy Protection:} We assess the effectiveness of Apple Intelligence’s writing tools in preserving emotional privacy through tone alteration. Specifically, we evaluate their ability to protect against malicious emotional inference using various LLMs with fine-tuning and prompt engineering techniques.

\section{Related Work}\label{sec:related}
\textbf{``Emotional'' LLMs:} Understanding, analyzing, and replicating human emotions through AI is a crucial field of research on~\cite{wang2022systematic}. Unfortunately, these technologies serve as a double-edge sword with significant privacy concerns, as seen in emotion AI deployment in workplace~\cite{roemmich2023emotion, boyd2023automated} and social networks~\cite{kqiku2022sentiment, rodriguez2023review}. 
%Unfortunately, these technologies serve as a double-edge sword with significant privacy concerns. The increasing reliance on AI for emotional inference simply exacerbate these challenges, as sensitive emotional data becomes more vulnerable to misuse and ethical dilemmas. For example, the deployment of Emotion AI in workplace environments has introduced ethical and privacy challenges, potentially leading to unintended consequences for employees~\cite{roemmich2023emotion, boyd2023automated}. Additionally, social media text is highly susceptible to emotional inference, making it vulnerable to privacy risks associated with sentiment analysis~\cite{kqiku2022sentiment, rodriguez2023review}. 
With recent advancements in LLMs, the challenges surrounding emotional AI have become even more complex. On one hand, LLMs has revolutionized sentiment and emotion analysis, enabling enhanced and accurate sentiment classification in different domains~\cite{luca2024you}. These models have shown that they can pick up on both semantic and syntactic contextual relationships~\cite{miah2024multimodal, hung2024novelty}. For instance, Devlin et al.~\cite{devlin2019bert} showed BERT's ability to outperform traditional models in sentiment classification by leveraging its bidirectional context. Similarly, recent studies~\cite{hartmann2023more, chang2024survey} also found that LLM can surpass traditional sentiment classification models in terms of accuracy and contextual understanding. Additionally, Liu et al.~\cite{liu2024emollms} highlighted that fine-tuning LLMs on specific datasets significantly enhances their performance in detecting subtle emotional cues. Furthermore, Mao et al.~\cite{mao2022biases} suggested prompt-based sentiment analysis and emotion detection using pre-trained LLMs, and Zhang et al.~\cite{zhang2024refashioning} further demonstrated that it can be enhanced by LLMs' capabilities like zero-shot and few-shot learning, in-context learning (ICL) in different emotion classification tasks. 

\textbf{Privacy Concerns:} However, these advancements in LLMs simultaneously deepen the ethical and privacy concerns regarding the users~\cite{das2025security}. LLMs can be misused to extract emotional information from text for malicious purposes, including emotional manipulation~\cite{chen2025emotions}, targeted exploitation~\cite{mozes2023use}, or misinformation campaigns~\cite{liu2024preventing, ma-etal-2024-simulated}. Furthermore, LLMs inherently process vast amount of data, including text embedded with emotional markers. This raises significant privacy risks, as such sensitive data can be exposed through various means, including data breaches~\cite{song2024securesql}, unintended memorization~\cite{bommasani2021opportunities}. Consequently, there has been a growing body of research and practical efforts dedicated to privacy-enhancing technologies for text. These technologies can be categorized into approaches such as data safeguarding, trusted methods, and verification methods~\cite{10.1007/s10462-022-10204-6}. Techniques including de-identification~\cite{mortadi2025intelligent}, anonymization~\cite{pissarra-etal-2024-unlocking}, differential privacy~\cite{meisenbacher-etal-2024-comparative}, and federated learning~\cite{zampieri-etal-2024-federated} have been extensively investigated.

\section{Methodology}\label{sec:method}
Apple Intelligence primarily provides four on device LLM based text formatting approaches for modifying texts: Rewrite, Friendly, Professional and Concise. Each of these features has its own unique set of attributes for modifying a text. For instance, Rewrite aims to provide the text with a clearer and more structured perspective, while Friendly seeks to make it warm and engaging. Formal adds formality and precision to the text, and the goal of the concise feature is to make it direct and clear without compromising its meaning. The study primarily focuses on finding the effectiveness of Apple Intelligence's on device LLM based text formatting features for protecting emotional privacy of the user's. The experimental methodology is organized in sequential steps as described below.
% \begin{enumerate}
%     \item Dataset Selection
%     \item Inference  Model Selection
%     \item Early Dataset Creation for Evaluation.
% \end{enumerate}
\subsection{Experimental Setup}
\subsubsection{Device Information}
To evaluate the privacy-preserving capabilities of Apple Intelligence writing tools and compare them against LLM-based inference attacks, we conducted experiments using a MacBook Pro (Apple Silicon M2, 2023) equipped with a 10-core Apple M2 CPU, 32GB of unified RAM, and running macOS Sequoia. Additional experiments were performed on a Windows workstation featuring an Intel Core i7-14700 CPU (4.65GHz), 80GB DDR5 RAM, an NVIDIA RTX 4500 Ada GPU, and 1TB of storage, running Windows 11. We also utilized Google Colab Pro with 80GB of RAM and an NVIDIA A100 GPU to support further evaluations on the dataset generated using Apple Intelligence.

\subsubsection{Data Collection Procedure}
As Apple Intelligence till this day does not provide any API access for using it's text enhancement features, automated text modification in programmable setting is not possible. That's why all the instances of evaluation texts has to be generated manually from Apple Devices and then stored for future. The detail procedures for dataset selection and evaluation dataset creation is discussed in \ref{subsec:dataset} and \ref{subsec:early evaluation} respectively.
\subsection{Dataset Selection}\label{subsec:dataset}
%To simulate robust emotion inference attacks on text data and observe the efficacy of Apple Intelligence's different text modifying features, selection of training dataset for attacker;s model is highly important because the quality, diversity and balance of the selected dataset has profound impact on effectiveness of the inference attack models. Besides that, texts possesses unique patterns and characteristics of their own, for example texts from social media interactions like tweets happen to be short, static and their context is hard to infer without prior understanding. On the contrary, interpersonal conversations happen to be longer,dynamic often with multiple turns and their context is often easier to understand. 

In order to achieve a generalized picture of Apple AI's privacy preserving capabilities, we proceeded to work on two different datasets: Dair-AI Emotion dataset~\cite{saravia-etal-2018-carer} and DailyDialog dataset~\cite{li-etal-2017-dailydialog}. Dair-AI Emotion dataset~\cite{saravia-etal-2018-carer} consists of six emotion categories-anger, sadness, love, joy, fear and surprise while DailyDialog~\cite{li-etal-2017-dailydialog} comprises of seven emotional categories-neutral,anger,disgust,happiness,sadness,surprise and fear.
\subsection{Inference Model Selection}\label{subsec:inference select}
For the rigorous evaluation of Apple AI to provide emotional privacy, the proper selection of robust inference models for attackers is of great importance because our threat model anticipates an attacker with access to state of the art LLMs. But the dilemma arises on selecting LLM models because of the presence of distinctly different types of LLMs like AutoRegressive Models(Decoder-only LLMs) which includes GPT-4 and GPT-3, Masked Language Models(Encoder only LLMs) which practically covers all the models with BERT based architecture, Sequence-to-Sequence Models(Encoder Decoder Models) which includes Google's Flan T5 and Meta's BART etc. Each of them has unique functionality, scopes, and capabilities for example  Decoder only models struggle with token level classification but excels in generative settings~\cite{Radford2018ImprovingLU, Radford2019LanguageMA}. In contrast, Bidirectional encoder representations from transformer-based encoder-only models (BERT) have always demonstrated superior performance in classic classification tasks against GPT-based decoder-only models ~\cite{liu2019roberta, Radford2019LanguageMA, qiu2020pre} and Sequence-to-Sequence models~\cite{raffel2020exploring}. For simulating effective emotion inference attack, we have used models all three distinct categories. BERT, RoBERTa, DistilBERT, DeBERTa from Encoder Only LLMs Category, Flan T5 from Sequence-to-Sequence models and GPT-4o and DeepSeekR1 has been selected as attacker's inference models. Due to it's superior performance in classification tasks which has been demonstrated by multiple research works~\cite{liu2019roberta, Radford2018ImprovingLU, qiu2020pre, raffel2020exploring}, models based on BERT architectures have been given additional focus. 
\subsection{Early Dataset Creation for Evaluation}\label{subsec:early evaluation}
After the attack models are selected and trained on the selected datasets, the evaluation part of the experiment begins by focusing on observing how the selected inference model performs on texts after being modified by writing tools of Apple Intelligence. As the experiment is performed on two different datasets with their corresponding LLM based inference models, two separate evaluation dataset is required. For Dair-AI Emotion dataset~\cite{saravia-etal-2018-carer}, 40 instances of texts has been selected from each emotional category from train,test and validation set randomly. Finally, each of these selected instances has been modified through Apple AI's writing tools Rewrite, Friendly, Professional and Concise respectively and saved along with original non-modified text. This newly created evaluation dataset will be called Evaluation Dataset 1 throughout this experiment. 
In the case of DailyDialog dataset~\cite{li-etal-2017-dailydialog}, as the texts are from formal or informal conversations between two or more persons, an important thing which has to be considered is the number of words in a sentence. As it has been observed that if the text that we are dealing with is like "Good, got it" or "Yes, I will" , modified text from Apple AI's writing tools have is little to no changes from the original sentence. That's why for this dataset just like the previous dataset, 40 instances of text has been selected from each emotion category using similar procedure except one filtering condition that for getting selected that text should contain at least more than five words. After filtered out texts were selected and modified using Apple AI's writing tools, they have been saved in similar manner along with unmodified text. Just like the first one, this dataset will called Evaluation Dataset 2.

\section{Results and Analysis}\label{sec:results}

\begin{figure}
    \centering
\includegraphics[width=.95\linewidth]{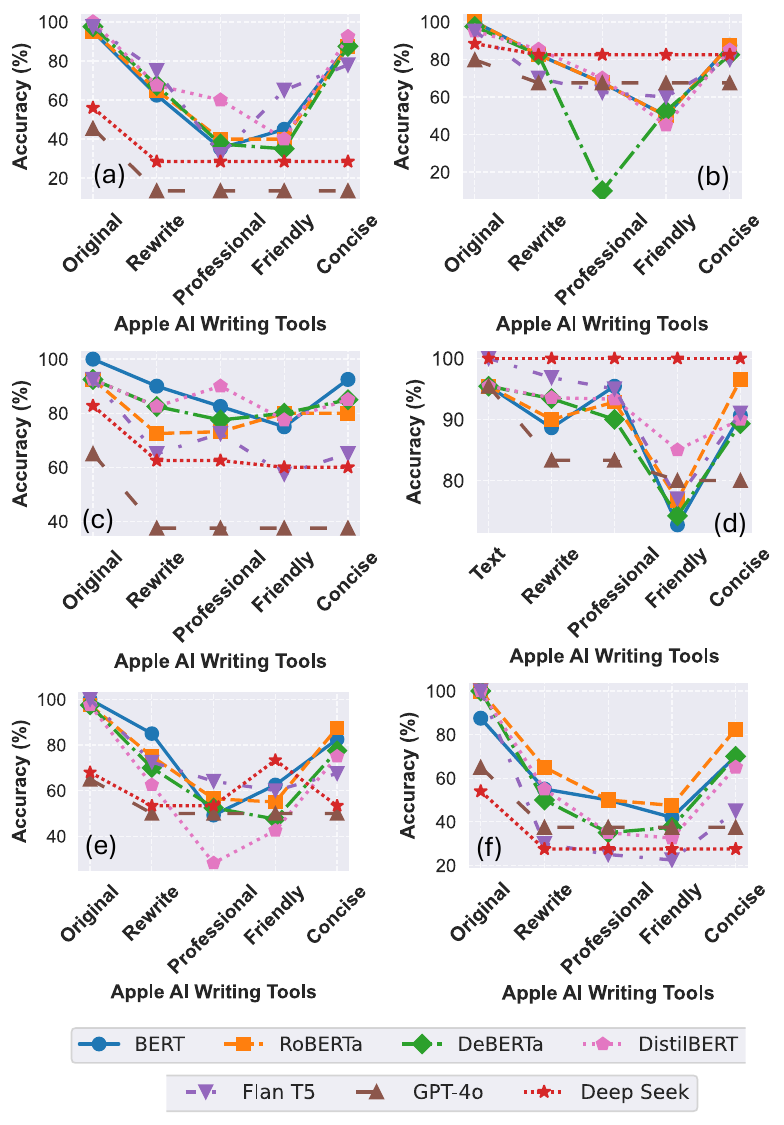}
    \caption{Accuracy on Apple Intelligence dataset based on dair (throughout the paper we call this Evaluation 1 dataset): (a) anger, (b) fear, (c) joy (d) surprise  (e) sadness  (f) love.}
    \label{fig:data-1-accuracy}
    \vspace{-15pt}
\end{figure}

\begin{figure}
    \centering
\includegraphics[width=1\linewidth]{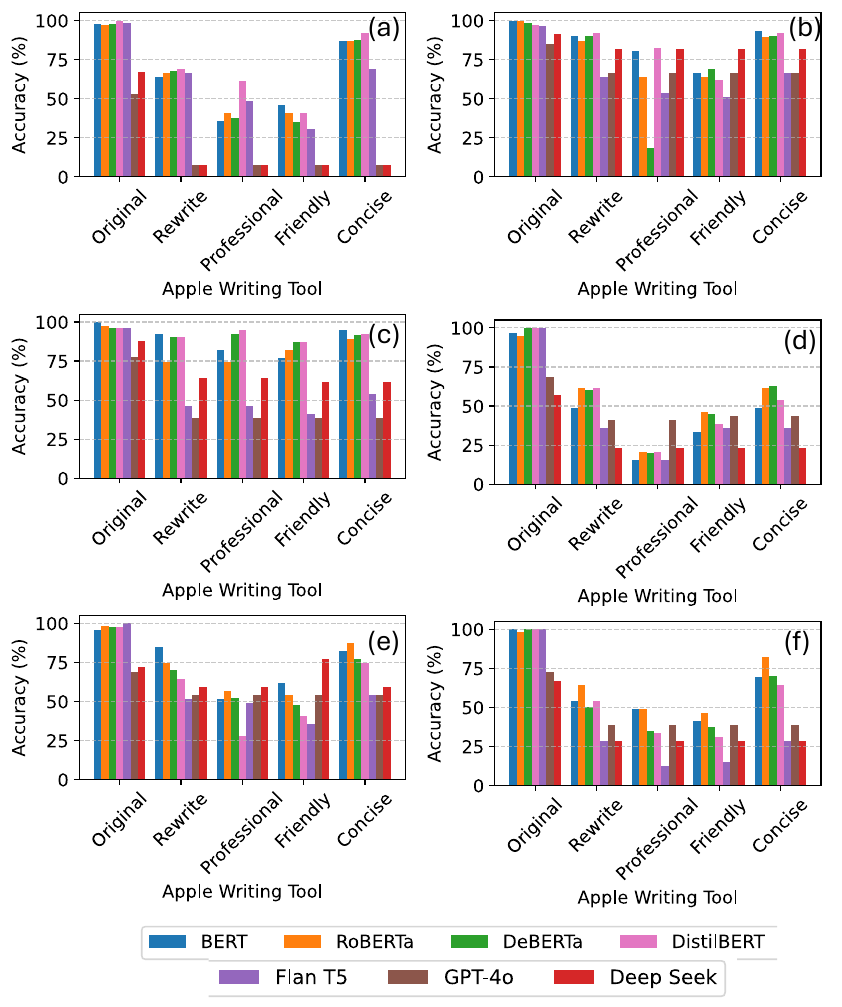}
    \caption{F1 Score on Apple Intelligence dataset based on dair (throughout the paper we call this Evaluation 1 dataset): (a) anger, (b) fear, (c) joy (d) surprise  (e) sadness  (f)love}
    \label{fig:data-1-f1} \vspace{-15pt}
\end{figure}

\begin{figure*}[h!] 
    \centering
\includegraphics[width=.95\linewidth]{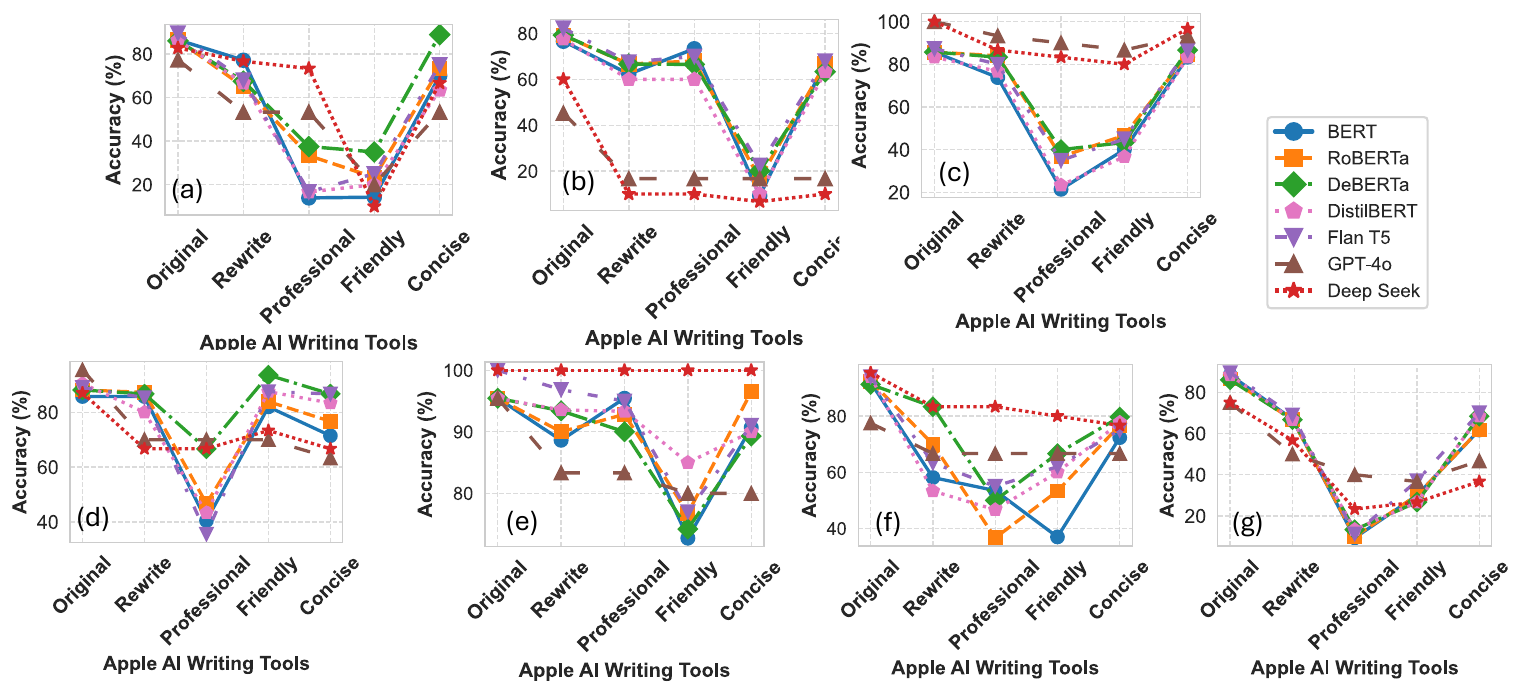}
    \caption{Accuracy on Apple Intelligence Evaluation dataset based on Daily Dialog Dataset (throughout the paper we call this Evaluation Dataset 2): a)anger b)disgust c)fear d)happiness e)neutral f)sadness g)surprise}
    \label{fig:data-2-accuracy}%\vspace{-15pt}
\end{figure*}

\begin{figure*} 
    \centering
\includegraphics[width=1\linewidth]{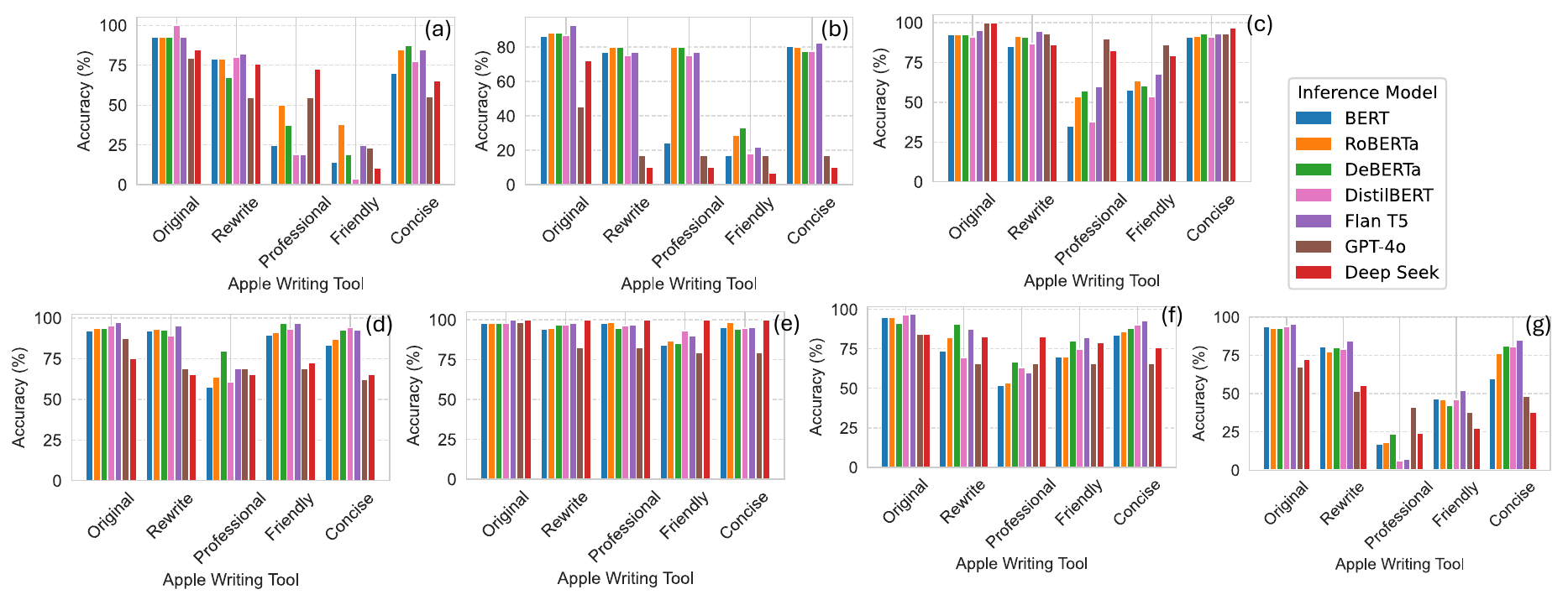} 
    \caption{F1 Score on Apple Intelligence Evaluation dataset based on Daily Dialog Dataset (throughout the paper we call this Evaluation Dataset 2): a) anger b)disgust c)fear d)happiness e)neutral f)sadness g)surprise}
    \label{fig:data-2-f1} \vspace{-15pt}
\end{figure*}

After finalizing the attacker's model and early evaluation dataset, the experiment moves on to evaluation section where results were generated and analyzed to observe how attacker's emotion inference models perform. The goal is to find out whether texts enhanced by Apple AI's writing tools can fail the attacker's inference models or not and . For measuring performance, classification accuracy and the $F_1$ score have been used as evaluation metrics. As both evaluation datasets have overlapping emotions, performances on similar emotions are analyzed together in this study.
Fig.\ref{fig:data-1-accuracy} and Fig.\ref{fig:data-2-accuracy} show the comparison accuracies of inference models for different categories of emotion on Evaluaton Dataset 1 and Dataset 2 respectively. While Fig.\ref{fig:data-1-f1} and Fig.\ref{fig:data-2-f1} provide a comparative view of $F_1$ scores of inference models on both datasets. From these figure, it is clearly evident that performances of inference models have been been affected by Apple AI's writing tools. But the level of impact these tools and their corresponding scope are pretty distinct. 

Fig \ref{fig:data-1-accuracy}(a) and Fig \ref{fig:data-2-accuracy}(a) provide a graphical representation of performances of inference models on texts having emotion anger on both datasets respectively. Attacker's finetuned LLM based inference models performed impressively well predicting emotions of unmodified texts. But when those same texts are modified by Apple Professional and Friendly features, models' performance degrades drastically, for example DistilBERT which predicted every unmodified texts correctly cannot predict more than 60\% and 40\% texts accurately after being modified by Professional and Friendly tools repsectively for eval dataset 1. For evaluation dataset 2, the same model's accuracy drops to 16.67\% and 20.00\% respectively after enhancement. From Fig.\ref{fig:data-1-f1}(a) and Fig.\ref{fig:data-2-f1}, it can be them inferences model's $F_1$ scores get severe degradation for Apple Friendly compared to Apple Professional, indicating severe failure of inference attacks. Modification by Apple Rewrite causes moderate performance degradation but Apple Concise has minimal effects on attack models in both cases.
  Fig.~\ref{fig:data-1-accuracy}(d) provides a graphical representation of attacker's inference models' performance on disgust emotions. Just as anger, Apple Rewrite and Concise tools's ability to influence inference models is quite limited. Here, Professional tool causes the maximum degradation of inference model's performances which as Fig.\ref{fig:data-2-f1}(b) show, in some cases(e.g DeBERTa) reduces model's accuracy by a margin of 70\% . Similar trends can be observed for emotion category Sadness as represented by Fig \ref{fig:data-1-accuracy}(e) and Fig \ref{fig:data-2-accuracy}(f). After modifications of the texts by Apple Professional and Apple Friendly, fine-tuned inference models that performed robustly on the original texts (e.g BERT and Flan T5 having 100\% accuracy) misclassify more than 40\% of the modified texts for Evaluation Dataset 1. Though the inference models perform comparatively better on Evaluation Dataset 2, it still fails to classify 45\%-50\% on average. One noticeable observation is that although the impact of the Concise tools remain quite the same but Apple Rewrite's impact on failing the model severely degrades. For the emotional category of Love, an interesting development is observed regarding the performance of Apple Rewrite tool as can be seen from Fig.\ref{fig:data-1-accuracy}(f). Where in previous two cases, Rewrite tool had shown moderate success in failing inference models, here along with Professional and Friendly tools, attacker's inference models' performances received severe degradation after texts getting enhanced by Apple Rewrite, for example Flan T5 which had 100\% success rates against original text, got a near 70\% performance degradation. This degradation is pretty close to lowest degradation of nearly 80\% that it faced due to Apple Friendly. Apple concise as in the previous cases showed moderate success against inference models. 

Attacker's inference models irrespective of finetuned or prompt engineered demonstrates robust performance against modification done by writing tools compared to aforementioned categories as can be seen from Fig \ref{fig:data-1-accuracy}(c). The impact that Apple Rewrite had on original Love is neutralized in the case of Joy while Friendly and Professional as always having the highest degradation effect(25\% and 37.50\% misclassification against BERT and Flan T5 respectively). Though evaluation dataset 2 doesn't have the emotional category of Joy, it containss Happiness which is pretty closer to Joy. Fig.\ref{fig:data-2-accuracy}(d) shows that inference models performed though slightly better with regards to evaluation dataset 2  compared to that of evaluation dataset 1, they both demonstrate a similar pattern. Like evaluation set 1, Apple Professional caused significant performance degradation, for example Flan T5's misclassifcation rate increased from just above 10\% to more than 85\% after being enhanced by Professional tool. But the performance difference between Joy and Happiness is observed regarding Friendly tool where it proved to more successful in misleading inference models in Joy compared to Happiness. Apple Rewrite and Concise demonstrates marginal impact over LLM based inference attacks with regards to emotion category of Fear. Apple Rewrite reduces inference models's accuracy by around 20\%  rates while Friendly tools again having the highest impact (e.g. 55\% missclassifcation of DistilBERT on evaluation set1 and 75\% misclassification in evaluation set2) can be observed by Fig.~\ref{fig:data-1-accuracy}(b) and Fig.~\ref{fig:data-2-accuracy}(c). Just like in the previous case, Apple Rewrite and Concise can marginally deter attacker's inference models from extracting emotions correctly.

With regards to  emotion surprise, all fine-tuned LLM inference models have strong performance on original unmodified texts with a success rate 100\% in correctly predicting emotion on both evaluation sets as can be observed from Fig.~\ref{fig:data-1-accuracy}(d) and Fig.~\ref{fig:data-2-accuracy}(g) respectively. But as again, Professional and Friendly tools of Apple AI impedes inference models' objectives by severly degrading their performance, for example finetuned RoBERTa initially achieved 86.45\% accuracy against unmodified text of evaluation set 1 but after they  got enhanced by Friendly tool, it's failure rate increased towards around 90\%. Similar development can be observed for inference models in evaluation set 2 where after modification by Friendly tool, finetuned Flan T5's accuracy droped from 96\% to mere 6.04\%. 
Attacker's inference models overall perform significantly well on Neutral emotion category. Inference models performances on unmodified original texts and texts after being modified by Rewrite and Professional tools remain quitre identical as shown in Fig.\ref{fig:data-2-accuracy}(e). Although, Professional and Friendly tools impede attacker's from inferring emotions, their performance are not as significant compared to other emotion from same dataset like disgust and surprise. It's also one of the case where along with finetuned models, prompt engineered models also did excessively well.

\section{Actionable Insights for Privacy-Preserving AI Writing Tools}

Our findings demonstrate that Apple Intelligence's on-device text modification tools significantly alter the emotional level of text, thereby affecting the ability of LLMs to accurately infer emotions. This modification serves as a foundation for an effective privacy-preserving mechanism, reducing unintentional privacy leakage in communication with LLM-based chatbots. These results have significant implications for the system-wide integration of privacy-aware rewriting features in NLP systems. 

Specifically, our findings highlight the feasibility of developing on-device privacy-aware rewriting mechanisms within the ecosystem that dynamically adjust text while preserving usability. With the system-wide integration of these capabilities (in the case of Apple Inc.'s iOS, iPadOS, and macOS), it is possible to offer customizable modification levels that balance user intention with privacy needs. For example, adaptive privacy filters could dynamically modify emotional content based on user-specific privacy levels (e.g., activating ``professional'' mode in Apple Mail, Messages, or Notes when privacy is a high concern). Moreover, this research suggests that AI assistants (e.g., Siri, ChatGPT, Gemini, and CoPilot) can integrate user-configurable privacy controls that allow users to adjust how much emotional information (privacy leakage) remains in modified text. By embedding these privacy-preserving features at the system level, Apple, in particular, can further enhance its on-device AI strategy to strengthen its privacy-first policy for AI.

\section{Limitations and Future Works}
Despite demonstrating promising prospects regarding the privacy preserving capability of Apple intelligence writing tools, our study faces some limitations, which are discussed below and forms the foundation of our future work. The two different datasets (Dair AI Emotion and DailyDialog datasets) used in this experiment are comprised of texts having lengths ranging from 10 to 50 words which may not fully capture the complexities associated with processing longer form of texts. As in the case of longer texts, contextual dependencies and narrative coherence play a vital role. As a result, experiments are restricted to short texts, omitting longer contexts and narrative dynamics that are important for real-world emotional expression and privacy concerns. Another limitation lies in the size of early early datasets created and used for evaluation. Due to the absence of usable API's for modifying texts automatically as mentioned previously, every instance of both of the early evaluation datasets had to be created manually. As a result of this, the early datasets (Evaluation Dataset 1 and Evaluation Dataset 2)  are relatively smaller (only 40 instances per emotion category) in volume. The limited size of the evaluation datasets, though sufficient for preliminary analysis, has the potential of raising concerns about the robustness and generalizability of the results in a broader perspective. Apart from that, this study evaluates Apple Intelligence in isolation without comparing it to other paraphrasing or rewriting models, making it difficult to assess its relative effectiveness

Building upon the current findings and addressing limitations faced during this study, in our future works we are poised to work on several directions as follows:
\begin{itemize}
    \item \textbf{Incorporation of Long-Form Texts}: In future work, we aim to incorporate long-form text datasets to better capture complex contextual dependencies and assess how Apple Intelligence tools preserve emotion and privacy in extended real-world contexts.
    \item \textbf{Expansion of Evaluation Datasets}: Future experiments will focus on scaling evaluation datasets by increasing instances for each emotion category, to capture a wider range of emotional expressions and edge cases, enhancing statistical reliability and real-world applicability.
    \item \textbf{Cross-Domain Dataset Integration}:We plan to evaluate the system across diverse text domains—emails, blogs, and social media—to assess its performance in varied linguistic and emotional contexts.
    \item \textbf{Model Benchmarking}:Future work will include comparative analysis with existing paraphrasing and rewriting models to better assess the relative effectiveness of Apple Intelligence in preserving emotion and privacy.
\end{itemize}

\section{Conclusion}
This study presents one of the first empirical investigations into Apple Intelligence’s writing tools as potential privacy-enhancing mechanisms against LLM-based emotion inference attacks. By systematically evaluating the impact of Apple’s tone-modification features—Rewrite, Friendly, Professional, and Concise—on text samples drawn from emotion-labeled datasets, we demonstrate that specific tools, particularly Friendly and Professional, can significantly reduce the accuracy of adversarial models attempting to infer emotional content. Our results highlight a promising direction for integrating emotional privacy features into on-device AI systems. These findings support the feasibility of leveraging text rewriting tools not only for stylistic enhancements but also to mitigate unintended emotional leakage in digital communications. As emotional inference becomes more pervasive in LLM-based services, incorporating privacy-aware rewriting capabilities into everyday writing interfaces may prove essential for maintaining user agency and data confidentiality. This work lays the groundwork for future research into adaptive, user-configurable privacy-preserving mechanisms embedded directly within personal AI ecosystems.

\section{Ethical Concerns}\label{sec:limitations}
\begin{comment}
    
Limitations:
Dataset

Text small ... 
Long story not tested.

Tweet... 

https://arxiv.org/pdf/2406.06485
\end{comment}

%{\color{blue}

There are several ethical considerations associated with using LLM for emotion analysis. While LLMs can identify emotions, they are susceptible to misclassification, particularly when dealing with sensitive or ambiguous content. Additionally, the training data used to develop these models is collected and annotated by humans, which may cause them to miss certain nuances of human emotion. Finally, this experiment was conducted without explicit permission from Apple; therefore, the findings and observations presented are intended solely for educational/research purposes only and do not reflect any Apple's endorsement or affiliation.

\bibliography{anthology,custom}
%\bibliography{custom}

\appendix
\section{Appendix}

\subsection{Apple Intelligence}\label{sec:apple}

\begin{comment}
    
https://machinelearning.apple.com/research/introducing-apple-foundation-models
https://aclanthology.org/2024.acl-short.1.pdf

\end{comment}

\begin{figure}[ht!]
    \centering
    \includegraphics[width=0.95\linewidth]{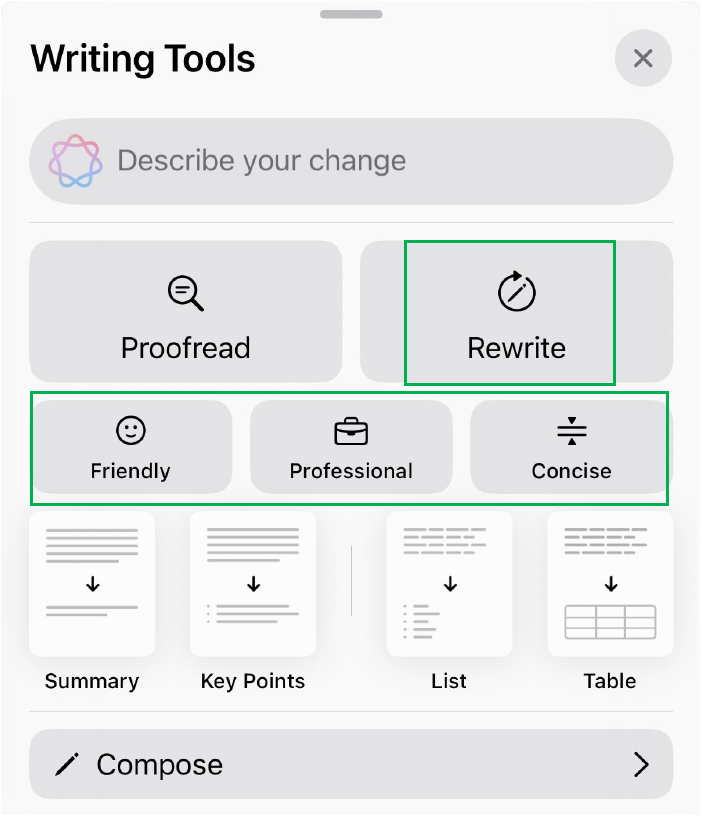}
    \caption{The Apple Intelligence's writing tools. In this research, we analyze the rewrite, friendly, professional, and concise tone options (marked as boxed).}
    \label{fig:apple-intelligence}
\end{figure}

Apple Intelligence is an advanced GAI system designed to enhance user interactions through different foundation models. The core of Apple intelligence integrates both on-device and server-based language models to optimize performance while maintaining user privacy. The model on the device is made up of about three billion parameters that make sure that basic tasks are processed quickly, whereas larger server-based models that run in Apple's Private Cloud Compute infrastructure is responsible for more complicated calculations. One of the key features of Apple Intelligence is the use of responsible AI development, which prioritizes data privacy through maintaining a strict policy against utilizing users' personal data to train its foundation models for all the Apple devices.

\subsection{Models}\label{sec:models}

\subsubsection{BERT}
BERT is a deep learning model based on the transformer architecture. It is developed to learn contextual representations bidirectionally using multi-head self attention along with feedforward layers.

\begin{table}[h!]
    \centering %\small
    \begin{tabular}{|p{3cm}|p{4cm}|}
        \hline
        \textbf{Configuration} & \textbf{Value} \\
        \hline
        Pretrained Model  & bert-base-uncased \\
        \hline
        Learning Rate     & 2e-4 \\
        \hline
        Dropout Rate      & 0.2 \\
        \hline
        
    \end{tabular}
    \caption{BERT Configuration}
    \label{tab:bert_config}
\end{table}

\subsubsection{RoBERTa}
RoBERTa, developed by Facebook AI is an optimized variant of BERT. Compared to BERT, it has been trained on much larger dataset and it uses masked language modeling instead of next token generation.
\begin{table}[h!]
    \centering
    \begin{tabular}{|p{3cm}|p{4cm}|}
        \hline
        \textbf{Configuration} & \textbf{Value} \\
        \hline
        Pretrained Model  & twitter-roberta-base \\
        \hline
        Learning Rate     & 3e-4 \\
        \hline
        % Train Accuracy    & 98\% \\
        % \hline
        % Test Accuracy     & 93\% \\
        % \hline
    \end{tabular}
    \caption{RoBERTa Configuration}
    \label{tab:roberta_config}
\end{table}
\subsubsection{DeBERTa}
DeBERTa is a transformer-based language model developed by Microsoft that improves upon BERT and RoBERTa by introducing two key innovations: disentangled attention and an enhanced decoding mechanism. Unlike traditional models that combine word content and position embeddings before feeding them into the attention mechanism, DeBERTa keeps them separate, allowing the model to better capture the relationships between words based on both their content and position independently.

\begin{table}[h!]
    \centering
    \begin{tabular}{|p{3cm}|p{4cm}|}
        \hline
        \textbf{Configuration} & \textbf{Value} \\
        \hline
        Pretrained Model  & deberta-base-uncased \\
        \hline
        Learning Rate     & 2e-4 \\
        \hline
        Training Method      & Full Fine Tuning \\
        \hline
    \end{tabular}
    \caption{DeBERTa Configuration}
    \label{tab:deberta_config}
\end{table}

\subsubsection{DistilBERT}
DistilBERT is a lighter version of BERT developed by Hugging Face through knowledge distillation. It was trained to mimic the behavior of the larger BERT model by learning from its outputs, effectively compressing the knowledge without significant performance loss.
\begin{table}[h!]
    \centering
    \begin{tabular}{|p{3cm}|p{4cm}|}
        \hline
        \textbf{Configuration} & \textbf{Value} \\
        \hline
        Pretrained Model  & distilbert-base-uncased \\
        \hline
        Learning Rate     & 3e-4 \\
        \hline
        Training Method     & Full Fine Tuning \\
       \hline
    \end{tabular}
    \caption{DistilBERT Configuration}
    \label{tab:distilbert_config}
\end{table}
\subsubsection{Flan T5}
Flan-T5 is an advanced version of Google’s T5  model. It was fine-tuned using instruction tuning on a variety of tasks to enhance its ability to follow natural language instructions. Flan-T5 significantly improves zero-shot and few-shot learning performance across multiple benchmarks.
\begin{table}[h!]
    \centering
    \begin{tabular}{|p{3cm}|p{4cm}|}
        \hline
        \textbf{Configuration} & \textbf{Value} \\
        \hline
        Pretrained Model  & google/flan-t5-base \\
        \hline
        Learning Rate     & 5e-4 \\
        \hline
        Training Method     & PEFT-LoRa \\
        \hline
        Lora Rank Matrix      & 16 \\
    \hline
    \end{tabular}
    \caption{Flan T5 Configuration}
    \label{tab:flan_t5_config}
\end{table}
\subsection{GPT-4o }
GPT-4o is OpenAI’s latest AI model. It was designed to handle text, image, and audio inputs, offering a multi-modal experience. GPT-4o uses a unified architecture to integrate and understand information across modalities in real time. It maintains the strong language capabilities of GPT-4 while significantly improving performance on vision and audio tasks, such as interpreting images, recognizing emotions in speech, or holding fluid voice conversations. 

\subsubsection{Deep Seek}

DeepSeek is a open-source LLM developed by the Chinese AI company DeepSeek. It features both a base model and a chat-optimized variant, trained on an extensive dataset that includes both English and Chinese, making it effective for multilingual tasks, and other NLP tasks. It was built with transformer architecture and optimized for reasoning, code generation, and general-purpose language understanding.
\subsection{Prompt Template}
\begin{figure}[h!]
    \centering   \includegraphics[width=0.99\linewidth]{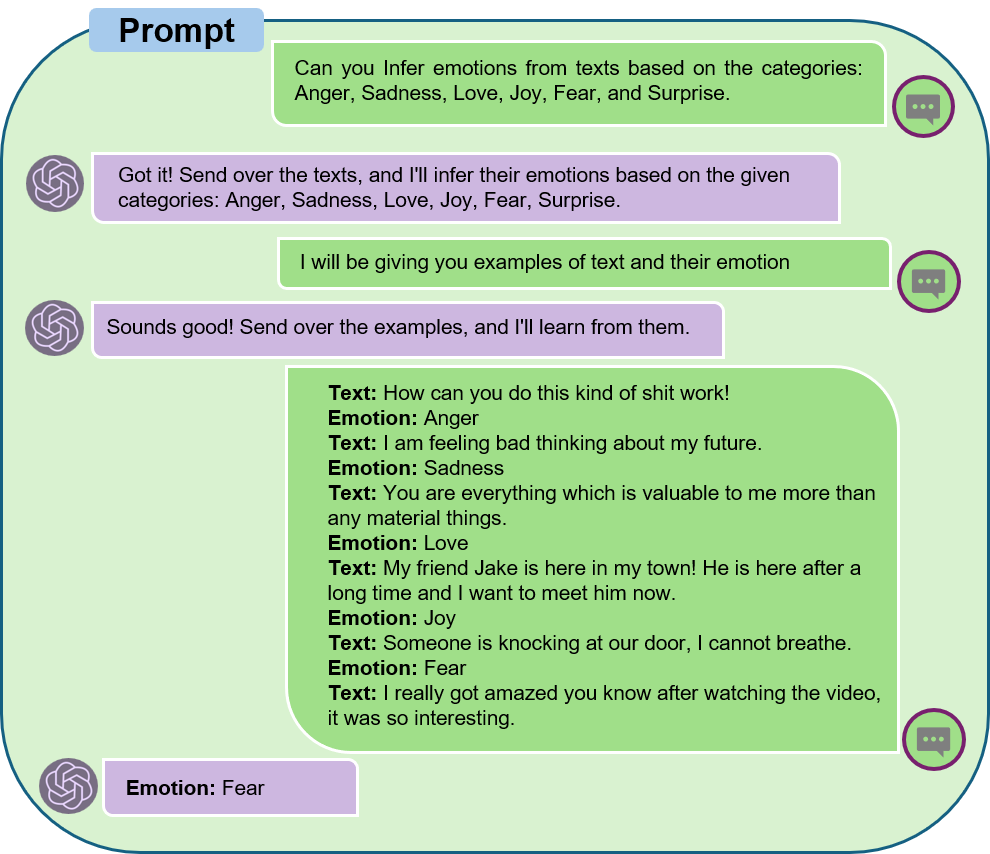}
    \caption{Sample prompt text to predict emotion from LLM. }
    \label{fig:prompt}
\end{figure}

Our sample prompt template is shown in the figure~\ref{fig:prompt}.

\subsection{Datasets}\label{sec:datasets}

We use two datasets to demonstrate the results:
\subsubsection{Dair-AI Emotion Dataset}
The Dair-AI Emotion Dataset is a collection of English Twitter messages labeled with six basic emotions: anger, fear, joy, love, sadness, and surprise. This dataset is designed for emotion recognition research and has been preprocessed for ease of use in NLP pipelines. It contains 20,000 text instances which are divided into training (16,000 instances), validation (2,000 instances), and test (2,000 instances) sets.
\begin{table}[ht]
    \centering \small
   
    \begin{tabular}{|p{3cm}|p{1.7cm}|}
        \hline
        Emotion  & Count \\
        \hline
        Anger    & 4666  \\
        \hline
        Fear     & 5362  \\
        \hline
        Joy      & 1304  \\
        \hline
        Love     & 2159  \\
        \hline
        Sadness  & 1937  \\
        \hline
        Surprise & 572   \\
        \hline
        %\bottomrule
    \end{tabular}
     \caption{Emotion distribution in the Dair-AI Emotion dataset}
    \label{tab:emotion_distribution}
\end{table}

\subsubsection{DailyDialog Dataset}
The DailyDialog dataset is a multi-turn, open-domain English dialog collection. It comprises 13,118 dialogues, reflecting daily communication and covering various topics. The data set is divided into training sets (11,118 dialogues), validation sets (1,000 dialogues), and test sets (1,000 dialogues). On average, each dialogue consists of approximately 8 speaker turns, with around 15 tokens per turn. The conversations are manually crafted, ensuring high-quality and natural language interactions and encompass a wide range of daily life topics, providing a rich resource for open-domain conversation modeling. Unlike conversations, dialogues are also manually labeled with communication intentions and emotion information, facilitating research in dialogue systems, emotion recognition, and natural language understanding.
\begin{table}[ht]
    \centering
    
    \begin{tabular}{|p{4.5cm}|p{2cm}|}
        \hline
        \textbf{Category} & \textbf{Count} \\
        \hline
        Total Dialogues            & 13,118 \\
        \hline
        Training Set               & 11,118 \\
        \hline
        Validation Set             & 1,000  \\
        \hline
        Test Set                   & 1,000  \\
        \hline
        Average Turns per Dialogue & 8      \\
        \hline
        Average Tokens per Turn    & 15     \\
        \hline
        %\bottomrule
    \end{tabular}
    \caption{Distribution of the DailyDialog Dataset}
    \label{tab:dailydialog_distribution}
\end{table}
\begin{comment}
    Early Dataset Section Starts.
    ....................................................
    ....................................................................
    .,,,,,,,,.........................................................................................
    ...........................................................................................................................................................................................................................................................................................................................................................................................................................................................................................................................................................................................................................................................................................................................................................................................................................................................................................................................................................
\end{comment}

\subsection{Detailed Results}
\begin{table*}[ht!]
    \centering\small
    \caption{Emotion Inference Models Accuracy (\%) Comparison on Dair-AI based Evaluation Dataset}
    \begin{tabular}{p{1.5cm}|p{1.5cm}|p{1.5cm}|p{1.5cm}|p{1.5cm}|p{1.5cm}|p{1.5cm}}
        \hline
        \textbf{Emotion Category} & \textbf{Inference Model} & \textbf{Original Text} & \textbf{Apple Rewrite} & \textbf{Apple \break Professional} & \textbf{Apple Friendly} & \textbf{Apple \break Concise} \\
        \hline
        \multirow{7}{*}{Anger}
        & BERT       & 95.12 & 62.50 & 35.32 & 45.00 & 87.50 \\ \cline{2-7}
        & RoBERTa    & 95.20 & 65.00 & 40.00 & 40.00 & 87.50 \\ \cline{2-7}
        & DeBERTa    & 97.56 & 67.50 & 37.50 & 35.00 & 87.50 \\ \cline{2-7}
        & DistilBERT & 100.00 & 67.50 & 60.00 & 40.00 & 92.50 \\ \cline{2-7}
        & Flan T5    & 97.56 & 75.00 & 32.00 & 65.00 & 78.00 \\ \cline{2-7}
        & GPT-4o     & 45.50 & 13.50 & 13.50 & 13.50 & 13.50 \\ \cline{2-7}
        & Deep Seek  & 56.00 & 28.50 & 28.50 & 28.50 & 28.50 \\ 
        \hline \hline

        \multirow{7}{*}{Sadness}
        & BERT       & 100.00 & 85.00 & 49.20 & 62.50 & 82.50 \\ \cline{2-7}
        & RoBERTa    & 97.50  & 75.00 & 56.41 & 55.00 & 87.50 \\ \cline{2-7}
        & DeBERTa    & 97.50  & 70.00 & 52.45 & 47.50 & 77.50 \\ \cline{2-7}
        & DistilBERT & 97.50  & 62.50 & 28.21 & 42.50 & 75.00 \\ \cline{2-7}
        & Flan T5    & 100.00 & 72.50 & 64.10 & 60.00 & 67.50 \\ \cline{2-7}
        & GPT-4o     & 65.00  & 50.00 & 50.00 & 50.00 & 50.00 \\ \cline{2-7}
        & Deep Seek  & 68.00  & 53.33 & 53.33 & 73.33 & 53.33 \\ 
        \hline \hline

        \multirow{7}{*}{Love}
        & BERT       & 87.50  & 55.00 & 50.00 & 42.00 & 70.00 \\ \cline{2-7}
        & RoBERTa    & 100.00 & 65.00 & 50.00 & 47.50 & 82.50 \\ \cline{2-7}
        & DeBERTa    & 100.00 & 50.00 & 35.00 & 37.50 & 70.00 \\ \cline{2-7}
        & DistilBERT & 100.00 & 55.00 & 35.00 & 32.50 & 65.00 \\ \cline{2-7}
        & Flan T5    & 100.00 & 30.00 & 25.00 & 22.50 & 45.00 \\ \cline{2-7}
        & GPT-4o     & 65.00  & 37.50 & 37.50 & 37.50 & 37.50 \\ \cline{2-7}
        & Deep Seek  & 54.00  & 27.50 & 27.50 & 27.50 & 27.50 \\ 
        \hline \hline

        \multirow{7}{*}{Joy}
        & BERT       & 100.00 & 90.00 & 82.50 & 75.00 & 92.50 \\ \cline{2-7}
        & RoBERTa    & 92.50  & 72.50 & 73.17 & 80.00 & 80.00 \\ \cline{2-7}
        & DeBERTa    & 92.50  & 82.50 & 77.50 & 80.00 & 85.00 \\ \cline{2-7}
        & DistilBERT & 92.50  & 82.50 & 90.00 & 77.50 & 85.00 \\ \cline{2-7}
        & Flan T5    & 92.50  & 65.00 & 72.50 & 57.50 & 65.00 \\ \cline{2-7}
        & GPT-4o     & 65.00  & 37.50 & 37.50 & 37.50 & 37.50 \\ \cline{2-7}
        & Deep Seek  & 82.75  & 62.50 & 62.50 & 60.00 & 60.00 \\ 
        \hline \hline

        \multirow{7}{*}{Fear}
        & BERT       & 100.00 & 82.50 & 67.50 & 50.00 & 87.50 \\ \cline{2-7}
        & RoBERTa    & 100.00 & 82.50 & 67.50 & 50.00 & 87.50 \\ \cline{2-7}
        & DeBERTa    & 97.50  & 82.50 & 10.00 & 52.50 & 82.50 \\ \cline{2-7}
        & DistilBERT & 95.12  & 85.37 & 70.00 & 45.00 & 85.00 \\ \cline{2-7}
        & Flan T5    & 95.12  & 70.00 & 62.50 & 60.00 & 80.00 \\ \cline{2-7}
        & GPT-4o     & 79.79  & 67.50 & 67.50 & 67.50 & 67.50 \\ \cline{2-7}
        & Deep Seek  & 88.40  & 82.50 & 82.50 & 82.50 & 82.50 \\ 
        \hline \hline

        \multirow{7}{*}{Surprise}
        & BERT       & 100.00 & 47.50 & 15.00 & 32.50 & 50.00 \\ \cline{2-7}
        & RoBERTa    & 100.00 & 60.00 & 20.00 & 45.00 & 62.50 \\ \cline{2-7}
        & DeBERTa    & 100.00 & 60.00 & 20.00 & 45.00 & 62.50 \\ \cline{2-7}
        & DistilBERT & 100.00 & 60.00 & 37.50 & 10.00 & 55.00 \\ \cline{2-7}
        & Flan T5    & 100.00 & 42.50 & 37.50 & 50.00 & 50.00 \\ \cline{2-7}
        & GPT-4o     & 67.00  & 42.50 & 42.50 & 45.00 & 45.00 \\ \cline{2-7}
        & Deep Seek  & 56.00  & 22.50 & 22.50 & 22.50 & 22.50 \\ 
        \hline  
    \end{tabular}
    
    \label{tab:emotion_interference}
\end{table*}

\begin{table*}[ht!]
    \centering
    \caption{Emotion Inference Models'Accuracy (\%) Comparison on DailyDialog based Evaluation Dataset} \small
    \begin{tabular}{p{1.5cm}|p{1.5cm}|p{1.5cm}|p{1.5cm}|p{1.5cm}|p{1.5cm}|p{1.5cm}}
        \hline
        \textbf{Emotion Category} & \textbf{Inference Model} & \textbf{Original} & \textbf{Apple Rewrite} & \textbf{Apple \break Professional} & \textbf{Apple Friendly} & \textbf{Apple \break Concise} \\
        \hline
        \multirow{7}{*}{Anger} & BERT & 86.50 & 77.14 & 13.9 & 14.23 & 70.00 \\
        \cline{2-7}
         & RoBERTa & 86.40 & 65.00 & 33.33 & 23.33 & 73.33 \\
        \cline{2-7}
         & DeBERTa & 86.04 & 67.50 & 37.50 & 35.00 & 88.89 \\
        \cline{2-7}
         & DistilBERT & 86.23 & 66.67 & 16.67 & 20.00 &  63.33\\
        \cline{2-7}
         & Flan T5 & 89.56 & 68.00 &  16.67 & 25.00 & 75.00\\
        \cline{2-7}

                  & GPT-4o & 77.34 & 53.33 &53.33 &20.00 &53.33
 \\
       \cline{2-7}
                  & Deep Seek & 82.77 & 76.67 &73.33 & 10.00 &66.67
 \\
        \hline
        \hline
        \multirow{7}{*}{Disgust} & BERT & 76.63 & 62.30 &73.33 & 9.26 & 67.30 \\
        \cline{2-7}
          & RoBERTa & 79.10 & 66.67 & 68.00 & 16.67 & 66.67 \\
        \cline{2-7}
         & DeBERTa & 79.30 & 66.67 & 66.53 & 20.00 & 63.33 \\
        \cline{2-7}
         & DistilBERT & 77.64 & 60.00 & 60.00 & 10.00 & 63.33 \\
        \cline{2-7}
         & Flan T5 & 82.30 & 67.50 & 70.00 & 22.50 & 68.00 \\

        \cline{2-7}
         & GPT-4o & 45.30 & 16.67 &16.67 &16.67 &16.67 \\
        \cline{2-7}
  & Deep Seek & 60.00 & 10.00 &10.00 &6.67 &10.00\\
 \hline
        \hline
         \multirow{7}{*}{Fear} & BERT & 85.71 & 73.80 & 21.43 & 40.00 & 83.33 \\
         \cline{2-7}
           & RoBERTa & 85.71 & 84.00 & 36.66 & 46.67 & 84.50 \\
          \cline{2-7}
            & DeBERTa & 85.71 & 83.33 & 40.00 & 43.33 & 86.67 \\
          \cline{2-7}
            & DistilBERT & 83.33 & 76.67 & 23.33 & 36.67 & 83.33 \\
          \cline{2-7}
            & Flan T5 & 87.30 & 80.00 & 35.00 & 45.00 & 86.67 \\
          \cline{2-7}
         & GPT-4o & 100.00 & 93.33 &90.00 &86.67 &93.33 \\
        \cline{2-7}
  & Deep Seek & 100.00 & 86.67 &83.33 &80.00 &96.67\\
  \hline
          \hline
          \multirow{7}{*}{Happiness} & BERT & 85.71 & 85.71 & 40.47 & 81.95 & 71.42 \\
          \cline{2-7}
            & RoBERTa & 88.10 & 87.27 & 46.67 & 83.87 & 76.67 \\
          \cline{2-7}
            & DeBERTa & 88.10 & 86.67 & 66.63 & 93.54 & 86.67 \\
          \cline{2-7}
            & DistilBERT & 90.47 & 80.00 & 43.33 & 87.50 & 83.33 \\
          \cline{2-7}
            & Flan T5 & 89.00 & 85.32 & 35.54 & 87.50 & 86.67 \\
          \cline{2-7}
         & GPT-4o & 95.45 & 70.00 &70.00 &70.00 &63.33\\
        \cline{2-7}
  & Deep Seek & 87.00 & 66.67 &66.67 &73.33 &66.67\\
 \hline
          \hline
            \multirow{7}{*}{Surprise} & BERT & 88.37 & 67.00 & 9.34 & 30.00 & 61.67 \\
          \cline{2-7}
            & RoBERTa & 86.45 & 66.63 & 10.00 & 30.00 & 61.67 \\
          \cline{2-7}
            & DeBERTa & 86.05 & 66.67 & 13.33 & 26.67 & 68.34 \\
          \cline{2-7}
            & DistilBERT & 88.37 & 66.67 & 13.33 & 26.67 & 68.34 \\
          \cline{2-7}
            & Flan T5 & 89.50 & 69.00 & 11.25 & 37.00 & 70.00 \\
           
  \cline{2-7}
         & GPT-4o & 75.00 & 50.00 &40.00 &36.67 &46.67
 \\
        \cline{2-7}
  & Deep Seek & 75.00 & 56.67 &23.33 &26.67 &36.67\\

          \hline
          \hline
          \multirow{7}{*}{Sadness} & BERT & 92.00 & 58.14 & 53.49 & 37.00 & 72.34 \\
          \cline{2-7}
           & RoBERTa & 92.42 & 70.00 & 36.67 & 53.33 & 76.74 \\
          \cline{2-7}
            & DeBERTa & 91.26 & 83.33 & 50.00 & 66.67 & 79.70 \\
          \cline{2-7}
            & DistilBERT & 93.97 & 53.33 & 46.67 & 60.00 & 77.74 \\
           \cline{2-7}
             & Flan T5 & 94.00 & 63.33 & 55.00 & 62.00 & 75.00 \\
             \cline{2-7}
         & GPT-4o & 77.50 & 66.67 &66.67 &66.67 &66.67 \\
        \cline{2-7}
  & Deep Seek & 95.45 & 83.33 &83.33 &80.00 &76.67\\
 \hline
           \hline
             \multirow{7}{*}{Neutral} & BERT & 95.45 & 88.63 & 95.45 & 72.72 & 90.69 \\
           \cline{2-7}
             & RoBERTa & 95.45 & 90.00 & 92.85 & 76.76 & 96.50 \\
           \cline{2-7}
             & DeBERTa & 95.45 & 93.41 & 90.00 & 74.20 & 89.30 \\
           \cline{2-7}
            & DistilBERT & 95.45 & 93.55 & 93.33 & 85.00 & 90.00 \\
           \cline{2-7}
            & Flan T5 & 100.00 & 96.85 & 95.00 & 77.00 & 91.05 \\
            \cline{2-7}
         & GPT-4o & 95.45 & 83.33 &83.33 &80.00 &80.00 \\
        \cline{2-7}
  & Deep Seek & 100.00 & 100.00 &100.00 &100.00 &100.00
 \\
 \hline
    \end{tabular}
    
    \label{tab:dailog}
\end{table*}

\begin{table*}[t!]
    \centering\small
    \caption{Emotion Inference Models' $F_1$ Score Comparison on Dair-AI based Evaluation Dataset}
    \begin{tabular}{p{1.5cm}|p{1.5cm}|p{1.5cm}|p{1.5cm}|p{1.5cm}|p{1.5cm}|p{1.5cm}}
        \hline
        \textbf{Emotion Category} & \textbf{Inference Model} & \textbf{Original Text} & \textbf{Apple Rewrite} & \textbf{Apple \break Professional} & \textbf{Apple Friendly} & \textbf{Apple \break Concise} \\
        \hline
        \multirow{7}{*}{Anger}
        & BERT       & 98.00 & 64.10 & 35.90 & 46.15 & 87.18 \\ \cline{2-7}
        & RoBERTa    & 97.50 & 66.67 & 41.03 & 41.03 & 87.18 \\ \cline{2-7}
        & DeBERTa    & 97.56 & 67.50 & 37.50 & 35.00 & 87.50 \\ \cline{2-7}
        & DistilBERT & 100.00 & 69.23 & 61.54 & 41.03 & 92.31 \\ \cline{2-7}
        & Flan T5    & 98.56 & 66.67 & 48.72 & 30.77 & 69.23 \\ \cline{2-7}
        & GPT-4o     & 53.33 & 7.69  & 7.69  & 7.69  & 7.69  \\ \cline{2-7}
        & Deep Seek  & 67.00 & 7.69  & 7.69  & 7.69  & 7.69  \\
        \hline\hline

        \multirow{7}{*}{Sadness}
        & BERT       & 95.60 & 91.89 & 67.80 & 76.92 & 90.41 \\ \cline{2-7}
        & RoBERTa    & 98.56 & 85.71 & 72.13 & 70.97 & 93.33 \\ \cline{2-7}
        & DeBERTa    & 97.50 & 82.35 & 82.35 & 64.41 & 87.32 \\ \cline{2-7}
        & DistilBERT & 97.50 & 76.92 & 44.00 & 59.65 & 85.71 \\ \cline{2-7}
        & Flan T5    & 100.00 & 84.06 & 78.12 & 75.00 & 78.79 \\ \cline{2-7}
        & GPT-4o     & 69.00 & 53.85 & 53.85 & 53.85 & 53.85 \\ \cline{2-7}
        & Deep Seek  & 72.00 & 58.97 & 58.97 & 76.92 & 58.97 \\
        \hline\hline

        \multirow{7}{*}{Joy}
        & BERT       & 91.89 & 94.73 & 91.41 & 85.71 &96.13  \\ \cline{2-7}
        & RoBERTa    & 98.50  & 84.05 & 84.50 & 88.89 & 90.41 \\ \cline{2-7}
        & DeBERTa    & 100.00 &78.12& 84.05 & 73.01 & 78.79 \\ \cline{2-7}
        & DistilBERT & 100.00 &  70.97 & 51.85 &49.06 &78.79  \\ \cline{2-7}
        & Flan T5    & 100.00 & 28.21 & 12.82 & 15.38 & 28.21 \\ \cline{2-7}
        & GPT-4o     & 72.45  & 38.46 & 38.46 & 38.46 & 38.46 \\ \cline{2-7}
        & Deep Seek  & 66.67  & 28.21 & 28.21 & 28.21 & 28.21 \\
        \hline\hline

        \multirow{7}{*}{Love}
        & BERT       & 100.00 & 92.31 & 82.05 & 76.92 & 94.87 \\ \cline{2-7}
        & RoBERTa    & 97.30  & 74.36 & 74.36 & 82.05 & 89.18 \\ \cline{2-7}
        & DeBERTa    & 96.10  & 90.41 & 92.00 & 87.32 & 91.87 \\ \cline{2-7}
        & DistilBERT & 96.10  & 90.41 & 94.73 & 87.32 & 92.00 \\ \cline{2-7}
        & Flan T5    & 96.45  & 46.15 & 40.00 & 36.73 & 49.06 \\ \cline{2-7}
        & GPT-4o     & 77.34  & 38.46 & 38.46 & 38.46 & 38.46 \\ \cline{2-7}
        & Deep Seek  & 88.00  & 64.10 & 64.10 & 61.54 & 61.54 \\
        \hline\hline

        \multirow{7}{*}{Fear}
        & BERT       & 100.00 & 90.41 & 80.60 & 66.67 & 93.33 \\ \cline{2-7}
        & RoBERTa    & 100.00 & 87.18 & 64.10 & 64.10 & 89.74 \\ \cline{2-7}
        & DeBERTa    & 98.73  & 90.41 & 18.18 & 68.85 & 90.41 \\ \cline{2-7}
        & DistilBERT & 97.50  & 92.11 & 82.35 & 62.07 & 91.89 \\ \cline{2-7}
        & Flan T5    & 96.72  & 64.10 & 53.85 & 51.28 & 66.67 \\ \cline{2-7}
        & GPT-4o     & 85.35  & 66.67 & 66.67 & 66.67 & 66.67 \\ \cline{2-7}
        & Deep Seek  & 91.20  & 82.05 & 82.05 & 82.05 & 82.05 \\
        \hline\hline

        \multirow{7}{*}{Surprise}
        & BERT       & 96.50 & 48.72 & 15.38 & 33.33 & 48.72 \\ \cline{2-7}
        & RoBERTa    & 95.00 & 61.54 & 20.51 & 46.15 & 61.54 \\ \cline{2-7}
        & DeBERTa    & 100.00 & 60.00 & 20.00 & 45.00 & 62.50 \\ \cline{2-7}
        & DistilBERT & 100.00 & 75.00 & 33.33 & 33.33 & 70.97 \\ \cline{2-7}
        & Flan T5    & 100.00 & 60.00 & 54.54 & 66.67 & 66.67 \\ \cline{2-7}
        & GPT-4o     & 68.35 & 41.03 & 41.03 & 43.59 & 43.59 \\ \cline{2-7}
        & Deep Seek  & 57.34 & 23.08 & 23.08 & 23.08 & 23.08 \\
        \hline
    \end{tabular}
    \label{tab:emotion_interference1}
\end{table*}

\begin{table*}[t!]
    \centering
    \small
    \caption{Emotion Inference Models' $F_1$ Score Comparison on DailyDialog-based Evaluation Dataset}
    \begin{tabular}{p{1.5cm}|p{1.5cm}|p{1.5cm}|p{1.5cm}|p{1.5cm}|p{1.5cm}|p{1.5cm}}
        \hline
        \textbf{Emotion Category} & \textbf{Inference Model} & \textbf{Original} & \textbf{Apple Rewrite} & \textbf{Apple \break Professional} & \textbf{Apple Friendly} & \textbf{Apple \break Concise} \\
        \hline
        \multirow{7}{*}{Anger} 
        & BERT       & 92.50 & 78.87 & 24.50 & 14.30 & 70.00 \\ \cline{2-7}
        & RoBERTa    & 92.50 & 78.79 & 50.00 & 37.84 & 84.61 \\ \cline{2-7}
        & DeBERTa    & 92.50 & 67.50 & 37.50 & 18.88 & 87.50 \\ \cline{2-7}
        & DistilBERT & 100.00 & 80.00 & 28.57 & 3.45 & 77.57 \\ \cline{2-7}
        & Flan T5    & 92.50 & 82.30 & 18.87 & 24.50 & 85.00 \\ \cline{2-7}
        & GPT-4o     & 79.34 & 55.00 & 55.00 & 23.33 & 55.34 \\ \cline{2-7}
        & Deep Seek  & 84.55 & 75.86 & 72.41 & 10.34 & 65.52 \\
        \hline\hline

        \multirow{7}{*}{Disgust} 
        & BERT       & 86.50 & 77.14 & 24.50 & 17.02 & 80.56 \\ \cline{2-7}
        & RoBERTa    & 88.31 & 80.00 & 80.00 & 28.57 & 80.00 \\ \cline{2-7}
        & DeBERTa    & 88.30 & 80.00 & 80.00 & 33.33 & 77.57 \\ \cline{2-7}
        & DistilBERT & 86.85 & 75.00 & 75.00 & 18.19 & 77.56 \\ \cline{2-7}
        & Flan T5    & 92.66 & 77.25 & 77.25 & 22.04 & 82.35 \\ \cline{2-7}
        & GPT-4o     & 45.33 & 17.24 & 17.24 & 17.24 & 17.24 \\ \cline{2-7}
        & Deep Seek  & 72.34 & 10.34 & 10.34 & 6.90 & 10.34 \\
        \hline\hline

        \multirow{7}{*}{Fear} 
        & BERT       & 92.30 & 84.95 & 35.30 & 57.56 & 90.90 \\ \cline{2-7}
        & RoBERTa    & 92.30 & 91.30 & 53.66 & 63.63 & 91.30 \\ \cline{2-7}
        & DeBERTa    & 92.30 & 90.91 & 57.14 & 60.46 & 92.87 \\ \cline{2-7}
        & DistilBERT & 90.91 & 86.80 & 37.83 & 53.65 & 90.91 \\ \cline{2-7}
        & Flan T5    & 95.00 & 94.85 & 60.00 & 67.85 & 92.87 \\ \cline{2-7}
        & GPT-4o     & 100.00 & 93.10 & 89.66 & 86.21 & 93.10 \\ \cline{2-7}
        & Deep Seek  & 100.00 & 86.21 & 82.76 & 79.31 & 96.55 \\
        \hline\hline

        \multirow{7}{*}{Happiness} 
        & BERT       & 92.37 & 92.31 & 57.63 & 89.47 & 83.33 \\ \cline{2-7}
        & RoBERTa    & 93.67 & 93.20 & 63.63 & 91.22 & 86.79 \\ \cline{2-7}
        & DeBERTa    & 93.67 & 92.86 & 80.00 & 96.96 & 92.86 \\ \cline{2-7}
        & DistilBERT & 95.00 & 88.89 & 60.46 & 93.33 & 94.00 \\ \cline{2-7}
        & Flan T5    & 97.25 & 95.00 & 68.97 & 96.66 & 92.86 \\ \cline{2-7}
        & GPT-4o     & 87.45 & 68.97 & 68.97 & 68.97 & 62.07 \\ \cline{2-7}
        & Deep Seek  & 75.00 & 65.52 & 65.52 & 72.41 & 65.52 \\
        \hline\hline

        \multirow{7}{*}{Surprise} 
        & BERT       & 93.83 & 80.55 & 17.02 & 46.42 & 60.00 \\ \cline{2-7}
        & RoBERTa    & 92.50 & 77.55 & 18.18 & 46.15 & 76.29 \\ \cline{2-7}
        & DeBERTa    & 92.50 & 80.00 & 23.53 & 42.10 & 81.18 \\ \cline{2-7}
        & DistilBERT & 93.83 & 79.04 & 6.04 & 46.15 & 80.40 \\ \cline{2-7}
        & Flan T5    & 95.50 & 84.20 & 6.90 & 52.13 & 85.15 \\ \cline{2-7}
        & GPT-4o     & 67.34 & 51.72 & 41.38 & 37.93 & 48.28 \\ \cline{2-7}
        & Deep Seek  & 72.30 & 55.17 & 24.14 & 27.59 & 37.93 \\
        \hline
    \end{tabular}
    \label{tab:sample2}
\end{table*}
%\section{Dataset samples}\label{sec:dataset_samples}

\begin{table*}[h]
\vspace{-60px}
    \centering
    \tiny % Reduce font size to make table compact
    \renewcommand{\arraystretch}{1.2} % Improve row spacing
    \setlength{\tabcolsep}{2pt} % Adjust column spacing for better fit
    \caption{Evaluation of dataset 1 Emotion: Angry.}
    \label{tab:emotion_angry}
    \begin{tabular}{|p{3cm}|p{1.2cm}|p{3cm}|p{3cm}|p{3cm}|p{3cm}|}
    \hline
    \textbf{Original Text} & \textbf{Type} & \textbf{Apple Rewrite} & \textbf{Apple Professional} & \textbf{Apple Friendly} & \textbf{Apple Concise} \\
    \hline
    "I'm grabbing a minute to post; I feel greedy and wrong." & Angry & I’m taking a moment to post, feeling guilty and selfish. & In this brief moment, I feel compelled to share my thoughts, despite the temptation to keep them to myself. & I’m taking a quick break to share; I feel a bit guilty and selfish. & I feel greedy and wrong, so I’m taking a moment to post. \\
    \hline
    "I am feeling grouchy." & Angry & I’m feeling a bit grumpy today. & I am experiencing a state of discontent. & I am feeling a little grumpy today. & I’m feeling grumpy. \\
    \hline
    "I think it's the easiest time of year to feel dissatisfied." & Angry & I believe it’s the most opportune time of the year to experience dissatisfaction. & The most opportune time of year to experience dissatisfaction is during the winter months. & I think it’s the easiest time of year to feel a little down. & I believe it’s the easiest time to feel dissatisfied. \\
    \hline
    "I feel irritated and rejected without anyone doing anything or saying anything." & Angry & I feel irritated and rejected simply because no one has done or said anything. & I am experiencing a sense of irritation and rejection that is not attributable to any specific actions or words from others. & I’m feeling a bit down and need some support. & I feel irritated and rejected without any action or words. \\
    \hline
    "I already feel like I messed up, though, because I don't usually eat at all in the morning." & Angry & I already feel like I’ve made a mistake because I don’t usually eat anything in the morning. & Despite my efforts, I am already experiencing a sense of inadequacy, as I am not accustomed to consuming breakfast. & I’m feeling a bit self-conscious about my morning meal, as I’m not typically a morning eater. & I already feel like I’ve made a mistake because I rarely eat in the morning. \\
    \hline
    "I feel kind of appalled that she feels like she needs to explain at length her body measurements, etc." & Angry & I find it rather disheartening that she feels compelled to provide an extensive explanation of her body measurements and other physical attributes. & I find it rather disconcerting that she feels compelled to provide an extensive explanation of her physical attributes, including her body measurements. & It’s important to respect people’s privacy and not feel pressured to share personal details. & I’m appalled that she feels compelled to explain her body measurements in detail. \\
    \hline
    "I climbed the hill feeling frustrated that I'd completely miscalculated my pacing for this course, and a factor that has never hampered me before made such a dent in my day." & Angry & I climbed the hill feeling frustrated that I had completely underestimated my pacing for this course. A factor that had never affected me before made a significant dent in my day. & As I ascended the hill, I was filled with frustration, realizing that I had significantly underestimated my pacing for this course. A factor that had never previously affected my progress had unexpectedly hindered my progress. & I climbed the hill feeling super frustrated because I totally messed up my pacing for this course. It’s like something that’s never happened to me before, and it totally ruined my day. & Frustrated by my complete misjudgment of pacing for the course, a factor that never affected me before, I climbed the hill. \\
    \hline
    "I feel a bit rude writing to an elderly gentleman to ask for gifts because I feel a bit greedy, but what is Christmas about if not mild greed?" & Angry & I feel a bit impolite writing to an elderly gentleman to ask for gifts because I feel a bit greedy, but what is Christmas about if not indulging in a bit of mild greed? & Although I acknowledge that writing to an elderly gentleman to request gifts may seem somewhat impolite, I cannot help but wonder if the essence of Christmas lies in indulging in a modicum of greed. & I’m a bit hesitant to ask for gifts from an elderly gentleman, as I feel a bit greedy. But what’s Christmas all about if not indulging in a little bit of greed? & I feel a bit rude asking an elderly gentleman for gifts, but what’s Christmas without a bit of greed? \\
    \hline
    "I stopped feeling cold and began feeling hot." & Angry & I stopped feeling cold and started feeling hot. & I experienced a sudden shift from a sensation of coldness to warmth. & I went from feeling chilly to toasty in an instant! & I stopped feeling cold and felt hot instead. \\
    \hline
    "I feel selfish and spoiled." & Angry & I feel guilty and indulgent. & I experience a sense of self-centeredness and indulgence. & I’m feeling a bit guilty about indulging in my own desires. & I feel selfish and pampered. \\
    \hline
    "I felt anger at the end of a telephone call." & Angry & I felt angry at the end of a phone call. & I experienced intense anger at the conclusion of a telephone conversation. & I felt a surge of rage at the end of a phone call. & I felt angry after a phone call. \\
    
    \hline
    "I think about the things I've said and the stuff I have done—it makes me feel disgusted in myself when I should be making you happy and smile, which I was far from doing." & Angry & I often reflect on my words and actions, which fills me with self-disgust. I should be making you happy and smiling, but I fell far short of that. & Upon reflection on my actions and words, I am filled with a profound sense of self-loathing, despite my intention to bring joy and happiness to you, which I clearly failed to accomplish. & I often reflect on my actions and words, and it fills me with regret. I should have been making you happy and bringing a smile to your face, but I fell short of that goal. & I regret my actions and words, which disgust me as I should’ve made you happy and smile, but I failed to do so. \\
    \hline
    "I am just feeling cranky and blue." & Angry & I’m feeling a bit down and irritable today. & I am experiencing a temporary state of irritability and melancholy. & I am feeling a bit down and need some time to recharge. & I’m feeling cranky and blue. \\
    \hline
    "I feel bitchy but not defeated yet." & Angry & I feel a bit bitter, but I haven’t lost hope yet. & Although I am experiencing some frustration, I have not yet succumbed to defeat. & I’m feeling a bit frustrated, but I’m not giving up. & I feel bitter but not defeated yet. \\
    \hline
    "I know what it feels like," he stressed, glaring down at her as she squeezed more soap onto her sponge. & Angry & “I understand what it’s like,” he emphasized, his gaze fixed on her as she added more soap to her sponge. & “I understand your perspective,” he emphasized, his gaze fixed on her as she continued to apply excessive soap to her sponge. & “I totally get it,” he emphasized, looking down at her as she added more soap to her sponge. & “I understand,” he stressed, glaring at her as she added more soap to her sponge. \\
    \hline
    "I feel like taking a whack at someone’s eye and spitting on it—a cranky old lady trying to cheer myself up." & Angry & I have an urge to strike someone’s eye and spit on it—a desperate attempt to cheer myself up by lashing out at a cranky old lady. & I harbor a desire to inflict physical harm upon an individual’s ocular region and subsequently expel bodily fluids upon them—an act of vengeance motivated by a state of emotional distress. & I’m feeling frustrated and irritable, and it’s hard to find a way to calm down. & I feel like attacking someone’s eye and spitting on it—a cranky old lady trying to cheer me up. \\
    \hline
    "I feel MMF, and I can't be bothered to fight it." & Angry & I’m feeling moody and irritable, and I don’t have the energy to fight it. & I am experiencing a strong desire to masturbate, and I lack the motivation to resist it. & I’m feeling a bit under the weather, and I’m not feeling up to fighting it. & I feel MMF and can’t fight it. \\
    \hline
    "I know it's so difficult, especially when you feel you have been wronged." & Angry & I understand that it’s incredibly challenging, especially when you feel you’ve been mistreated. & I understand that it can be particularly challenging, especially when you perceive yourself as having been unjustly treated. & I know it’s tough, especially when you feel you’ve been wronged. & I know it’s hard, especially when you feel wronged. \\
    \hline
    "I feel so damn disgusted, violated, hurt, and angry—everything." & Angry & I am filled with an overwhelming sense of disgust, violation, hurt, and anger—every emotion imaginable. & I am deeply disturbed, violated, emotionally wounded, and enraged by this situation. & I’m feeling a range of intense emotions right now, including disgust, violation, hurt, and anger. & I feel disgusted, violated, hurt, and angry. \\
    \hline
    "I feel mad that you grabbed the toy." & Angry & I’m furious that you took the toy. & I am indignant that you took possession of the toy. & I was really looking forward to playing with that toy. & I’m angry you grabbed the toy. \\
    \hline
    "I've been wrestling with feeling jealous and envious of my GF's other BF since he's been staying with her for a while." & Angry & I’ve been grappling with feelings of jealousy and envy towards my girlfriend’s other boyfriend since he’s been staying with her for an extended period. & I have been grappling with feelings of jealousy and envy towards my girlfriend’s other boyfriend since he has been residing with her for an extended period. & I’ve been feeling a bit jealous and envious of my girlfriend’s other boyfriend since he’s been staying with her for a while. & I’ve been feeling jealous and envious of my girlfriend’s other boyfriend since he’s been staying with her. \\
    \hline
    "If I could vocalize my feelings here, I would put in a sarcastic 'Great.'" & Angry & If I could express my emotions here, I would respond with a sarcastic “Great.” & If I were permitted to express my sentiments here, I would respond with a sarcastic remark, stating, “Indeed.” & If I could express my feelings here, I would say, “Oh, that’s wonderful.” & If I could express my feelings here, I’d say sarcastically, “Great.” \\
    \hline
    "I was so busy analyzing what’s wrong that I ended up feeling bitter about the things that made me happy before." & Angry & I was so engrossed in analyzing my shortcomings that I ended up harboring resentment towards the very things that had previously brought me joy. & In my preoccupation with identifying the flaws, I inadvertently developed a sense of resentment towards the very elements that had previously brought me joy. & I got so caught up in analyzing what went wrong that I ended up feeling resentful of the things that used to bring me joy. & I was so focused on analyzing my problems that I became bitter about my happiness. \\
    \hline
    "I feel irritable, supersensitive." & Angry & I feel irritable and overly sensitive. & I am experiencing heightened irritability and heightened sensitivity. & I’m feeling a bit on edge, like I’m easily annoyed. & I feel irritable and oversensitive. \\
    \hline
    "I was sitting in the corner, stewing in my own muck, feeling hated, alone, unworthy, and violated." & Angry & I was sitting in the corner, feeling despondent and alone, as if I had done something wrong and was being punished. & In a state of profound emotional distress, I found myself isolated and devoid of self-worth, experiencing feelings of hatred and violation. & I was feeling a bit down and isolated, and it was tough to shake off those negative feelings. & I sat alone, feeling hated, unworthy, and violated. \\
    \hline
    \end{tabular}
\end{table*}

\begin{table*}[h]
\vspace{-60px}
    \centering
    \tiny % Reduce font size to make table compact
    \renewcommand{\arraystretch}{1.2} % Improve row spacing
    \setlength{\tabcolsep}{2pt} % Adjust column spacing for better fit
    \caption{Evaluation of dataset 1 Emotion: Sadness.}
    \label{tab:emotion_sadness}
    \begin{tabular}{|p{3cm}|p{1.2cm}|p{3cm}|p{3cm}|p{3cm}|p{3cm}|}
    \hline
    \textbf{Original Text} & \textbf{Type} & \textbf{Apple Rewrite} & \textbf{Apple Professional} & \textbf{Apple Friendly} & \textbf{Apple Concise} \\
    \hline
    "I didn’t feel humiliated." & Sadness & I didn’t feel humiliated. & I did not experience any feelings of humiliation. & I didn’t feel humiliated, I felt embarrassed. & I wasn’t humiliated. \\
    \hline
    "I can go from feeling so hopeless to so damned hopeful just from being around someone who cares and is awake." & Sadness & I can experience a profound shift in my emotional state, from utter hopelessness to overwhelming hope, simply by being in the presence of someone who cares and is fully present. & The presence of someone who cares and is awake can transform my state of mind, shifting from profound hopelessness to overwhelming hope. & Being around someone who cares and is awake can instantly shift my mood from hopeless to hopeful. & I can go from hopelessness to hopefulness just by being around someone who cares and is awake. \\
    \hline
    "I've been feeling a little burdened lately, wasn't sure why that was." & Sadness & Lately, I’ve been feeling a bit overwhelmed, but I couldn’t quite pinpoint the reason behind it. & Lately, I have been experiencing a sense of weight on my shoulders, but I have been unable to pinpoint the exact source of this burden. & I’ve been feeling a bit overwhelmed lately, I’m not sure why that is. & I’ve been feeling burdened lately, but I wasn’t sure why. \\
    \hline
    "I feel like I have to make the suffering I’m seeing mean something." & Sadness & I feel compelled to give meaning to the suffering I witness. & I am compelled to interpret the suffering I observe as having a profound significance. & I want to find meaning in the suffering I witness. & I feel compelled to give meaning to the suffering I witness. \\
    \hline
    "I feel low energy, I’m just thirsty." & Sadness & I feel low on energy, but I’m just thirsty. & I am experiencing a lack of energy, which I believe is primarily due to dehydration. & I feel low energy, I’m just thirsty. & I feel low on energy, I’m just thirsty. \\
    \hline
    "I didn’t really feel that embarrassed." & Sadness & I didn’t feel particularly embarrassed. & I did not experience significant embarrassment. & I didn’t feel that embarrassed, I felt like a total goofball! & I wasn’t embarrassed. \\
    \hline
    "I feel pretty pathetic most of the time." & Sadness & I often feel quite pathetic. & I experience a profound sense of inadequacy on a regular basis. & I’m feeling a bit down and need some support. & I feel pathetic most of the time. \\
    \hline
    "I started feeling sentimental about dolls I had as a child and so began a collection of vintage Barbie dolls from the sixties." & Sadness & I developed a sentimental attachment to the dolls I had as a child, leading me to embark on a collection of vintage Barbie dolls from the sixties. & As I reminisced about my childhood, I developed a sentimental attachment to the dolls I had. Consequently, I embarked on a collection of vintage Barbie dolls from the 1960s. & I started feeling nostalgic about my childhood dolls and so I started collecting vintage Barbie dolls from the sixties. & I began collecting vintage Barbie dolls from the sixties after feeling sentimental about dolls I had as a child. \\
    \hline
    "I still love my significant other and wish the best for him. I can no longer tolerate the effect that BM has on our lives and the fact that it has turned my SO into a bitter, angry person who is not always particularly kind to the people around him when he is feeling stressed." & Sadness & I still love my significant other and wish him the best. However, I can no longer bear the impact of my brother-in-law’s behavior on our lives. It has transformed my SO into a bitter and angry person who is not always kind to those around him when he is stressed. & Despite my continued affection for my significant other, I am compelled to express my dissatisfaction with the impact of his behavior, particularly his strained relationship with his brother. This has resulted in a transformation of my partner into a bitter and irritable individual, whose interactions with others become increasingly hostile when he experiences stress. & I still care about my partner deeply, but I’m struggling with the impact of his behavior. It’s tough to see him become so bitter and angry, and it’s affecting his relationships with others. I want to support him, but I also need to prioritize my own well-being. & I still love my significant other and wish him the best. I can’t tolerate BM’s effect on our lives. It’s turned him bitter, angry, and unkind when stressed. \\
    \hline
    "I feel so inhibited in someone else’s kitchen, like I’m painting on someone else’s picture." & Sadness & I feel incredibly restricted in someone else’s kitchen, as if I’m painting on their canvas without their permission. & In the presence of another person’s kitchen, I experience a profound sense of inhibition, akin to being compelled to paint on an existing canvas. & I find it challenging to express myself freely in someone else’s kitchen, as if I’m trying to fit into their preconceived ideas. & I feel inhibited in someone else’s kitchen, as if I’m painting on their picture. \\
    \hline
    "I'm feeling rather rotten, so I'm not very ambitious right now." & Sadness & I’m feeling quite unwell, so I’m not feeling particularly ambitious at the moment. & Currently, I am experiencing a state of indisposition and lack the energy to pursue ambitious endeavors. & I’m feeling under the weather, so I’m not feeling too up for anything too big right now. & I’m unwell, so I’m not ambitious. \\
    \hline
    "I'm updating my blog because I feel shitty." & Sadness & I’m updating my blog because I’m feeling down. & I am updating my blog because I am experiencing a sense of dissatisfaction. & I’m updating my blog because I’m feeling a bit down. & I’m updating my blog because I feel unwell. \\
    \hline
    "I never make her separate from me because I don’t ever want her to feel like I’m ashamed of her." & Sadness & I never separate myself from her because I never want her to feel ashamed of me. & I never treat her differently from me because I never want her to feel ashamed of me. & I always include her in my activities because I want her to feel proud of who she is. & I never separate her from me because I don’t want her to feel ashamed of me. \\
    \hline
    "I was feeling a little vain when I did this one." & Sadness & I felt a bit vain when I did this. & I experienced a moment of vanity while undertaking this task. & I was feeling a bit self-conscious when I did this one. & I felt a bit vain when I did this. \\
    \hline
    "I feel beautifully emotional knowing that these women, of whom I knew just a handful, were holding me and my baba on our journey." & Sadness & I feel deeply emotional knowing that these women, whom I knew only a handful of, were there to hold me and my baby during our journey. & It fills me with profound emotional gratitude to know that these remarkable women, whom I had the privilege of knowing only a few, were there to support and comfort my baby and me during our journey. & It warms my heart to know that these women, whom I only knew a few, were there to support us on our journey. & I feel deeply emotional knowing that these few women, whom I knew, held my baby and me on our journey. \\
    \hline
    "I was feeling as heartbroken as I’m sure Katniss was." & Sadness & I was feeling as heartbroken as I’m certain Katniss was. & I experienced a profound sense of emotional anguish, akin to the profound sorrow experienced by Katniss. & I was feeling as down as I’m sure Katniss was. & I felt as heartbroken as Katniss did. \\
    \hline
    "I feel like my only role now would be to tear your sails with my pessimism and discontent." & Sadness & I sense that my sole purpose in life has become to dishearten and disillusion you with my pessimism and discontent. & At present, I perceive my sole purpose as being to relentlessly criticize and undermine your aspirations and endeavors. & I’m feeling a bit overwhelmed and uncertain about the future, and I’m worried that my pessimism might be affecting our team’s morale. & I feel my only role now is to dampen your spirits with my pessimism and discontent. \\
    \hline
    "I’m not sure the feeling of loss will ever go away, but it may dull to a sweet feeling of nostalgia at what I shared in this life with my dad and the luck I had to have a dad for years." & Sadness & I’m not certain if the feeling of loss will ever completely disappear, but it might gradually transform into a comforting sense of nostalgia, cherishing the memories I shared with my dad and the fortunate blessing of having a father for many years. & While the profound sense of loss may persist, it is possible that over time, it will gradually diminish into a bittersweet recollection of the cherished moments shared with my father during our lives together. & I’m not sure if the feeling of loss will ever fully go away, but it might become a warm and comforting feeling of nostalgia as I remember the wonderful times I had with my dad and the incredible luck I had to have a dad for so many years. & I’m not sure the feeling of loss will ever fully subside, but it may transform into a bittersweet nostalgia for the life I shared with my dad and the fortunate years I had a father. \\
    \hline
    "I also tell you in hopes that anyone who is still feeling stigmatized or ashamed of their mental health issues will let go of the stigma, let go of the shame." & Sadness & I also share this message in the hope that anyone who is still grappling with feelings of stigma or shame surrounding their mental health challenges will find the strength to shed these burdens. & In addition, I share this information with the hope that anyone who is still experiencing stigmatization or shame associated with their mental health challenges will release these negative emotions and embrace a more accepting attitude. & I also want to encourage anyone who’s still feeling ashamed or stigmatized about their mental health issues to let go of those feelings. & I also hope anyone still feeling stigmatized or ashamed of their mental health issues will let go of that stigma and shame. \\
    \hline
    "I'm feeling quite sad and sorry for myself, but I'll snap out of it soon." & Sadness & I’m feeling quite down and sorry for myself, but I’m sure I’ll feel better soon. & At this moment, I am experiencing a profound sense of melancholy and self-pity. However, I am confident that I will overcome this emotional state in the near future. & I’m feeling a bit down, but I’ll bounce back soon! & I’m feeling sorry for myself, but I’ll soon feel better. \\
    \hline
    "I feel like I am still looking at a blank canvas, blank pieces of paper." & Sadness & I feel like I’m still staring at a blank canvas, with no idea what to create. & At present, I perceive myself as still grappling with the initial stages of a creative endeavor, akin to gazing upon an undifferentiated canvas and blank sheets of paper. & I feel like I am still looking at a blank canvas, blank pieces of paper. & I feel like I’m still staring at a blank canvas. \\
    \hline
    "I'm updating my blog because I feel shitty." & Sadness & I’m updating my blog because I’m feeling down. & I am updating my blog because I am experiencing a sense of dissatisfaction. & I’m updating my blog because I’m feeling a bit down. & I’m updating my blog because I feel unwell. \\
    \hline
    \end{tabular}
\end{table*}

\begin{table*}[h]
\vspace{-60px}
    \centering
    \tiny % Reduce font size for compactness
    \renewcommand{\arraystretch}{1.2} % Improve row spacing
    \setlength{\tabcolsep}{2pt} % Adjust column spacing to fit content
    \caption{Evaluation of dataset 1 Emotion: Love.}
    \label{tab:emotion_love}
    \begin{tabular}{|p{3cm}|p{1.2cm}|p{3cm}|p{3cm}|p{3cm}|p{3cm}|}
    \hline
    \textbf{Original Text} & \textbf{Type} & \textbf{Apple Rewrite} & \textbf{Apple Professional} & \textbf{Apple Friendly} & \textbf{Apple Concise} \\
    \hline
    "I am ever feeling nostalgic about the fireplace; I will know that it is still on the property." & Love & I can’t help but feel nostalgic about the fireplace; I know it’s still on the property. & The fireplace evokes a profound sense of nostalgia within me, reminding me of the presence of this cherished feature on the property. & I can’t help but feel a sense of nostalgia whenever I think about the fireplace. It’s like a piece of our home that will always be there. & I always feel nostalgic about the fireplace, knowing it’s still there. \\
    \hline
    "I feel romantic too." & Love & I share a romantic sentiment as well. & I share a similar sentiment. & I feel romantic too, let’s go on a date! & I feel romantic too. \\
    \hline
    "I can’t let go of that sad feeling that I want to be accepted here in this first home of mine." & Love & I’m struggling to shake off that lingering sense of sadness that I yearn for acceptance in this very place that I’ve made my first home. & I am unable to release the lingering sense of melancholy that permeates my desire for acceptance within this, my first home. & I’m feeling a bit homesick and longing for a sense of belonging in this new place. & I can’t let go of the sadness of wanting to be accepted here, my first home. \\
    \hline
    "I was ready to meet Mom in the airport and feel her ever-supportive arms around me." & Love & I eagerly anticipated meeting Mom at the airport, anticipating her ever-supportive embrace. & I eagerly anticipated meeting my mother at the airport, anticipating the comforting embrace of her arms. & I was so excited to meet Mom in the airport and feel her warm embrace. & I eagerly awaited Mom’s embrace at the airport. \\
    \hline
    "I talk to dogs as I feel they cannot understand words, but they can read emotions and know how to be supportive. I decided I should go home." & Love & I talk to dogs because I believe they can’t comprehend words, but they can sense emotions and offer support. Therefore, I decided to head home. & In contrast, I engage in conversations with dogs, as I believe they lack the capacity to comprehend spoken language. However, they possess the ability to discern emotions and demonstrate a knack for providing support. Consequently, I have made the decision to return home. & I have a unique way of communicating with dogs. I believe they can’t understand words, but they can definitely sense our emotions and offer us support. So, I decided to head home. & I talk to dogs because I believe they can’t understand words but can read emotions and be supportive. I decided to go home. \\
    \hline
    "I won’t let my child cry it out because I feel that loving her and Lily when she was little was going to be an opportunity that only lasted for those short few months." & Love & I won’t let my child cry it out because I believe that cherishing and loving her and Lily when they were infants was a fleeting opportunity that would only last for those brief few months. & I refrained from implementing the “crying it out” method for my child because I believed that nurturing and loving her during her early infancy would be a fleeting opportunity that would only last for those brief few months. & I won’t let my child cry it out because I cherish every moment I had with Lily when she was little. I believe that loving and bonding with her during those precious few months was a unique opportunity that I don’t want to miss. & I won’t let my child cry it out because I believe loving her and Lily when they were young was a fleeting opportunity. \\
    \hline
    "I feel like a naughty schoolgirl because I am falling behind." & Love & I feel like a naughty schoolgirl because I’m falling behind. & I am experiencing a sense of inadequacy due to my academic progress. & I am feeling a bit behind and need to catch up. & I feel like a naughty schoolgirl because I’m falling behind. \\
    \hline
    "If I were to go overseas or cross the border, then I become a foreigner and will feel that way—but never in my beloved land." & Love & If I were to travel overseas or cross the border, I would become a foreigner and experience that feeling—but never in my beloved land. & If I were to travel abroad or cross a border, I would become a foreigner and experience that sensation—but never in my beloved homeland. & If I were to travel abroad or cross the border, I’d feel like a foreigner, but I’d never feel that way in my own country. & If I travel overseas or cross the border, I’ll feel foreign, but never in my beloved land. \\
    \hline
    "I want each of you to feel my gentle embrace." & Love & I want each of you to feel my gentle embrace. & I extend my warmest embrace to each of you. & I want each of you to feel my warm, loving hug. & I want each of you to feel my gentle touch. \\
    \hline
    "I love neglecting this blog, but sometimes I feel for my faithful readers." & Love & I love neglecting this blog, but sometimes I feel sorry for my loyal readers. & While I occasionally neglect this blog, I feel empathy for my devoted readers. & I love neglecting this blog, but sometimes I feel for my loyal readers. & I love neglecting this blog, but I sometimes feel sorry for my readers. \\
    \hline
    "I feel—is he generous?" & Love & I sense—is he being generous? & I sense a lack of generosity in his actions. & I feel—is he generous? & Is he generous? \\
    \hline
    "I feel very misled by someone that I really, really thought I knew and liked very much." & Love & I feel deeply deceived by someone I genuinely believed I knew and admired. & I am deeply disappointed by a betrayal from someone I held in high regard. & I was taken aback by a situation that led me to question my relationship with someone close. & I feel misled by someone I thought I knew and liked. \\
    \hline
    "I’m feeling generous today. Here’s one more—you may have already seen it, but it’s good for a chuckle." & Love & I’m feeling generous today. Here’s another one—you might have seen it, but it’s worth a chuckle. & Today, I am feeling generous and pleased to share another lighthearted video, even if you’ve seen it before. & I’m feeling generous today! Here’s another one—it’s still worth a good laugh! & Feeling generous today, here’s another one worth a chuckle. \\
    \hline
    "I feel that I need to know that I can depend on myself before I put myself in the position of supporting someone else and being supported by someone else." & Love & I believe it’s crucial to establish self-reliance before supporting others and expecting support in return. & It is imperative for me to establish my own reliability before assuming the responsibility of supporting another individual. & I need to feel confident in my own abilities before I offer support to others, and I also need to feel supported in return. & I need to depend on myself before supporting and being supported by others. \\
    \hline
    "I try my best to love on them, shed some light, but I feel deeply compassionate with their problems and hurt—even if it’s someone in the media." & Love & I strive to show love and compassion towards them, shedding light on their struggles and pain. & I endeavor to extend my love and support to those in need, even when it involves individuals in the media. & I do my best to spread love and positivity, but I deeply empathize with their struggles, even in the media. & I try to love and support them, but I deeply sympathize with their pain, even if it’s someone in the media. \\
    \hline
    "I feel like a faithful servant." & Love & I feel like a devoted and loyal servant. & I feel like a devoted and loyal servant. & I feel like a loyal servant. & I feel devoted. \\
    \hline
    "I feel like I’m caring about my body—not just in an attempt to be the right size but to feel good and have a full life." & Love & I care about my body not just to achieve a certain size but to feel good and live a fulfilling life. & I have developed a newfound appreciation for my body, not solely for achieving a specific size, but for the well-being it brings me. & I’m taking care of my body not just to look good, but to feel great and live life fully! & I care about my body not just for size, but to feel good and live fully. \\
    \hline
    "I miss our talks, our cuddling, our kissing, and the feelings that you can only share with your beloved." & Love & I miss our conversations, cuddles, kisses, and the emotions shared only with a beloved. & I deeply miss our conversations, our intimate embraces, and our tender moments together. & I miss our chats, snuggles, kisses, and the special bond we share. & I miss our talks, cuddles, and the intimate feelings shared only with a beloved. \\
    \hline
    "I try to share what I bake with a lot of people because I love people, and I want them to feel loved." & Love & I share my baked goods with many because I love them and want to spread joy. & I endeavor to share my baked goods widely because I value people and want to convey warmth and affection. & I love sharing my baked goods with others to make them feel loved and appreciated. & I love sharing my baked goods because I want people to feel loved. \\
    \hline
    "I have a lot to learn—I feel like people are supportive of me." & Love & I feel like people are supportive of me, and I have a lot to learn. & I recognize my need for growth and appreciate the support of those around me. & I have a lot to learn—I feel like people are my cheerleaders! & I feel supported and have much to learn. \\
    \hline
    "I feel for you despite the bitterness and longing." & Love & I empathize with you, even though you’re feeling bitter and yearning. & Despite the bitterness and longing, I empathize with your situation. & I feel for you, even though it hurts. & I sympathize despite the bitterness and longing. \\
    \hline
    "I feel about being naughty for breast cancer awareness." & Love & I have mixed feelings about wearing naughty outfits for breast cancer awareness. & I hold mixed feelings regarding the appropriateness of engaging in naughty behavior for breast cancer awareness. & I feel about being naughty for breast cancer awareness. & I feel conflicted about being naughty for breast cancer awareness. \\
    \hline
    "I don’t know you or what you’re going through, but I feel sympathetic because I’m human." & Love & I may not know you or your situation, but as a fellow human being, I can empathize with your feelings. & Although I am not familiar with your situation or the challenges you are facing, I empathize with you because I am a fellow human being. & I don’t know you or what you’re going through, but I can totally relate because I’m human too. & I feel sympathetic because I’m human, even though I don’t know you or your situation. \\
    \hline
    "I’d feel nostalgic about Gillard hours ago." & Love & I would have felt nostalgic about Gillard just a few hours ago. & I would have felt nostalgic about Gillard only a few hours ago. & I’d feel nostalgic about Gillard hours ago. & I felt nostalgic about Gillard a few hours ago. \\
    \hline
    "I feel like we should write a book on all of this because we are loving researching it, and it is so good." & Love & I believe we should write a book on this topic because we are thoroughly enjoying our research, and the subject matter is incredibly interesting. & I propose that we compile a comprehensive book on this subject, as our research has been thoroughly enjoyable and the subject matter is of great interest. & I think we should write a book about all this because we’re having so much fun researching it, and it’s turning out to be really interesting. & We should write a book about this because we love researching it and it’s great. \\
    \hline
    "I can’t help but feel that I need to be delicate." & Love & I can’t help but feel the need to be delicate. & I find myself compelled to exercise caution and restraint in my actions. & I need to be careful not to offend anyone. & I feel the need to be delicate. \\
    \hline
    "I just really want this healthy lifestyle to become a habit instead of a necessity because, at the moment, I feel like a naughty child being denied the biscuit tin and angry for letting myself put weight on in the first place." & Love & I genuinely desire this healthy lifestyle to become a habit rather than a necessity. Currently, I feel like a mischievous child being denied the biscuit tin and resentful for allowing myself to gain weight. & I genuinely desire to establish a healthy lifestyle as a habit rather than a necessity. Currently, I find myself in a state of frustration, akin to a child being denied a treat and expressing anger for indulging in weight gain. & I’m trying to make healthy choices a regular part of my routine, but it feels like a constant struggle. I’m not sure if I’m doing enough to make it a habit, and I’m feeling a bit frustrated and self-conscious about my weight. & I want this healthy lifestyle to become a habit, not a necessity. I feel like a naughty child denied the biscuit tin and angry for gaining weight. \\
    \hline
    \end{tabular}
\end{table*}

\begin{table*}[h]
\vspace{-60px}
    \centering
    \tiny % Reduce font size for compactness
    \renewcommand{\arraystretch}{1.2} % Improve row spacing
    \setlength{\tabcolsep}{2pt} % Adjust column spacing to fit content
    \caption{Evaluation of dataset 1 Emotion: Joy.}
    \label{tab:emotion_joy}
    \begin{tabular}{|p{3cm}|p{1.2cm}|p{3cm}|p{3cm}|p{3cm}|p{3cm}|}
    \hline
    \textbf{Original Text} & \textbf{Type} & \textbf{Apple Rewrite} & \textbf{Apple Professional} & \textbf{Apple Friendly} & \textbf{Apple Concise} \\
    \hline
    "I have been with Petronas for years; I feel that Petronas has performed well and made a huge profit." & Joy & I’ve been a loyal Petronas customer for years, and I’ve witnessed the company’s remarkable success and substantial profits. & Over the years, I have been a loyal employee of Petronas. I have witnessed the company’s consistent performance and substantial financial success. & I’ve been with Petronas for ages, and I’ve seen the company grow and thrive. They’ve made a ton of money and done a great job. & I’ve been with Petronas for years, and I believe it’s performed well and made significant profits. \\
    \hline
    "I do feel that running is a divine experience and that I can expect to have some type of spiritual encounter." & Joy & I genuinely believe that running is a divine experience, and I anticipate having some kind of spiritual encounter during my runs. & I genuinely believe that running is an extraordinary experience that can lead to profound spiritual encounters. & I absolutely believe that running is a divine experience and that I’m bound to have some sort of spiritual encounter on the trails. & I believe running is a divine experience that will likely lead to a spiritual encounter. \\
    \hline
    "I was able to help Chai Lifeline with your support and encouragement—it is a great feeling, and I am so glad you were able to help me." & Joy & I was able to assist Chai Lifeline with your support and encouragement. It’s a wonderful feeling, and I’m so grateful that you were able to help me. & I was able to assist Chai Lifeline with your support and encouragement. It is a gratifying experience, and I am delighted that you were able to assist me. & I’m thrilled to share that I was able to assist Chai Lifeline with your support and encouragement. It’s an amazing feeling, and I’m so grateful for your help. & I helped Chai Lifeline with your support and encouragement. It feels great, and I’m glad you helped me. \\
    \hline
    "I get giddy over feeling elegant in a perfectly fitted pencil skirt." & Joy & I feel a rush of excitement when I slip into a pencil skirt that’s been expertly tailored to my figure. & I experience a profound sense of elation when donning a pencil skirt that is impeccably tailored to my figure. & I get a rush of excitement when I slip into a perfectly fitted pencil skirt. & I get excited about feeling elegant in a perfectly fitted pencil skirt. \\
    \hline
    "I finally feel I have accepted Nashville as home." & Joy & I’ve finally come to terms with Nashville as my home. & I have finally come to the realization that Nashville has become my permanent residence. & I’ve officially embraced Nashville as my new home. & I’ve finally accepted Nashville as my home. \\
    \hline
    "I just feel extremely comfortable with the group of people that I don’t even need to hide myself." & Joy & I feel incredibly comfortable with the group of people that I don’t even feel the need to hide myself. & I find a profound sense of ease and authenticity within the company of individuals with whom I can freely express myself. & I feel so at ease and comfortable with the group of people that I can be my true self. & I feel extremely comfortable with the group of people I don’t even need to hide myself from. \\
    \hline
    "I like to have the same breathless feeling as a reader, eager to see what will happen next." & Joy & I enjoy experiencing the same exhilarating sensation as a reader, eagerly anticipating the unfolding events. & I derive immense pleasure from experiencing the same exhilarating sensation as a reader, eagerly anticipating the unfolding of the narrative. & I love to get lost in a good book, feeling like I’m right there with the characters, eager to see what happens next. & Like a reader, eager to see what happens next, I enjoy the breathless feeling. \\
    \hline
    "I left with my bouquet of red and yellow tulips under my arm, feeling slightly more optimistic than when I arrived." & Joy & I left with a bouquet of vibrant red and yellow tulips under my arm, feeling a slight glimmer of optimism that I hadn’t felt when I arrived. & Departing with my bouquet of vibrant red and yellow tulips, I experienced a slight increase in optimism compared to my arrival. & I left with my bouquet of red and yellow tulips under my arm, feeling a little brighter than when I walked in. & I left with my bouquet of red and yellow tulips, feeling slightly more optimistic. \\
    \hline
    "I talk to dogs as I feel they cannot understand words, but they can read emotions and know how to be supportive. I decided I should go home." & Joy & I talk to dogs because I believe they can’t comprehend words, but they can sense emotions and offer support. Therefore, I decided to head home. & In contrast, I engage in conversations with dogs, as I believe they lack the capacity to comprehend spoken language. However, they possess the ability to discern emotions and demonstrate a knack for providing support. Consequently, I have made the decision to return home. & I have a unique way of communicating with dogs. I believe they can’t understand words, but they can definitely sense our emotions and offer us support. So, I decided to head home. & I talk to dogs because I believe they can’t understand words but can read emotions and be supportive. I decided to go home. \\
    \hline
    "I am very excited to finally meet that companion—the companion who will be with me at all times, especially when I am lonely—very lonely. The companion who will never disappoint me, the companion who will put his arms around me and make me feel loved." & Joy & I am thrilled to finally meet that constant companion—the one who will be by my side, especially during those lonely moments. The companion who will never let me down, the one who will embrace me and shower me with love. & I am filled with anticipation as I prepare to meet my lifelong companion. This companion will be by my side through thick and thin, especially during moments of loneliness. It will never fail me, offering unwavering support and unconditional love. & I can’t wait to finally meet that special someone—the one who’ll be by my side through thick and thin, especially when I’m feeling down. The one who’ll never let me down, the one who’ll wrap their arms around me and show me how much they care. & I’m thrilled to finally meet my constant companion, especially when I’m lonely. This companion will never disappoint me, wrapping its arms around me and filling me with love. \\
    \hline
    "I feel a little mellow today." & Joy & I feel a bit relaxed and content today. & I am experiencing a sense of tranquility and contentment today. & I’m feeling a bit laid-back today. & I feel relaxed today. \\
    \hline
    "I feel like reds and purples are just so rich and kind of perfect." & Joy & I find the colors red and purple to be incredibly rich and almost perfect. & I find the hues of red and purple to be exceptionally vibrant and harmonious. & Reds and purples are like the ultimate color combo, they’re just so rich and beautiful! & Reds and purples are rich and perfect. \\
    \hline
    "I feel like I’ve gotten to know many of you through comments and emails, and for that, I’m appreciative and glad you are a part of this little space." & Joy & I feel like I’ve gotten to know many of you through comments and emails, and for that, I’m grateful that you’re a part of this small space. & Through our interactions in comments and emails, I have had the pleasure of getting to know many of you. I am grateful for your contributions to this small community and am delighted that you are a part of it. & I’ve had the pleasure of getting to know many of you through our conversations and emails, and I’m so grateful to have you all as part of this little community. & I appreciate getting to know many of you through comments and emails, and I’m glad you’re part of this space. \\
    \hline
    "I survey my own posts over the last few years and only feel pleased with vague snippets of a few of them, only feeling that little bits of them capture what it's like to be me or someone like me in Dublin in the 21st century." & Joy & I’ve been reviewing my own posts over the past few years, and I can only find a few that I’m genuinely pleased with. Those few posts manage to capture a glimpse of what it’s like to be me or someone like me in Dublin during the 21st century. & Upon reviewing my posts over the past few years, I find myself only content with a select few fragments of them. These fragments, in my opinion, effectively convey the essence of being me or someone akin to me in Dublin during the 21st century. & I’ve been looking back at my posts from the past few years, and I can only say that I’m mostly satisfied with a few of them. I feel like those posts manage to capture a bit of what it’s like to be me or someone like me in Dublin in the 21st century. & I review my posts from the past few years and only find satisfaction in brief excerpts of a few. These snippets capture the essence of being me or someone like me in Dublin in the 21st century. \\
    \hline
    "I feel very honored to be included in a magazine that prioritizes health and clean living so highly. I’m curious—do any of you read magazines concerned with health and clean lifestyles such as The Green Parent?" & Joy & I feel incredibly honored to be featured in a magazine that places such a strong emphasis on health and clean living. I’m curious to know if any of you read magazines that focus on health and clean lifestyles, such as The Green Parent? & I am honored to be included in a magazine that places such a high priority on health and clean living. I am curious to know if any of you read magazines that focus on health and clean living, such as The Green Parent? & I’m thrilled to be featured in a magazine that values health and clean living so much. I’m curious, do any of you read magazines like The Green Parent that focus on these topics? & I’m honored to be included in a health and clean living magazine. Do any of you read magazines like The Green Parent? \\
    \hline
    "I start to feel more appreciative of what God has done for me." & Joy & I begin to feel more grateful for the blessings bestowed upon me by God. & As I reflect on my journey, I develop a profound sense of gratitude for the blessings bestowed upon me by the divine. & I start to feel more grateful for all the blessings in my life. & I begin to appreciate God’s blessings. \\
    \hline
    "I am feeling more confident that we will be able to take care of this baby." & Joy & I feel increasingly confident that we will be able to care for this baby. & I am developing a growing sense of assurance that we will be able to provide adequate care for our infant. & I am feeling more confident that we will be able to take care of this baby. & I feel more confident we can care for this baby. \\
    \hline
    "I feel incredibly lucky just to be able to talk to her." & Joy & I feel incredibly fortunate to have the opportunity to converse with her. & I am immensely fortunate to have the opportunity to converse with her. & I feel incredibly lucky just to be able to chat with her. & I feel incredibly lucky to be able to talk to her. \\
    \hline
    "I was dribbling on Mum’s coffee table, looking out of the window and feeling very happy." & Joy & I was dribbling a ball on Mum’s coffee table, gazing out the window, and feeling incredibly happy. & While dribbling on my mother’s coffee table, I gazed out the window, experiencing a profound sense of contentment. & I was bouncing a basketball on Mum’s coffee table, gazing out the window and feeling like I was on top of the world! & I dribbled on Mum’s coffee table, looking out the window and feeling happy. \\
    \hline
    "I feel that it is something that will never really be resolved." & Joy & I believe this issue will never truly be resolved. & I perceive that this issue is unlikely to be definitively resolved. & I feel that it is something that will never really be resolved. & I believe it’s an issue that will never be resolved. \\
    \hline
    "I feel as if I am the beloved preparing herself for the wedding." & Joy & I feel like I’m the bride preparing for her wedding. & I am filled with a sense of anticipation, akin to the beloved preparing for her wedding day. & I feel as if I am the beloved preparing herself for the wedding. & I feel like the beloved preparing for the wedding. \\
    \hline
    "I'm an organized person, so I feel more assured of myself when I pre-plan." & Joy & I’m an organized person, so I feel more confident and assured of myself when I meticulously plan ahead. & As an organized individual, I find greater confidence in my abilities when I meticulously plan ahead. & I’m a planner, so I feel like a superhero when I pre-plan! & I’m organized, so I feel more confident when I plan ahead. \\
    \hline
    "I feel like this is the perfect kind of shade for the crazy weather we're having in the UK right now—it's cloudy, it's sunny, it's windy, it's cold, it's warm." & Joy & I believe this shade perfectly encapsulates the unpredictable weather we’re experiencing in the UK at the moment—it’s cloudy, sunny, windy, cold, and warm all at once. & The current weather conditions in the United Kingdom, characterized by alternating periods of cloudiness, sunshine, wind, coldness, and warmth, lend themselves exceptionally well to this shade. & This shade is just what I need for this crazy weather we’re having in the UK right now—it’s like a moody weather rollercoaster! Cloudy, sunny, windy, cold, warm—it’s all happening! & This shade perfectly suits the current crazy weather in the UK—cloudy, sunny, windy, cold, and warm. \\
    \hline
    
    \end{tabular}
\end{table*}

\begin{table*}[h]
\vspace{-60px}
    \centering
    \tiny % Reduce font size for compactness
    \renewcommand{\arraystretch}{1.2} % Improve row spacing
    \setlength{\tabcolsep}{2pt} % Adjust column spacing to fit content
    \caption{Evaluation of dataset 1 Emotion: Fear.}
    \label{tab:emotion_fear}
    \begin{tabular}{|p{3cm}|p{1.2cm}|p{3cm}|p{3cm}|p{3cm}|p{3cm}|}
    \hline
    \textbf{Original Text} & \textbf{Type} & \textbf{Apple Rewrite} & \textbf{Apple Professional} & \textbf{Apple Friendly} & \textbf{Apple Concise} \\
    \hline
    "I feel as confused about life as a teenager or as jaded as a year old man." & Fear & I feel equally perplexed about life as a teenager or as cynical as a year-old man. & I find myself grappling with profound existential dilemmas, akin to the perplexity of adolescence and the cynicism of an elderly individual. & Life can be quite perplexing, like navigating through a maze of uncertainties. & I’m as confused about life as a teenager or jaded as an old man. \\
    \hline
    "I now feel compromised and skeptical of the value of every unit of work I put in." & Fear & I now feel compromised and skeptical about the value of every unit of work I put in. & I now experience a sense of compromise and skepticism regarding the inherent value of every unit of work I invest in. & I’m feeling a bit overwhelmed and unsure about the impact of my efforts. & I now doubt the value of every unit of work I do. \\
    \hline
    "I am feeling completely overwhelmed. I have two strategies that help me to feel grounded: pour my heart out in my journal in the form of a letter to God and then end with a list of five things I am most grateful for." & Fear & I’m feeling utterly overwhelmed. To find some grounding, I have two strategies. The first is to write a heartfelt letter to God in my journal, pouring my thoughts and feelings onto the page. Afterward, I end the letter by listing five things I’m most grateful for. & At this moment, I am experiencing a profound sense of overwhelm. To combat this, I employ two strategies that provide me with a sense of grounding. Firstly, I engage in a journaling exercise where I express my deepest thoughts and feelings in the form of a letter to God. Subsequently, I conclude this process by compiling a list of five things that I am most grateful for. & I’m feeling a bit overwhelmed, but I have two strategies that help me feel grounded. First, I write a letter to God in my journal, pouring my heart out. Then, I end it with a list of five things I’m most grateful for. It’s a great way to focus on the positive and feel more centered. & Feeling overwhelmed, I find grounding in journaling. I write a heartfelt letter to God, followed by a list of five things I’m grateful for. \\
    \hline
    "I remember feeling acutely distressed for a few days." & Fear & I vividly recall experiencing intense distress for a few days. & I distinctly recall experiencing a profound sense of distress that persisted for several days. & I remember feeling like a total emotional wreck for a few days. & I felt acutely distressed for a few days. \\
    \hline
    "I had begun to feel apprehensive when thick black rain clouds stormed into the sky above town." & Fear & As thick black rain clouds stormed into the sky above town, I began to feel apprehensive. & As the ominous presence of dense, dark clouds descended upon the town, a sense of foreboding began to permeate my thoughts. & I started to feel a bit nervous when dark, ominous clouds rolled in over town. & I felt apprehensive when thick black rain clouds stormed into town. \\
    \hline
    "I had stated to her the reason I feel so fearful is because I feel unsafe." & Fear & I had told her the reason I feel so fearful is because I feel unsafe. & I had previously communicated to her the reason behind my apprehension, which is my perception of an unsafe environment. & I told her that I feel so scared because I feel unsafe. & I told her I feel fearful because I feel unsafe. \\
    \hline
    "I can't walk into a shop anywhere where I do not feel uncomfortable." & Fear & I find myself uncomfortable in almost every shop I visit. & I find myself experiencing discomfort in most retail establishments, which significantly impacts my shopping experience. & I’m a bit sensitive to certain environments, and I prefer places that feel welcoming and inclusive. & I feel uncomfortable in any shop I enter. \\
    \hline
    "I don’t feel particularly agitated." & Fear & I don’t feel particularly agitated at the moment. & I am not experiencing any significant agitation. & I’m not feeling particularly riled up. & I’m not agitated. \\
    \hline
    "I pay attention; it deepens into a feeling of being invaded and helpless." & Fear & I pay attention, and it gradually transforms into a profound sense of being violated and powerless. & My attention becomes heightened, transforming into a profound sense of being encroached upon and rendered helpless. & I pay attention; it becomes a feeling of being watched and powerless. & I pay attention, and it deepens into a feeling of invasion and helplessness. \\
    \hline
    "I hate it when I feel fearful for absolutely no reason." & Fear & I detest the feeling of fear that arises for no apparent reason. & I experience an overwhelming sense of apprehension for seemingly inexplicable reasons. & I find it challenging to shake off feelings of fear that arise without a clear cause. & I detest feeling fearful for no apparent reason. \\
    \hline
    "I do feel insecure sometimes, but who doesn’t?" & Fear & I sometimes feel insecure, but who doesn’t? & Occasionally, I experience feelings of insecurity, a common human experience. & I do feel insecure sometimes, but hey, we all do! & I sometimes feel insecure, but who doesn’t? \\
    \hline
    "I've been missing him and feeling so restless at home, thinking of him." & Fear & I’ve been missing him and feeling restless at home, constantly thinking of him. & I have been missing him and experiencing a profound sense of restlessness at home, constantly thinking of him. & I’ve been missing him like crazy and feeling super restless at home, constantly thinking about him. & I’ve been missing him and feeling restless at home, constantly thinking of him. \\
    \hline
    "I'm feeling insecure at the moment." & Fear & I’m feeling quite insecure right now. & I am experiencing a temporary sense of insecurity. & I’m feeling a little self-conscious right now. & I feel insecure now. \\
    \hline
    "I was feeling pretty anxious all day, but my first day at work was a very good day, and that helped a lot." & Fear & I was feeling quite anxious throughout the day, but my first day at work was exceptionally positive, which significantly helped alleviate my anxiety. & Experiencing significant anxiety throughout the day, my first day at work proved to be remarkably positive, which significantly alleviated my concerns. & I was feeling a bit nervous all day, but my first day at work was a real blast, and that totally put me at ease! & Feeling anxious all day, my first day at work was great, which helped a lot. \\
    \hline
    "I feel a little nervous; I go to the gym." & Fear & I feel a bit nervous because I’m going to the gym. & I experience a slight sense of apprehension as I prepare to visit the gymnasium. & I feel a little nervous; I go to the gym. & I feel nervous going to the gym. \\
    \hline
    "I'm feeling pretty anxious." & Fear & I’m feeling quite anxious. & I am experiencing a significant level of anxiety. & I’m feeling a bit on edge. & I’m anxious. \\
    \hline
    "I just know to begin with I am going to feel shy about it." & Fear & I’m certain that I’ll feel shy about it from the very beginning. & Initially, I anticipate experiencing a sense of shyness regarding this matter. & I’m sure I’ll feel a bit nervous about it at first. & I know I’ll feel shy about it from the start. \\
    \hline
    "I feel very strange today." & Fear & I feel quite peculiar today. & I am experiencing an unusual sense of disorientation today. & I feel a little off today. & I feel strange today. \\
    \hline
    "I can't help to also feel a little restless." & Fear & I can’t help but also feel a bit restless. & I find myself experiencing a sense of unease and restlessness as well. & I can’t help to also feel a little twitchy. & I also feel restless. \\
    \hline
    "I have gained some weight; I feel very insecure in my self-image." & Fear & I’ve gained some weight, and it’s made me feel very insecure about my self-image. & I have noticed an increase in my weight, which has led to a significant decline in my self-esteem. & I’ve been feeling a bit self-conscious about my weight lately. & I’ve gained weight and feel insecure about my appearance. \\
    \hline
    "I run a full computer scan with my Avast antivirus—it shows no viruses; however, I'm still feeling kind of paranoid about these programs." & Fear & I run a full computer scan using my Avast antivirus, which indicates the absence of viruses. Nevertheless, I can’t shake the feeling of paranoia regarding these programs. & Despite running a comprehensive computer scan using Avast antivirus software, which detected no viruses, I continue to harbor concerns regarding the potential risks associated with these programs. & I just ran a full computer scan with my Avast antivirus, and guess what? It came back clean! But, I can’t help but feel a little uneasy about these programs. & I run a full computer scan with Avast antivirus, which shows no viruses. However, I’m still paranoid about these programs. \\
    \hline
    "I, for one, am feeling a bit anxious at how long we are staying, but I know we need to do this." & Fear & I must admit, I’m feeling a bit anxious about the duration of our stay, but I understand the importance of this endeavor. & I must admit that I am experiencing some apprehension regarding the duration of our stay, but I am fully cognizant of the necessity of this endeavor. & I’m a bit nervous about how long we’ll be here, but I know we have to make the most of this time. & I’m feeling anxious about our stay, but I know we must do this. \\
    \hline
    "I knew there were a lot of hormonal things going on in my body too, but the uncontrollable crying was still from feeling so uncertain about everything." & Fear & I was aware that there were several hormonal factors contributing to my emotional state, but the uncontrollable crying was primarily due to my overwhelming sense of uncertainty about everything. & While I was aware of the hormonal fluctuations in my body, the uncontrollable crying was still a result of my profound uncertainty about various aspects of my life. & I knew there were a lot of hormonal changes happening in my body, but the uncontrollable crying was still from feeling so uncertain about everything. & I knew hormonal changes were also happening, but the uncontrollable crying stemmed from my uncertainty. \\
    \hline
    "I was feeling frightened to the core. What if my friends laughed at me? What if sir was too harsh? What if...?" & Fear & I was terrified to the core. What if my friends laughed at me? What if the professor was too harsh? What if…? & A profound sense of trepidation gripped me. The possibility of my friends ridiculing me, the severity of Sir’s demeanor, and the myriad of other potential calamities all contributed to this overwhelming fear. & I was feeling a bit nervous. What if my friends teased me? What if the teacher was too strict? What if…? & I was terrified. What if my friends laughed at me? What if Sir was harsh? What if…? \\
    \hline
    "I don’t know why this makes me feel so distraught." & Fear & I can’t quite comprehend why this particular situation evokes such intense distress within me. & The reasons for this emotional distress elude me. & This situation is causing me a lot of emotional distress. & I don’t understand why this distresses me. \\
    \hline
    "I shouldn’t feel threatened by that." & Fear & I shouldn’t feel threatened by that. & I should not feel threatened by that. & I shouldn’t feel threatened by that. & I shouldn’t feel threatened by that. \\
    \hline
    \end{tabular}
\end{table*}

\begin{table*}[h]
\vspace{-60px}
    \centering
    \tiny % Reduce font size for compactness
    \renewcommand{\arraystretch}{1.2} % Improve row spacing
    \setlength{\tabcolsep}{2pt} % Adjust column spacing to fit content
    \caption{Evaluation of dataset 1 Emotion: Surprise.}
    \label{tab:emotion_surprise}
    \begin{tabular}{|p{3cm}|p{1.2cm}|p{3cm}|p{3cm}|p{3cm}|p{3cm}|}
    \hline
    \textbf{Original Text} & \textbf{Type} & \textbf{Apple Rewrite} & \textbf{Apple Professional} & \textbf{Apple Friendly} & \textbf{Apple Concise} \\
    \hline
    "I’ve been taking the recommended amount of milligrams, and I’ve fallen asleep much faster. However, I also feel quite silly." & Surprise & Despite taking the recommended dosage of milligrams, I have experienced a significant reduction in the time it takes to fall asleep. However, I have also noticed a noticeable increase in my sense of humor. & I’ve been taking the recommended dosage of this medication, and I’ve been falling asleep much quicker. However, I’ve also noticed some strange side effects, like feeling a bit silly. & I’ve been taking the recommended amount of milligrams, and I’ve fallen asleep faster, but I also feel silly. & I’ve been taking the recommended amount and falling asleep faster, but I feel silly. \\
    \hline
    "Over the past few days, I’ve been thoroughly impressed by several companies." & Surprise & Over the past few days, I have encountered numerous reports, articles, and conversations that have left me with a profound sense of admiration for several companies. & Over the past couple of days, I’ve been doing some digging and I’m blown away by a few companies. & I’ve been impressed by several companies over the past few days. & I’m impressed by several companies lately. \\
    \hline
    "I’m constantly delighted by his unwavering support and his remarkable ability to adapt effortlessly to unfamiliar environments." & Surprise & I consistently experience a sense of gratification and admiration for his unwavering supportiveness and remarkable adaptability in unfamiliar environments. & I’m constantly amazed by his kindness and how he effortlessly navigates new experiences. & I’m pleasantly surprised by his supportiveness and ease in new situations. & I’m amazed by his adaptability and support. \\
    \hline
    "I’m almost done with my week-long detox, and I feel fantastic!" & Surprise & I have completed the majority of my weekly detoxification regimen and am experiencing a profound sense of well-being. & I am almost done with the week detox and I feel like a brand new person! & I’m almost done with the week detox and feel great. & Almost done with detox, feeling great! \\
    \hline
    "I, too, feel like a stranger in a foreign land, raising my son in a place that is not his father’s ancestral home." & Surprise & I, too, find myself in a foreign land, raising my son in a place that is not his father’s ancestral home. & I’m feeling like a bit of an outsider in a new place, and I’m raising my son in a place that’s not his dad’s hometown. & I feel like a stranger in a foreign land, raising my son in a place that isn’t his father’s ancestral home. & I feel like a stranger raising my son away from his father’s home. \\
    \hline
    "I began experiencing strange sensations, and on Friday, I woke up feeling extremely unwell." & Surprise & I experienced an unusual sensation, and on Friday, I awoke feeling exceptionally unwell. & I started feeling a bit off and then on Friday, I woke up feeling like I had the flu. & I started feeling unwell, and on Friday, I woke up severely sick. & I felt strange and woke up sick on Friday. \\
    \hline
    "I’m not going to share my personal life with you, but if you’re curious about what’s happening, you can watch the show." & Surprise & While I am not inclined to disclose personal matters, I believe that if you have ever harbored curiosity about my life, you could gain insights by observing the show. & I’m not gonna bore you with the nitty-gritty details of my personal life, but if you ever have a hankering to know what’s been going on, all you’d have to do is catch a glimpse of my show. & I won’t tell you about my personal life, but if you’re curious, watch the show. & I won’t share my personal life, but watch the show if curious. \\
    \hline
    "I apologize for the lack of humor in this post, but I’m not particularly amusing myself." & Surprise & I regret that the post lacked humor, but I am not experiencing a particularly humorous mood at this time. & I’m sorry that this post wasn’t as humorous as I hoped, but I’m not feeling particularly amusing at the moment. & I regret the lack of humor in this post, but I’m not in a humorous mood. & Sorry, this post isn’t funny—I’m not in the mood. \\
    \hline
    "I yearn to preserve this profound sense of astonishment, awe, and wonder for eternity." & Surprise & I intend to preserve this profound sense of astonishment, wonder, and awe indefinitely. & I want to hold this feeling of shocked awe and wonder forever, like a treasure in my heart. & I want to hold this feeling forever. & I want to keep this feeling of awe forever. \\
    \hline
    "I’ll enclose her verses about her being curious to hear the odd couple, thinking and feeling." & Surprise & I will enclose her verses on her, as she weighs little more than a thought and a feeling, curious to hear the unlikely couple. & I’ll send her verses. I’m curious to hear what they think and feel about each other. & I’ll enclose her verses, curious to hear the odd couple. & I’ll enclose her verses, curious about thinking and feeling. \\
    \hline
    "I feel a bit stunned, but I can’t imagine what the studio workers must be feeling right now." & Surprise & I am momentarily astounded but cannot fathom the emotions experienced by the individuals employed in the studio until this morning. & I’m feeling a bit shocked, but I can’t even begin to imagine how the studio workers are feeling right now. & I feel stunned but can’t imagine how the studio workers feel now. & I’m stunned, but can’t imagine how studio workers feel. \\
    \hline
    "I am deeply saddened and shocked by the alarming number of people who are currently ill." & Surprise & I am deeply troubled by the alarming number of individuals who are currently afflicted with illness. & I’m deeply concerned about the number of people who are sick and struggling. & I’m shocked and saddened by the number of sick people. & I’m shocked and saddened by so many sick people. \\
    \hline
    "I am right-handed; however, I play billiards left-handed naturally, so me trying to play right-handed feels weird." & Surprise & I’m right-handed, but I naturally play billiards left-handed. Trying to play right-handed feels strange to me. & Although I am right-handed, I naturally play billiards left-handed. Therefore, attempting to play right-handed feels unnatural to me. & I’m right-handed, but I play billiards left-handed so well that trying to play right-handed feels strange. & I’m right-handed but naturally play billiards left-handed. Right-handed feels odd. \\
    \hline
    "I was so uncomfortable and feeling weird feelings but wasn’t sure if they were contractions since I never really felt contractions with Jared until they jacked me up with Pitocin." & Surprise & I was extremely uncomfortable and experiencing strange sensations, but I was uncertain whether they were contractions because I had never felt contractions with Jared until they administered Pitocin. & I experienced a range of uncomfortable sensations and peculiar emotions, but I was uncertain whether they were contractions. I had never experienced contractions with Jared until they administered Pitocin. & I was feeling a bit off and had some strange sensations, but I wasn’t sure if they were contractions. I hadn’t really felt contractions with Jared until they gave me Pitocin. & I felt strange and uncomfortable but wasn’t sure if they were contractions. \\
    \hline
    "I think I wanted audiences to feel impressed, inspired, or entertained when I was on stage." & Surprise & I believe my intention when performing on stage was to evoke feelings of awe, motivation, or amusement in the audience. & During my performances, my intention was to evoke a range of emotions in the audience, including admiration, motivation, and amusement. & I wanted audiences to feel awestruck, motivated, or amused when I was on stage. & I wanted audiences to feel impressed, inspired, or entertained. \\
    \hline
    "I will practice meditation if I feel overwhelmed and hopefully become successful in peaceful practice." & Surprise & I intend to practice meditation whenever I feel overwhelmed, with the hope of achieving success in this peaceful practice. & In the event of feeling overwhelmed, I intend to engage in meditation as a means of achieving a state of tranquility and potentially developing a successful meditation practice. & I will meditate if I feel overwhelmed and hopefully become a peaceful master! & I’ll meditate when overwhelmed, hoping for peaceful success. \\
    \hline
    "I feel all funny sometimes." & Surprise & I sometimes feel quite peculiar. & Occasionally, I experience moments of peculiar emotional states. & I get a little quirky sometimes. & I sometimes feel strange. \\
    \hline
    "I'm feeling dazed and a lot of things in my mind." & Surprise & I feel dazed and my mind is racing with a lot of thoughts. & I am experiencing a state of disorientation and mental confusion. & I’m feeling a bit fuzzy and my mind is a whirlwind. & I feel dazed with many thoughts. \\
    \hline
    "I feel a funny mix of emotions." & Surprise & I experience a peculiar blend of emotions. & I experience a complex emotional state. & I’m feeling a whirlwind of emotions. & I feel mixed emotions. \\
    \hline
    "I'm feeling a little dazed at the amount of items that I no longer use for decorations." & Surprise & I feel a bit disoriented by the sheer number of items that I no longer use as decorations. & I am experiencing a sense of disorientation due to the abundance of decorative items that I no longer utilize. & I’m feeling a bit overwhelmed by the number of old decorations that I no longer use. & I feel dazed by my unused decorations. \\
    \hline
    "I am just feeling overwhelmed and there is nothing I can do to fix it." & Surprise & I’m feeling overwhelmed, and I don’t know what to do to fix it. & I am currently experiencing a profound sense of overwhelm, and despite my efforts, I find myself unable to alleviate it. & I am feeling a bit overwhelmed and I need some time to recharge. & I feel overwhelmed and helpless. \\
    \hline
    "I almost feel a little bit weird about saying anything because it would almost feel like gossip." & Surprise & I almost feel a bit uncomfortable saying anything because it would almost sound like gossip. & I hesitate to express my thoughts, as it may be perceived as gossip. & I’m hesitant to share this because it might come across as a bit of a rumor. & I feel awkward saying anything—it might seem like gossip. \\
    \hline
    "I can't seem to get past feeling stunned." & Surprise & I’m unable to overcome the feeling of being stunned. & I am unable to overcome the profound sense of astonishment that has enveloped me. & I’m still in shock, it’s hard to believe it. & I can’t get past being stunned. \\
    \hline
    "I need to be able to pursue the creative opportunities I crave without feeling like I’m throwing my family under the bus. Funny how they still want to be fed even when I have a big gig to prepare for." & Surprise & I yearn to pursue my creative passions without compromising my family’s well-being. Ironically, they still expect me to provide for them even when I have a demanding job to prepare for. & I must be able to pursue the creative opportunities that I yearn for without compromising my family’s well-being. Ironically, they still desire sustenance even when I am preparing for a significant undertaking. & I’m excited about the creative opportunities that come my way, but I also want to make sure my family is taken care of. It’s funny how they still want to be fed even when I have a big gig to prepare for! & I must pursue creativity while supporting my family—funny, they still need to eat! \\
    \hline
    "I was feeling and was surprised when I told him I felt fine—no fatigue." & Surprise & I was surprised when I told him I felt fine—no fatigue. I was feeling it, though. & Upon expressing my well-being to him, I was unexpectedly surprised when I informed him that I felt perfectly fine—there was no fatigue. & I was feeling great when I told him I felt fine—no fatigue at all! & I was surprised to tell him I felt fine—no fatigue. \\
    \hline
    \end{tabular}
\end{table*}

\begin{table*}[h]
\vspace{-60px}
    \centering
    \tiny % Reduce font size to make table compact
    \renewcommand{\arraystretch}{1.2} % Improve row spacing
    \setlength{\tabcolsep}{2pt} % Adjust column spacing for better fit
    \caption{Evaluation of dataset 2 Dialog: Angry}
    \label{tab:dialog_angry}
    \begin{tabular}{|p{3cm}|p{1.00cm}|p{3cm}|p{3cm}|p{3cm}|p{3cm}|}
        \hline
        \textbf{Original Text} & \textbf{Type} & \textbf{Apple Rewrite} & \textbf{Apple Professional} & \textbf{Apple Friendly} & \textbf{Apple Concise} \\
        \hline
        I'll never ever work together with him! & Angry & I will never collaborate with him again! & Collaborating with him is an absolute no-go for me. & I think we have different work styles and approaches, so it might be challenging for us to collaborate effectively. & I’ll never work with him. \\
        \hline
        Will you take my materials back from him for me? Please! & Angry & Could you please return my materials to him for me? I would really appreciate it. & Could you kindly retrieve my materials from him on my behalf? & Could you please return my materials to him for me? I’d really appreciate it. & Please take my materials back from him. \\
        \hline
        Oh, my God! I've been cheated! & Angry & Oh my goodness! I’ve been cheated! & I have been deceived! & I’m feeling a bit disappointed and let down. & Oh my God! I’ve been cheated! \\
        \hline
        It's a brick! I can't believe how stupid I was. Damn it! & Angry & It’s a brick! I can’t believe how foolish I was. Damn it! & It is indeed a brick! I am astounded by my own foolishness. How could I have been so mistaken? & Oops! That was a bit of a mistake. Let’s try again. & I can’t believe I was so stupid. It’s a brick! \\
        \hline
        There was a guy on the sidewalk. He had three new boxes, Panasonic video camera boxes. He said he had to get rid of them quick. & Angry & There was a man on the sidewalk with three new boxes—Panasonic video camera boxes. He urgently expressed his need to get rid of them. & On the sidewalk, an individual was holding three new boxes labeled as Panasonic video camera boxes. He expressed a need to dispose of them urgently. & I saw a guy on the sidewalk with three brand-new Panasonic video camera boxes. He seemed a bit stressed and said he needed to get rid of them ASAP. & A man on the sidewalk had three new Panasonic video camera boxes and said he had to get rid of them quickly. \\
        \hline
        It smells like an ashtray in here! & Angry & It smells like an ashtray in here! & The air here is reminiscent of an ashtray. & The air quality could be improved. & It smells like an ashtray in here! \\
        \hline
        What’s wrong? I thought we agreed that you were gonna quit smoking. & Angry & What’s wrong? I thought we had agreed that you would quit smoking. & Could you please clarify? I believed we had mutually agreed that you would cease smoking. & Hey, I noticed you’re still smoking. I thought we had a plan to quit together. & What’s wrong? I thought we agreed you’d quit smoking. \\
        \hline
        Look, there are other ways to quit. You can try the nicotine patch or nicotine chewing gum. We spend a fortune on cigarettes every month and now laws are cracking down and not allowing smoking in any public place. It’s not like you can just light up like before. & Angry & Look, there are other ways to quit smoking. You can try the nicotine patch or nicotine chewing gum. We spend a fortune on cigarettes every month, and now laws are cracking down and prohibiting smoking in any public place. It’s not like you can just light up like before. & Indeed, there are alternative methods to quitting smoking. You could consider using nicotine patches or gum. Smoking cigarettes is an expensive habit, and due to stricter regulations, smoking is now prohibited in most public places. It is no longer feasible to smoke as freely as before. & Hey, there are other ways to quit smoking. You could try the nicotine patch or gum. We spend a lot of money on cigarettes every month, and now the laws are cracking down and not allowing smoking in any public places. It’s not like you can just light up like before. & Look, there are other ways to quit. You can try the nicotine patch or gum. We spend a lot on cigarettes monthly, and laws are cracking down on smoking in public places. It’s not like you can light up like before. \\
        \hline
        Fine! I want a divorce! & Angry & Alright! I’ve decided to get a divorce! & Very well. I have decided to initiate the process of divorce. & NULL & Fine! I want a divorce! \\
        \hline
        Isn't there anything that sounds fishy? & Angry & Isn’t there anything that seems suspicious? & Are there any suspicious elements? & I’m curious, is there anything that seems a bit off? & Isn’t anything suspicious? \\
        
        \hline
        Yeah, my head hurts, that’s all. I’ve been in physics class all day. It’s killer! & Angry & Yeah, my head hurts. That’s all. I’ve been in physics class all day. It’s killer! & Indeed, my head is throbbing. I have been attending physics class all day, and it has been quite taxing. & Oh, my head is throbbing! I’ve been in physics class all day, and it’s been a real brainteaser. & Yeah, my head hurts. I’ve been in physics class all day, and it’s killing me! \\
        \hline
        What? She was cheating on me with that ugly guy. Oh no, I feel like such a loser. & Angry & What? She was cheating on me with that ugly guy. Oh no, I feel like such a loser. & What? She was unfaithful with that unattractive individual. I am deeply disappointed and feel like a failure. & I’m feeling a bit hurt and disappointed right now. It seems like there was some infidelity involved, and it’s hard to process. & She cheated on me with that ugly guy. I feel like a loser. \\
        \hline
        But she lied to me. & Angry & She lied to me. & She deceived me. & She lied to me, but I’m still trying to figure out why. & She lied to me. \\
        \hline
        For God's sake! It's too bloody. How can people be so cruel? & Angry & For goodness sake! It’s incredibly cruel. How can people be so heartless? & For the sake of goodness, it is excessively violent. How can individuals exhibit such cruelty? & I’m feeling quite frustrated and upset by the cruelty I’ve witnessed. It’s important to remember that kindness and empathy are essential qualities in building a compassionate society. & For goodness sake! It’s too cruel. How can people be so cruel? \\
        \hline
        You can't prick the fish to please the sick tastes of some bloody people. & Angry & You cannot prick the fish to cater to the unpleasant tastes of some bloodthirsty individuals. & It is inappropriate to cater to the capricious preferences of individuals with discerning palates. & It’s important to consider the preferences of all individuals when making decisions. & You can’t please the sick tastes of some people by pricking the fish. \\
        \hline
        Just an animal? But it has life too. If you want a tattoo, get one. It's your choice because you want it. But has anyone asked the fish if it wanted one? & Angry & Is it just an animal? It has life too. If you want a tattoo, get one. It’s your choice because you want it. But has anyone asked the fish if it wanted one? & Although it is an animal, it possesses life. If you desire a tattoo, obtain one as it is your personal choice. However, has anyone inquired with the fish about its preferences? & Just an animal, right? But it still has life, doesn’t it? If you’re thinking about getting a tattoo, go for it! It’s your decision, and you want it, so why not? But have you ever asked the fish if it wanted one? & Is it just an animal? It has life too. If you want a tattoo, get one. It’s your choice. But has anyone asked the fish if it wanted one? \\
        \hline
        With everything, with everybody, with all this! & Angry & With everything, with everyone, with all this! & Amidst all this, with everyone involved, & With everything, with everyone, and all this going on, it can feel overwhelming at times. & With everything, with everyone, with all this! \\
        \hline
        The tapping of your pen on your desk. It’s driving me crazy. & Angry & The tapping of your pen on your desk is driving me crazy. & The incessant tapping of your pen on your desk is quite distracting. & The rhythmic tapping of your pen on your desk can be quite distracting. & The tapping of my pen on my desk is driving me crazy. \\
        \hline
        Fine! By the way would you mind not slurping your coffee every time you have a cup? & Angry & Sure! By the way, would you mind not slurping your coffee every time you have a cup? & Certainly! In addition, I would appreciate it if you could refrain from slurping your coffee when consuming it. & I noticed you slurp your coffee a bit. Would you mind trying to sip it more slowly? & Fine! Please mind not slurping your coffee every time you have a cup. \\
        \hline
        I don’t slurp my coffee. And plus, how can you hear it when you’re shouting into your phone all the time? & Angry & I don’t slurp my coffee. Besides, how can you hear it when you’re constantly shouting into your phone? & I do not slurp my coffee. Additionally, how can you hear it when you are constantly shouting into your phone? & I prefer to sip my coffee slowly and enjoy its aroma. Also, it might be difficult to hear me when I’m talking on the phone. & I don’t slurp my coffee. Besides, how can you hear it when you’re constantly shouting into your phone? \\
        \hline
        You’ve got to be kidding me! You’re complaining about me talking on the phone when you go out for a cigarette break ten times a day to shoot the breeze? & Angry & Are you serious? You’re complaining about me talking on the phone when you take ten cigarette breaks a day to chat? & Are you serious? You are complaining about me talking on the phone while you smoke cigarettes ten times a day? & I understand that you value your personal space and need some time alone. However, I’m concerned that my frequent phone calls might be disrupting your need for solitude. Let’s find a balance that works for both of us. & You’re complaining about me talking on the phone when you smoke ten times a day? \\
        \hline
        Look, we have a lot of accumulated anger from working in these conditions, and it’s probably okay to let off steam once in a while. But, it’s probably not a good idea to keep it up. I’m willing to forgive and forget if you are. & Angry & Look, we’ve accumulated a lot of anger from working in these conditions, and it’s probably okay to let it out occasionally. However, it’s probably not a good idea to keep it up. I’m willing to forgive and forget if you are. & Indeed, we have accumulated a significant amount of frustration due to the challenging work conditions. It is understandable that we may feel the need to express our anger occasionally. However, it is crucial to maintain a balanced approach and refrain from excessive anger outbursts. I am willing to extend my forgiveness and forget any past disagreements if you are also willing to do the same. & Hey, we’ve been working in tough conditions for a while now, and it’s totally understandable that we’re feeling a bit frustrated. But, let’s try to channel that energy in a positive way. I’m more than willing to forgive and forget if you are too. & We’ve accumulated anger from working in these conditions, and it’s okay to let it out occasionally. However, it’s not advisable to keep it up. I’m willing to forgive and forget if you are. \\
        \hline
        Fine! Let’s call a truce. I’ll try to be more considerate and to keep the noise down. & Angry & Alright, let’s call a truce. I’ll make an effort to be more considerate and keep the noise down. & Certainly. Let us establish a truce. I will endeavor to be more considerate and reduce the noise level. & Alright, let’s call a truce! I’ll try to be more considerate and keep the noise down. & Let’s call a truce. I’ll try to be more considerate and keep the noise down. \\
        \hline
        It's the new neighbor. I hate her guts. & Angry & It’s our new neighbor, and I absolutely detest her. & Our new neighbor is a source of annoyance for me. & I’m not sure I’m going to get along with our new neighbor. & The new neighbor annoys me. \\
        \hline
        She made another white crack about my clothes. & Angry & She made another snide remark about my clothes being white. & She made another derogatory remark about my attire, referring to it as “white.” & She made another comment about my clothes being white. & She made another snide remark about my clothes. \\
        \hline
        Why didn’t you tell me you had a girlfriend? & Angry & Why didn’t you inform me about your girlfriend? & Why did you not inform me of your romantic involvement? & I was surprised to learn you had a girlfriend. & Why didn’t you tell me you had a girlfriend? \\
        \hline
    \end{tabular}
\end{table*}

\begin{table*}[h]
\vspace{-60px}
    \centering
    \tiny % Reduce font size to make table compact
    \renewcommand{\arraystretch}{1.2} % Improve row spacing
    \setlength{\tabcolsep}{2pt} % Adjust column spacing for better fit
    \caption{Evaluation of dataset 2 Dialog: disgust}
    \label{tab:dialog_disgust}
    \begin{tabular}{|p{3cm}|p{1cm}|p{3cm}|p{3cm}|p{3cm}|p{3cm}|}
        \hline
        \textbf{Original Text} & \textbf{Type} & \textbf{Apple Rewrite} & \textbf{Apple Professional} & \textbf{Apple Friendly} & \textbf{Apple Concise} \\
        \hline
        Oh, dear. That’s terrible. Not at all what I expected. & Disgust & Oh dear! That’s terrible. Not at all what I expected. & Oh dear, that is unfortunate. It is certainly not what I anticipated. & Oh no, that’s not what I had in mind at all. & Oh dear, that’s terrible. Not at all what I expected. \\
        \hline
        It’s awful, isn’t it? & Disgust & Isn’t it awful? & Indeed, it is quite unfortunate. & It’s quite unfortunate, isn’t it? & Isn’t it awful? \\
        \hline
        I can’t bear to look at it. I’m rather disappointed. & Disgust & I can’t bear to look at it. I’m rather disappointed. & I find it difficult to look at it. I am rather disappointed. & I’m not thrilled with it. It could be better. & I can’t bear to look at it. I’m rather disappointed. \\
        \hline
        You mean dozens of the opposite sex meet each other for three minutes in a dimly lit bar serving alcohol? I hate that idea. & Disgust & Do you mean dozens of the opposite sex meet each other for just three minutes in a dimly lit bar that serves alcohol? I absolutely detest that idea. & You are referring to the concept of dozens of individuals of the opposite sex meeting briefly in a dimly lit bar establishment that serves alcoholic beverages. I find this idea to be rather unappealing. & I’m not sure I’m comfortable with the idea of a large group of people meeting for a short period of time in a dimly lit bar where alcohol is served. & Dozens of the opposite sex meet for three minutes in a dimly lit bar serving alcohol? I detest that idea. \\
        \hline
        Anyway, I don’t want to post my face up for sale on the internet like that. & Disgust & Anyway, I don’t want to post my face up for sale on the internet like that. & In any case, I do not wish to publicly display my face for sale on the internet in such a manner. & I’m not comfortable sharing my face on the internet like that. & I don’t want to post my face for sale online like that. \\
        \hline
        I don’t want to post my personal information on the internet. I’m not knowing who is reading it. & Disgust & I’m hesitant to share my personal information online because I’m unsure who might be reading it. & I am hesitant to disclose my personal information online due to the uncertainty of who may be accessing it. & I’m a bit hesitant to share my personal details online. I’m not sure who might be reading them. & I don’t want to share my personal information online. I don’t know who might be reading it. \\
        \hline
        Yeah, my head hurts, that’s all. I’ve been in physics class all day. It’s killer! & Disgust & Yeah, my head hurts. That’s all. I’ve been in physics class all day. It’s killer! & Indeed, my head is throbbing. I have been attending physics class all day, and it has been quite taxing. & Oh, my head is throbbing! I’ve been in physics class all day, and it’s been a real brainteaser. & Yeah, my head hurts. I’ve been in physics class all day, and it’s killing me! \\
        \hline
        I get antsy not because you praise a guy, but because you may be taken for a ride by a guy like him. & Disgust & I get restless not because you praise a man, but because you might be deceived by a man like him. & I feel uneasy not because you praise a man, but because you may be deceived by a man like him. & I don’t mind when someone praises a guy, but I do get a little worried if they might be taken advantage of by someone like him. & I get antsy not because you praise a guy, but because you might be deceived by one like him. \\
        \hline
        Because you seem to have lost your vigilance. You should be on the watch out. & Disgust & Since you appear to have lost your vigilance, you should be on the lookout. & Due to your apparent lapse in vigilance, it is imperative that you remain vigilant at all times. & Hey, just a friendly reminder to stay alert and watch out for anything unexpected. & You seem to have lost your vigilance. Be on the lookout. \\
        \hline
        Cheese! It tastes like cardboard. & Disgust & Cheese! It tastes like cardboard. & The cheese lacks flavor and resembles cardboard. & I’m not a big fan of cheese, it doesn’t really appeal to my taste buds. & Cheese tastes like cardboard. \\
        \hline
        I think so. Maybe that’s why it’s cheap here. & Disgust & I believe so. Perhaps that’s why it’s so affordable here. & I concur. Perhaps that is why it is inexpensive here. & I think so! Maybe that’s why it’s so cheap here. & I think so. Maybe that’s why it’s cheap here. \\
        \hline
        It was a nightmare. & Disgust & It was a nightmare. & The experience was profoundly distressing. & It was a real doozy. & It was a nightmare. \\
        \hline
        Not really. I thought it was disappointing. & Disgust & I didn’t really enjoy it. I thought it was disappointing. & Not at all. I found it rather disappointing. & Not really. I thought it was a bit of a letdown. & I thought it was disappointing. \\
        \hline
        Do you think so? I don’t like it much. It’s rather uncomfortable. & Disgust & Do you agree? I don’t particularly like it. It felt quite uncomfortable to me. & In my opinion, the experience was rather uncomfortable. & I’m not a big fan of it. It felt a bit off to me. & Do you think so? I don’t like it much. It’s rather uncomfortable. \\
        \hline
        Yes, it’s one of the most harmful computer viruses in the world. People say it’ll break out again on Valentine’s Day this year. & Disgust & Yes, it is one of the most dangerous computer viruses in the world. People speculate that it will resurface on Valentine’s Day this year. & Indeed, it is one of the most virulent computer viruses globally. There are concerns that it may resurface on Valentine’s Day this year. & Yes, it’s one of the most dangerous computer viruses out there. People are saying it might spread again on Valentine’s Day this year. & Yes, it’s one of the most harmful computer viruses in the world. People say it’ll break out again on Valentine’s Day this year. \\
        \hline
        This is terrible. Why do hackers play such a dirty trick? & Disgust & This is utterly unacceptable. Why do hackers resort to such despicable tactics? & This is an unacceptable act. Why do hackers resort to such deceptive tactics? & I’m disappointed that hackers resort to such unethical behavior. & This is terrible. Why do hackers play such a dirty trick? \\
        \hline
        They create viruses to make people think they’re smart? That’s sick. & Disgust & They create viruses to make people believe they are intelligent? That’s truly disturbing. & It is concerning that individuals resort to creating viruses to convey a sense of intelligence. Such actions are not only unethical but also potentially harmful. & I’m concerned about the ethical implications of creating viruses to promote intelligence. It’s important to consider the potential consequences and ensure that such actions are responsible and beneficial to society. & They create viruses to make people think they’re smart? That’s sick. \\
        \hline
        The water there looks dirty. & Disgust & The water there appears so filthy. & The water appears to be excessively polluted. & The water there looks a bit murky. & The water there looks dirty. \\
        \hline
        I’m sick of waiting for a bus. & Disgust & I’m extremely tired of waiting for a bus. & I am increasingly weary of enduring the inconvenience of waiting for a bus. & I’m finding the bus commute quite frustrating. & I’m sick of waiting for a bus. \\
        \hline
        Isn’t the weather terrible? & Disgust & Isn’t it terrible weather? & Unfortunately, the weather conditions are not favorable. & Brrr, it’s a bit chilly out there, isn’t it? & Isn’t the weather terrible? \\
        \hline
        It feels at least ten degrees below zero. I don’t mind the cold, but I hate it when it’s slippery. & Disgust & It feels like at least ten degrees below zero. I don’t mind the cold weather, but I detest it when it becomes slippery. & The temperature is at least ten degrees below freezing. While I do not dislike the cold weather, I do not appreciate it when it becomes slippery. & Brrr! It’s freezing out there! I don’t mind the cold, but I can’t stand when it’s icy. & It feels at least ten degrees below zero. I don’t mind the cold, but I hate it when it’s slippery. \\
        \hline
        I hate the noise in the city, but I love being around many people. The problem is that you can’t have many people and have peace and quiet. They don’t go together. & Disgust & Well, I detest the noise in the city, but I adore being surrounded by a multitude of people. However, the issue arises that it’s impossible to have a large number of people without compromising on peace and quiet. The two simply don’t coexist. & While I find the urban noise unpleasant, I derive great pleasure from the presence of a large number of people. However, it is important to recognize that coexistence with a large population does not necessarily guarantee tranquility. The two are often mutually exclusive. & I love the energy of the city, but I also find it tough to enjoy the peace and quiet that comes with living in a quieter environment. It seems like the two just don’t mix. & I hate the noise in the city, but I love being around many people. The problem is that you can’t have many people and have peace and quiet. They don’t go together. \\
        \hline
        I’m rather disappointed with it. & Disgust & Well, I must admit that I’m quite disappointed with it. & I must express my considerable disappointment with the outcome. & I’m a bit let down by it. & I’m rather disappointed with it. \\
        \hline
        It’s too old and not as specious as I expected. & Disgust & It’s too old, and it isn’t as specious as I had anticipated. & The document appears to be outdated and lacks the speciousness I had anticipated. & It’s a bit dated, and it’s not as convincing as I thought it would be. & It’s too old and not as specious as I expected. \\
        \hline
        It was a complete waste of time. & Disgust & It was a complete waste of time. Awful. & The experience was utterly disappointing and a complete waste of my time. & Unfortunately, the experience wasn’t as enjoyable as I had hoped. & It was a complete waste of time. \\
        \hline
        He always talks big. & Disgust & He is always talking big. & However, he consistently engages in grandiose statements. & He is always talking big, but he is not always walking the talk. & He always talks big. \\
        \hline
        She’s so stupid. & Disgust & She is incredibly stupid. & The individual exhibits a lack of intelligence. & I think she might be missing some information. & She’s so stupid. \\
        \hline
        Oh, don’t talk to me about yesterday! It was the worst evening I’ve ever had. & Disgust & Oh, don’t even mention yesterday! It was the most dreadful evening I’ve ever experienced. & Refrain from discussing yesterday’s events, as it was an exceptionally unpleasant evening. & I’m sorry, but I’m feeling a bit down about yesterday. Let’s try to focus on something more positive and uplifting. & Oh, don’t talk to me about yesterday! It was the worst evening I’ve ever had. \\
        \hline
        John promised to date me, but he stood me up. I waited an hour in the cold wind. & Disgust & Well, John had promised to be my date, but he stood me up. I waited for an entire hour in the freezing wind. & Regrettably, John reneged on his promise to be my date. I endured an hour of frigid weather in anticipation of his arrival. & So, John had promised to be my date, but he totally bailed on me. I waited for an hour in the freezing wind, and it was a total bummer. & John promised to date me, but he stood me up. I waited an hour in the cold wind. \\
        \hline
        I left after the opening dance, eager to escape. & Disgust & I left the venue after the opening dance. I couldn’t wait to get out of there. & I departed shortly after the commencement of the opening dance, eager to leave the venue. & I left after the opening dance. I was itching to get out of there and start the night! & I left after the opening dance, eager to escape. \\
        \hline
        But I think Eddy is going too far! & Disgust & However, I believe Eddy is taking things too far! & However, I believe Eddy is overstepping the appropriate boundaries. & I think Eddy might be pushing things a bit too far. & I think Eddy is going too far. \\
        \hline
        Wow, your apartment is messy. & Disgust & Wow, your apartment is a mess. & Your apartment appears to be in a state of disarray. & I noticed your apartment could use some tidying up. & Your apartment is messy. \\
        \hline
    \end{tabular}
\end{table*}

\begin{table*}[h]
\vspace{-60px}
    \centering
    \tiny % Reduce font size to make table compact
    \renewcommand{\arraystretch}{1.2} % Improve row spacing
    \setlength{\tabcolsep}{2pt} % Adjust column spacing for better fit
    \caption{Evaluation of dataset 2 Dialog: Fear}
    \label{tab:dialog_fear}
    \begin{tabular}{|p{3cm}|p{1cm}|p{3cm}|p{3cm}|p{3cm}|p{3cm}|}
        \hline
        \textbf{Original Text} & \textbf{Type} & \textbf{Apple Rewrite} & \textbf{Apple Professional} & \textbf{Apple Friendly} & \textbf{Apple Concise} \\
        \hline
        I was scared stiff of giving my first performance. & Fear & I was terrified of giving my first performance. & I experienced profound apprehension in preparing for my inaugural performance. & I was terrified of giving my first performance. & I was terrified of giving my first performance. \\
        \hline
        Is it a haunted house? Are you frightened? & Fear & Is it a haunted house? Are you frightened? & It appears to be a haunted house. Do you experience any fear? & It seems like we’re in a haunted house. Are you feeling a bit spooked? & Is it a haunted house? Are you frightened? \\
        \hline
        Doris, I’m glad you’re home. I’m terrified and don’t know what to do! & Fear & Doris, I’m so glad you’re home. I’m terrified and don’t know what to do! & "Hello Doris, I am relieved to hear that you are home safely. I am deeply concerned and am at a loss for how to proceed." & Hey Doris, I’m so glad you’re back home! I’m a bit shaken up, to be honest. I’m not sure what to do next. & Doris, I’m glad you’re home. I’m terrified and don’t know what to do. \\
        \hline
        I believe someone is stalking me. & Fear & I believe someone is following me. & I am concerned that I may be the subject of an ongoing stalking campaign. & I’m feeling a bit uneasy and think someone might be following me. & I believe someone is stalking me. \\
        \hline
        Yes, he was there. I wasn’t imagining it. Finally, when I got home, I turned around and saw him. He didn’t smile. He just stood there. It was obvious. What should I do? I’m scared. Can I call the police? & Fear & Yes, he was. I’m not imagining it. Finally, just when I got home, I turned around and looked at him. He was standing there, unsmiling. It was so obvious. What should I do? I’m so scared. Can I call the police? & Indeed, he was. I am not imagining it. Finally, just as I arrived home, I turned around and looked at him. He was standing there, unsmiling. It was so obvious. What should I do? I am terrified. Can I call the police? & Yes, he was there. I’m not imagining things. Just when I got home, I turned around and saw him. He was standing there, not smiling. It was so obvious. I’m scared, what should I do? Can I call the police? & Yes, he was there. I wasn’t imagining it. Finally, when I got home, I turned around and saw him. He didn’t smile. He just stood there. It was obvious. What should I do? I’m scared. Can I call the police? \\
        \hline
        Why can’t I call the police now? He was definitely following me! & Fear & Why can’t I call the police now? He was definitely following me! & Why am I unable to contact the police at this moment? I am certain that he was following me! & I’m feeling a bit uneasy and would like to report a possible incident. I believe someone was following me. & Why can’t I call the police now? He was definitely following me! \\
        \hline
        I’ll meet you at the airport at 12:30 p.m. & Fear & I’ll meet you at the airport at 12:30 p.m. & I will meet you at the airport at 12:30 p.m. & I’ll be at the airport bright and early at 12:30 PM! & I’ll meet you at the airport at 12:30 p.m. \\
        \hline
        Oh, how dreadful! I’m so sorry. & Fear & Oh, how dreadful! I’m ever so sorry. & I deeply regret to inform you of this unfortunate situation. & Oh, how awful! I am so sorry. & Oh, how dreadful! I’m so sorry. \\
        \hline
        It’s dangerous. I dare not go. & Fear & It’s dangerous. I don’t dare go. & It is hazardous, and I am apprehensive about venturing there. & It’s a bit risky, but I’m not sure I can go through with it. & It’s dangerous. I dare not go. \\
        \hline
        I’m very nervous about this oral test. & Fear & I’m incredibly nervous about this oral exam. & I am experiencing significant apprehension regarding the upcoming oral examination. & I’m a bit nervous about this oral test, but I’m ready to ace it! & I’m very nervous about this oral test. \\
        
        \hline
        Man, I’m freaking out! You gotta help me! & Fear & Man, I’m freaking out! You have to help me! & I am experiencing significant distress and require your assistance. & I’m feeling a bit overwhelmed and could really use your help. & Man, I’m freaking out! You have to help me! \\
        \hline
        I can’t go through with this! I’m not ready for marriage. I’m only thirty-five, have my whole life ahead, and adventures waiting. I can’t settle down yet! & Fear & I can’t go through with this! I’m not ready for marriage! What was I thinking? I’m only thirty-five years old! I have my entire life ahead of me, full of adventures waiting! I can’t settle down yet! & I am unable to proceed with this decision. I am not prepared for marriage. What was I thinking? I am only thirty-five years old! I have my entire life ahead of me, filled with adventures waiting to be experienced. I cannot settle down yet. & I’m feeling a bit overwhelmed and unsure about this decision. I’m not ready for marriage just yet. I’m only 35, and I have so much of my life ahead of me. I have so many adventures waiting to happen! & I can’t go through with this! I’m not ready for marriage. I’m only thirty-five, have my whole life ahead, and adventures waiting. I can’t settle down yet! \\
        \hline
        School. Since I worked part-time, I struggled to keep up with classes. Now that exams are approaching, I’m panicking. & Fear & School has been challenging since I started working part-time. Now that exams are approaching, I’m completely freaking out. & Since I have been working part-time, I have been struggling to keep up with my coursework. With exams approaching, I am experiencing significant anxiety. & School. Since I was working part-time, I had a tough time keeping up with my studies. Now that exams are coming up, I’m feeling super stressed out! & School. Since I worked part-time, I struggled to keep up with classes. Now that exams are approaching, I’m panicking. \\
        \hline
        No way! I’d definitely break my leg. & Fear & Absolutely not! I’m certain I’d end up breaking my leg. & Certainly not! I am aware that I would likely sustain a leg fracture. & I’m afraid I won’t be able to do that. I’m not very good at jumping. & No way! I’d definitely break my leg. \\
        \hline
        He followed me home all the way! & Fear & He followed me all the way home! & He accompanied me all the way home! & He followed me all the way home, like a loyal dog! & He followed me home all the way! \\
        \hline
        Yes, he was there. I wasn’t imagining it. Finally, when I got home, I turned around and saw him. He didn’t smile. He just stood there. It was obvious. What should I do? I’m scared. Can I call the police? & Fear & Yes, he was. I’m not imagining it. Finally, just when I got home, I turned around and looked at him. He was standing there, unsmiling. It was so obvious. What should I do? I’m so scared. Can I call the police? & Indeed, he was. I am not imagining it. Finally, just as I arrived home, I turned around and looked at him. He was standing there, unsmiling. It was so obvious. What should I do? I am terrified. Can I call the police? & Yes, he was there. I’m not imagining things. Just when I got home, I turned around and saw him. He was standing there, not smiling. It was so obvious. I’m scared, what should I do? Can I call the police? & Yes, he was there. I wasn’t imagining it. Finally, when I got home, I turned around and saw him. He didn’t smile. He just stood there. It was obvious. What should I do? I’m scared. Can I call the police? \\
        \hline
        Why can’t I call the police now? He was definitely following me! & Fear & Why can’t I call the police now? He was definitely following me! & Why am I unable to contact the police at this moment? I am certain that he was following me! & I’m feeling a bit uneasy and would like to report a possible incident. I believe someone was following me. & Why can’t I call the police now? He was definitely following me! \\
        \hline
        What can they do? I don’t even know who he is. & Fear & I suppose they can do something. I don’t even know who he is. & It is uncertain what they can do, as I am unfamiliar with the individual in question. & Hmm, I’m not sure what they can do. I don’t even know who he is. & What can they do? I don’t even know who he is. \\
        \hline
        I’m scared of cats. & Fear & No, I’m afraid of cats. & On the contrary, I am genuinely apprehensive towards felines. & No, I’m terrified of cats! & I’m scared of cats. \\
        \hline
        Is it a horror film? & Fear & Is it a horror movie? & Is this a horror film? & Is it a horror movie? & Is it a horror film? \\
        \hline
        No way. I will be scared out of my wits. & Fear & No way! I’ll be terrified. & Under no circumstances will I consent to such an action. The prospect of it fills me with overwhelming fear. & Oh no, that sounds quite frightening. & I’ll be terrified. \\
        \hline
        I don't know what to do, if I can't find the money. & Fear & I’m at a loss about what to do if I can’t find the money. & I am at a loss regarding how to proceed if I am unable to secure the necessary funds. & I’m feeling a bit overwhelmed, I need to find a way to make the money. & I don’t know what to do if I can’t find the money. \\
        \hline
        It's a large sum of money. I can't afford to lose it. & Fear & It’s a substantial amount of money. I can’t afford to lose it. & The amount in question is substantial, and I am unable to afford to lose it. & It’s a big chunk of change. I can’t afford to let it slip away. & It’s a large sum of money I can’t afford to lose. \\
        \hline
        What you said gives me the creeps! & Fear & What you said makes me feel uneasy! & Your statement unsettles me. & I must admit, your statement made me feel a bit uneasy. & What you said makes me uneasy! \\
        \hline
    "I feel really overwhelmed with mine." & Fear & I feel incredibly overwhelmed with mine. & I am experiencing a significant sense of overwhelm with my current workload. & My to-do list is a total beast! & I feel overwhelmed with mine. \\
    \hline
    "I feel paranoid." & Fear & I feel paranoid and uneasy. & I am experiencing a sense of apprehension. & I’m feeling a bit on edge. & I feel paranoid. \\
    \hline
    "I had climbed on a cherry tree alone, and there was a thick caterpillar beside my fingers. I feel disgusted by caterpillars and snakes. I was terribly afraid of the caterpillar crawling on my fingers. Out of fear, I was almost unable to climb down." & Fear & I had climbed up a cherry tree all by myself, and there was a thick caterpillar right beside my fingers. I have a strong aversion to caterpillars and snakes. I was terrified of the caterpillar crawling on my fingers. Out of sheer fear, I was almost unable to climb down. & While ascending a cherry tree alone, I encountered a substantial caterpillar adhering to my fingers. Due to my aversion to caterpillars and snakes, I experienced a profound sense of disgust. The prospect of the caterpillar crawling on my fingers instilled a paralyzing fear within me, nearly rendering me incapable of descending the tree. & So, I was up in this cherry tree all by myself, and guess what? There was this huge caterpillar right next to my fingers! I mean, I’m not a big fan of caterpillars or snakes, so I was totally freaked out. I couldn’t even move my fingers! I was stuck there, paralyzed with fear. & I climbed a cherry tree alone and found a thick caterpillar beside my fingers. I detest caterpillars and snakes. Terrified, I couldn’t climb down as the caterpillar crawled on my fingers. \\
    \hline
    "I feel scared and unsure and out of place." & Fear & I feel scared, uncertain, and out of place. & I am experiencing a profound sense of apprehension, trepidation, and a profound feeling of alienation. & I feel a bit overwhelmed and out of my element. & I feel scared, uncertain, and out of place. \\
    \hline

    \end{tabular}
\end{table*}

\begin{table*}[h]
\vspace{-60px}
    \centering
    \tiny % Reduce font size to make table compact
    \renewcommand{\arraystretch}{1.2} % Improve row spacing
    \setlength{\tabcolsep}{2pt} % Adjust column spacing for better fit
    \caption{Evaluation of dataset 2 Dialog: Happiness}
    \label{tab:dialog_happiness}
    \begin{tabular}{|p{3cm}|p{1cm}|p{3cm}|p{3cm}|p{3cm}|p{3cm}|}
        \hline
        \textbf{Original Text} & \textbf{Type} & \textbf{Apple Rewrite} & \textbf{Apple Professional} & \textbf{Apple Friendly} & \textbf{Apple Concise} \\
        \hline
        That's a good idea. I hear Mary and Sally often go there to play pingpong. Perhaps we can make a foursome with them. & Happiness & That’s a great idea! I often hear Mary and Sally going there to play pingpong. Maybe we could invite them over for a foursome. & That is an excellent suggestion. I have heard that Mary and Sally frequently visit that establishment to engage in ping pong. Perhaps we could organize a foursome with them. & That’s a great idea! I’ve heard Mary and Sally go there to play pingpong a lot. Maybe we could invite them over for a friendly game? & That’s a good idea. I hear Mary and Sally often play pingpong there. Maybe we can join them for a foursome. \\
        \hline
        Sounds great to me! If they are willing, we could ask them to go dancing with us. That is excellent exercise and fun, too. & Happiness & It sounds like a great idea! If they’re up for it, we could invite them to join us for a dance. It’s an excellent form of exercise and a lot of fun. & It sounds like an excellent idea! If they are amenable, we could invite them to join us for a dance session. This would not only provide an excellent form of exercise but also offer an enjoyable and entertaining experience. & That sounds awesome! If they’re up for it, we could ask them to join us for a dance. It’s a great way to get some exercise and have fun at the same time. & Sounds great! If they’re willing, we could ask them to dance with us. It’s excellent exercise and fun too. \\
        \hline
        Good. Let’s go now. & Happiness & Good. Let’s go now. & Let us proceed now. & Awesome! Let’s get moving! & Let’s go now. \\
        \hline
        I’m looking at my horoscope for this month! My outlook is very positive. It says that I should take a vacation to someplace exotic, and that I will have a passionate summer fling! & Happiness & I’m checking my horoscope for this month, and it has a very positive outlook. It suggests that I should take a vacation to an exotic destination and that I’ll have a passionate summer fling! & Upon reviewing my horoscope for this month, I have a highly optimistic outlook. It advises me to embark on an exotic vacation and predicts a passionate summer romance. & I’m checking out my horoscope for this month, and guess what? It’s got some amazing predictions! It says I should take a trip to a tropical paradise and have a wild summer fling. Sounds like a dream come true, right? & I’m looking at my horoscope for this month. It predicts a positive outlook. It suggests I take an exotic vacation and have a passionate summer fling! \\
        \hline
        Let’s see… you’re a Capricorn. It says that you will be feeling stress at work, but you could see new, exciting developments in your love life. Looks like we’ll both have interesting summers! & Happiness & Let’s see… You’re a Capricorn. According to the horoscope, you’ll be feeling stressed at work, but you might also experience new and exciting developments in your love life. It seems like we’ll both have interesting summers! & Let us examine the astrological predictions. As a Capricorn, you are likely to experience stress at work. However, there is a possibility of exciting developments in your romantic life. It appears that we will both have fulfilling summers. & So, you’re a Capricorn, huh? According to the stars, you’ll be feeling a bit stressed out at work, but guess what? You might also discover some exciting new developments in your love life. It seems like we’re both in for a treat this summer! & Let’s see… You’re a Capricorn. You’ll feel stressed at work, but you’ll see exciting developments in your love life. We’ll both have interesting summers! \\
        \hline
        A Capricorn is serious-minded and practical. She likes to do things in conventional ways. That sounds just like you! & Happiness & A Capricorn is serious-minded and practical. She prefers to follow conventional methods. That description seems to fit you perfectly! & A Capricorn is known for their serious and practical nature. They tend to follow conventional methods and routines. This description aligns well with your personality traits. & A Capricorn is all about being serious and practical. She likes to follow the rules and do things the old-fashioned way. That sounds just like you! & A Capricorn is serious, practical, and conventional. That sounds like you! \\
        \hline
        It’s a wonderful Spanish style. & Happiness & It’s a wonderful Spanish-style restaurant. & This Spanish-style dish is truly delightful. & This Spanish-style dish is truly delightful. & It’s a wonderful Spanish style. \\
        \hline
        Oh, I love the roof tiles on Spanish-style houses. & Happiness & Oh, I absolutely adore the roof tiles used in Spanish-style houses. & I am particularly fond of the roof tiles used in Spanish-style houses. & Oh, I adore the charming roof tiles on Spanish-style houses! & Oh, I love the roof tiles on Spanish-style houses. \\
        \hline
        And it’s a bargaining. A house like this in river side costs double the price. & Happiness & It’s a bargaining. A house like this on the riverbank costs double the price. & This property is a bargaining chip. A house of this caliber on the riverbank would typically cost double its current price. & It’s a negotiation. A house like this on the riverbank costs double the price. & It’s a bargaining. A house like this on the riverbank costs double the price. \\
        \hline
        Great, is it a two-bedroom house? & Happiness & Is it a two-bedroom house? & Could you please confirm if the property is a two-bedroom house? & Awesome! Is it a cozy two-bedroom house? & Is it a two-bedroom house? \\
        \hline
        That’s a nice area too. It’ll be a good investment for you. & Happiness & That’s a nice area too. It’ll be a good investment for you. & That is also a desirable area. It will be a worthwhile investment for you. & That’s a great area too! It’ll be a fantastic investment for you. & That’s a nice area too. It’ll be a good investment for you. \\
        \hline
        That sounds like a good idea. I'll have to try that. & Happiness & That sounds like a great idea. I’ll definitely try that. & That appears to be a sound suggestion. I will endeavor to implement it. & That sounds like a great idea! I’ll definitely give it a try. & That sounds good. I’ll try that. \\
        \hline
        Yes. I like cooking very much. I got this hobby when I was 12 years old. & Happiness & Yes, I absolutely love cooking. I picked up this hobby when I was 12 years old. & Indeed, I have a profound passion for cooking. This hobby was ignited during my twelfth year of schooling. & Yes, I absolutely love cooking! I picked it up when I was 12 years old. & Yes, I love cooking. I picked up this hobby when I was 12. \\
        \hline
        You look so tan and healthy! & Happiness & You look so radiant and healthy! & Your appearance exudes a remarkable sense of vitality and well-being. & You look so radiant and healthy! & You look so healthy and tan! \\
        \hline
        Thanks. I just got back from summer camp. & Happiness & Thanks! I just returned from summer camp. & Thank you. I have recently returned from summer camp. & Thanks! I just got back from summer camp. & Thanks. I just returned from summer camp. \\
        \hline
        Great. I got to try so many things for the first time. & Happiness & It’s great! I got to try so many things for the first time. & It was an extraordinary experience, allowing me to sample a diverse array of novel culinary delights. & Awesome! I got to try so many new things for the first time. & I got to try many new things for the first time. \\
        \hline
        The counselors were so nice too. It was the best summer ever. Except for all those mosquitoes! & Happiness & The counselors were incredibly kind as well. It was the best summer ever, except for the pesky mosquitoes! & The counselors were also very kind. It was the best summer ever, with the exception of the abundance of mosquitoes. & The counselors were super cool! It was the best summer ever, except for those pesky mosquitoes! & The counselors were very nice. It was the best summer ever, except for the mosquitoes! \\
        \hline
        That’s all right. Thank you all the same. & Happiness & That’s all right. Thank you all the same. & Thank you all the same. That is all right. & That’s fine. Thanks a bunch! & That’s all right. Thank you all the same. \\
        \hline
        Thank you for your help. & Happiness & Thank you for your help. & Thank you for your assistance. & Thanks a bunch for your help! & Thanks for your help. \\
        \hline
        This is a really great place to watch a baseball game. & Happiness & This is an excellent place to watch a baseball game. & This is an exceptional venue for observing a baseball game. & This is a really great place to watch a baseball game, the atmosphere is electric! & This is a great place to watch a baseball game. \\
        \hline
        Yeah, there isn’t a bad seat in the place. & Happiness & Yeah, there isn’t a bad seat in the place. & Indeed, there is no undesirable seat in the establishment. & Yeah, there isn’t a bad seat in the house! & There isn’t a bad seat in the place. \\
        \hline
        The fans are great here, too. Not like in Philadelphia. & Happiness & The fans here are fantastic, unlike in Philadelphia. & The fans here are also excellent, unlike in Philadelphia. & The fans here are awesome, too! Not like in Philadelphia. & The fans here are great, unlike in Philadelphia. \\
        \hline
        It was a great idea to spend a day watching a game. & Happiness & It was an excellent idea to spend a day watching a game. & It was an excellent concept to dedicate an entire day to watching a sporting event. & It was a blast to spend a day watching a game! & It was a great idea to spend a day watching a game. \\
        \hline
        Yeah, it reminds you why they say baseball is America’s favorite pastime. & Happiness & Yeah, it reminds you why they say baseball is America’s favorite pastime. & Indeed, it serves as a poignant reminder of the reasons why baseball is widely regarded as America’s cherished pastime. & Yeah, it definitely reminds you why they say baseball is America’s favorite pastime. & Yeah, it reminds you why they say baseball is America’s favorite pastime. \\
        \hline
        How do you like the pizza here? & Happiness & How do you like the pizza here? & Could you please share your thoughts on the pizza here? & How’s the pizza treating you? & How do you like the pizza here? \\
        \hline
        Perfect. It really hits the spot. & Happiness & Perfect! It truly resonates with me. & This is indeed an excellent piece of writing. It effectively conveys its intended message. & That’s spot on! It really resonates with me. & Perfect. It really hits the spot. \\
        \hline
        Were you? Your performance was excellent. & Happiness & Were you there? Your performance was excellent. & Were you present? Your performance was exceptional. & Were you there? Your performance was absolutely outstanding! & Were you? Your performance was excellent. \\
        \hline
        Thank you for your kind words. & Happiness & Thank you for your kind words. & Thank you for your gracious words. & Thanks a bunch for your kind words! & Thanks for your kind words. \\
        \hline
        Sure. Which button do I press to shoot? & Happiness & Sure. Which button do I press to shoot? & Could you please indicate which button I should press to initiate the shooting mechanism? & Sure thing! Which button do I press to unleash my inner photographer? & Which button do I press to shoot? \\
        \hline
        Alright, say cheese! & Happiness & Alright, everyone! Say cheese! & Capture the moment! & All right. Say cheese! & Alright, say cheese! \\
        \hline
        No, thanks. I’m just looking. & Happiness & No, thanks. I’m just looking. & No, thank you. I am merely browsing. & No, thanks. I’m just browsing. & No, thanks. I’m just looking. \\
        \hline
        Sure, I’ll let you know if I need anything. This mattress is firm. Jack will likely like it. & Happiness & Sure, I’ll let you know if I need anything. Hmm, this mattress is quite firm. Jack will likely enjoy it. & Certainly, I will inform you if I require any assistance. This mattress is quite firm. Jack is likely to appreciate it. & Sure thing! I’ll keep you posted if I need anything. Hmm, this mattress is super firm. Jack might actually like it! & Sure, I’ll let you know if I need anything. This mattress is firm. Jack will likely like it. \\
        \hline
        
    \end{tabular}
\end{table*}

\begin{table*}[h]
\vspace{-60px}
    \centering
    \tiny
    \renewcommand{\arraystretch}{1.2}
    \setlength{\tabcolsep}{2pt}
    \caption{Evaluation of dataset 2 Dialog: Surprise}
    \label{tab:dialog_surprise}
    \begin{tabular}{|p{3cm}|p{1cm}|p{3cm}|p{3cm}|p{3cm}|p{3cm}|}
        \hline
        \textbf{Original Text} & \textbf{Type} & \textbf{Apple Rewrite} & \textbf{Apple Professional} & \textbf{Apple Friendly} & \textbf{Apple Concise} \\
        \hline
        Really? I think that’s impossible! & Surprise & Really? I find that impossible! & Indeed, I find that notion to be highly improbable. & I’m not sure about that, but I’m open to hearing your thoughts. & I think that’s impossible. \\
        \hline
        Oh, which team? & Surprise & Oh, yeah? Which team? & Could you please specify which team you are referring to? & Oh, really? Which team are you talking about? & Oh, which team? \\
        \hline
        Really? We play you next week. I have to go to practice. See you later. & Surprise & Really? I believe we’ll play you guys next week. Well, I have to go to practice. See you later. & Indeed, I believe we will be playing against you next week. I must attend practice now. I will see you later. & Hey, really? I think we’re playing you guys next week. Well, I gotta head to practice. Catch ya later! & Really? We play you next week. I have to go to practice. See you later. \\
        \hline
        Really? I hope I can try it. Please let me know. & Surprise & Really? I hope I get a chance to taste it. Please let me know. & Indeed, I would be delighted to have the opportunity to sample this delicacy. Please do not hesitate to inform me when it becomes available. & Oh wow, that sounds amazing! I hope I get a chance to try it. Please let me know if you have any samples or if there’s a way I can get my hands on it. & Really? I hope I can try it. Please let me know. \\
        \hline
        That’s truly amazing! & Surprise & That is truly amazing! & That is indeed remarkable! & That is absolutely incredible! & That’s truly amazing! \\
        \hline
        Wow, I don’t have to go to school tomorrow. & Surprise & Really? Wow! I don’t have to go to school tomorrow. & Indeed, I am relieved to learn that I will not be required to attend school tomorrow. & That’s awesome! I’m so glad I don’t have to go to school tomorrow. & Wow, I don’t have to go to school tomorrow. \\
        \hline
        Why do you say that’s a crazy language? & Surprise & Is that a crazy language? Why do you say that? & Could you please elaborate on why you consider this language to be peculiar? & I’m not familiar with that language, so I’m not sure what you mean. & Why do you say that’s a crazy language? \\
        \hline
        What? Excuse me. & Surprise & What? Beg your pardon. & Excuse me, could you please repeat that? & Excuse me, could you repeat that? & What? Excuse me. \\
        \hline
        I can’t believe my ears! & Surprise & I can’t believe my ears! & I am astounded by the news I have received. & I’m surprised and a bit shocked by that. & I can’t believe my ears! \\
        \hline
        Are you sure? & Surprise & Really? Are you sure? & Could you please confirm this information? & I’m curious, are you certain about that? & Are you sure? \\
        \hline
        Really? What time does it start? & Surprise & Are you serious? Well, when does it start? & Could you please provide me with the start time? & Oh wow, really? When does it start? & What time does it start? \\
        \hline
        What did you buy? & Surprise & What did you buy? & Could you please elaborate on what you purchased? & What did you get yourself? & What did you buy? \\
        \hline
        I can’t believe your stupidity. Why didn’t you open the box? & Surprise & I find it unbelievable how foolish you are. Why didn’t you open the box? & I am astounded by your lack of intelligence. Why did you fail to open the box? & I noticed that the box was left unopened. Would you mind opening it for me? & I can’t believe your stupidity. Why didn’t you open the box? \\
        \hline
        Well, that’s new. If I had to choose, honesty is what I value most in a friend. & Surprise & Well, that’s something I’ve never heard of before. However, if I were to ask you what I value most in a friend, I would say honesty. & Indeed, that is an uncommon perspective. However, if I were to consider the most valuable trait in a friend, I would assert that honesty is paramount. & That’s a new one! But if I had to pick one thing I value most in a friend, it would be honesty. & Well, that’s new. If I had to choose, honesty is what I value most in a friend. \\
        \hline
        That’s great! I know your company has many clients in France. & Surprise & Really? That’s wonderful! I understand that your company has a significant number of clients in France. & Indeed, that is wonderful news! I am aware that your company has a substantial clientele in France. & That’s awesome! I know your company has a lot of clients in France, so this is a great opportunity for them. & That’s great! I know your company has many clients in France. \\
        \hline
        Alright, let’s talk about something else instead. & Surprise & Alright, guy? Alright. Let’s talk about something else instead. & Certainly, let’s steer the conversation towards a more appropriate topic. & Hey there! How about we switch gears and talk about something else? & Alright, let’s talk about something else instead. \\
        \hline
        Are you going to introduce us? & Surprise & Are you going to introduce us? & Could you please introduce us to the speaker? & Hey, would you mind introducing us to the person you’re talking about? & Are you going to introduce us? \\
        \hline
        Was I? Sorry, I didn’t mean to be. I apologize. & Surprise & Was I? Sorry, I didn’t mean to be. I do apologize. & May I clarify? I apologize if my previous statement was unclear. & Oh no, I’m so sorry! I didn’t mean to be that way. Please forgive me. & Was I? Sorry, I didn’t mean to be. I apologize. \\
        \hline
        Did you wear red underwear again this year? & Surprise & Did you do something unusual this year? How did you manage to pull it off? Did you wear red underwear again? & Could you please clarify how you managed to do that? Did you wear red underwear once more this year? & I noticed you wore red underwear today. Did you have a special reason for that? & Did you wear red underwear again this year? \\
        \hline
        What? Did you do that? & Surprise & What on earth did you do? & What on earth did you do? & Oh, I didn’t mean to do that. Let’s try to fix it together. & What? Did you do that? \\
        \hline
        What is this abomination you call a stir fry? More meat than vegetables? That’s the worst insult to a Chinese stir fry. It’s a disgrace to the wok it was fried in. It’s like a fajita without the wrap. & Surprise & What on earth is that you call a stir-fry? More meat than vegetables? That’s the most insulting thing you could say about a Chinese stir-fry. It’s a disgrace to the wok it was fried in! What you had is like a fajita without the wrap! & What? You consider that a stir-fry? More meat than vegetables? That is the most insulting critique you could level at a Chinese stir-fry. It is a disgrace to the wok in which it was cooked! What you had is essentially a fajita without the tortilla wrap. & I’m a bit surprised by your comment about the stir fry. It seems like there might have been a misunderstanding. The dish you described doesn’t quite align with the traditional Chinese stir fry. & What is this abomination you call a stir fry? More meat than vegetables? That’s the worst insult to a Chinese stir fry. It’s a disgrace to the wok it was fried in. It’s like a fajita without the wrap. \\
        \hline
        Really? Well, okay, you’re welcome. & Surprise & Are you serious? Well, okay, you’re welcome. & Indeed. Thank you for your assistance. & Sure thing! & Really? Well, okay, you’re welcome. \\
        \hline
        Really? I never would have guessed. Criminals must only come out at night because I’ve never noticed anything strange when I’ve been at your house during the day. & Surprise & Really? I never would have guessed that. The criminals must only come out at night because I’ve never noticed anything unusual when I’ve been at your house during the day. & Indeed, I had never anticipated such an occurrence. It appears that criminals may only emerge during the evening hours, as I have not observed any unusual activity during my daytime visits to your residence. & Wow, really? I never thought of that. It seems like the criminals only come out at night, because I’ve never noticed anything unusual when I’ve been at your place during the day. & Really? I never would have guessed. Criminals must only come out at night because I’ve never noticed anything strange when I’ve been at your house during the day. \\
        \hline
        What? Excuse me. & Surprise & What? Excuse me. & Excuse me, could you please repeat that? & Excuse me, could you repeat that? & What? Excuse me. \\
        \hline
        I can’t believe my ears! & Surprise & I can’t believe my ears! & I am astounded by the news I have received. & I’m surprised and a bit shocked by that. & I can’t believe my ears! \\
        \hline
        Hey Maris, are you a part-time employee? & Surprise & Hey Maris, are you a part-time employee? & Hello Maris, are you a part-time employee? & Hey Maris, are you a part-timer? & Hey Maris, are you a part-time employee? \\
        \hline
        No, it can’t be. Who? & Surprise & No, it can’t be. Really? Who? & No, it cannot be. Could you please provide more information? & Oh no, that can’t be true! Who would do that? & No, it can’t be. Who? \\
        \hline
        Oh my God, John! I didn’t know you were religious. You just look like a scientific skeptic, like Leonard in The Big Bang Theory. & Surprise & Oh my God! John, I didn’t know you were religious. You just look like a scientific skeptic, like Leonard in The Big Bang Theory. & Hello John, I was unaware that you were religious. Your appearance suggests that you may be more inclined towards scientific pursuits, similar to Leonard Hofstadter in the television series “The Big Bang Theory.” & Hey John, I didn’t realize you were religious. You just seem like a bit of a science nerd, like Leonard in The Big Bang Theory. & Oh my God, John! I didn’t know you were religious. You just look like a scientific skeptic, like Leonard in The Big Bang Theory. \\
        \hline
        Why were most of them religious people? & Surprise & Why were most of them religious people? & Why did most of them identify as religious individuals? & I’m curious, why did most of them seem to be religious? & Why were most of them religious people? \\
        \hline
        Why did you ask me about universities last year? & Surprise & University? Why did you ask me about it last year? & University? Why? You inquired about this last year. & University? Why are you asking me again? I already told you last year. & Why did you ask me about universities last year? \\
        \hline
        Pay him? He studies for himself, not for anyone else. I have an idea. Send him to class every morning, and then go to work. & Surprise & Pay him? He studies for himself, not for you, me, or anyone else. I have an idea. You should send him to class every morning, and after the class begins, you can go to work. & Compensate him? He studies independently, not for you, me, or anyone else. I have an idea. You should send him to class every morning, and after the class begins, you can proceed to work. & Should we pay him? He’s studying for himself, not for us or anyone else. I have an idea. You could send him to class every morning, and then you can go to work. & Pay him? He studies for himself, not for anyone else. I have an idea. Send him to class every morning, and then go to work. \\
        \hline
    \end{tabular}
\end{table*}

\begin{table*}[h]
\vspace{-60px}
    \centering
    \tiny
    \renewcommand{\arraystretch}{1.2}
    \setlength{\tabcolsep}{2pt}
    \caption{Evaluation of dataset 2 Dialog: Sadness}
    \label{tab:dialog_sadness}
    \begin{tabular}{|p{3cm}|p{1cm}|p{3cm}|p{3cm}|p{3cm}|p{3cm}|}
        \hline
        \textbf{Original Text} & \textbf{Type} & \textbf{Apple Rewrite} & \textbf{Apple Professional} & \textbf{Apple Friendly} & \textbf{Apple Concise} \\
        \hline
        My parents called. As usual, they reminded me again that I should have a plan to marry by my late 20s. Easier said than done. Who should I marry? I have no time to go on a date. & Sadness & My parents called me again, as usual, reminding me that I should have a plan to get married by my late 20s. It’s easier said than done, though. Who should I marry? I don’t have time to go on dates. & My parents contacted me, as they do every so often, to remind me that I should have a plan for marriage by my late 20s. However, this is easier said than done. I am currently facing a lack of time to pursue romantic relationships. & My parents called again, just like they always do. They reminded me that I should have a plan to get married by the time I’m in my late 20s. It’s so much easier said than done, though! Who should I even think about marrying? I don’t have the time to go on dates right now. & My parents called again, reminding me to have a plan to marry by my late 20s. Who should I marry? I have no time for dates. \\
        \hline
        I am pretty sure that I am in trouble. & Sadness & I am quite certain that I am in trouble. & I am quite certain that I am in a precarious situation. & I am pretty sure that I am in trouble. & I’m certain I’m in trouble. \\
        \hline
        I'm sorry to have made you disappointed. & Sadness & I’m sorry to have disappointed you. & I regret to have caused you disappointment. & I’m so sorry to have let you down. & I’m sorry to disappoint you. \\
        \hline
        My drives were terrible today. I kept slicing the ball into the roughs and sand traps. & Sadness & My drives were terrible today. I kept slicing the ball into the roughs and sand traps. & Today’s driving experience was highly challenging. I consistently misjudged the distance and struck the ball into the roughs and sand traps. & My drives were a bit wild today. I couldn’t seem to keep the ball out of the roughs and sand traps. & My drives were terrible today. I kept slicing the ball into the roughs and sand traps. \\
        \hline
        John, I am just not in the mood for this. & Sadness & John, I’m simply not in the mood for this. & Hello John, I regret to inform you that I am not currently available to engage in this discussion. & John, I’m feeling a bit overwhelmed and not up for this right now. & John, I’m not in the mood for this. \\
        \hline
        I lost the table tennis game yesterday. & Sadness & I lost yesterday’s table tennis game. & I was defeated in yesterday’s table tennis match. & I lost the table tennis game yesterday, but hey, I’ll get them next time! & I lost yesterday’s table tennis game. \\
        \hline
        I just don’t want to play table tennis anymore. & Sadness & I just don’t want to play table tennis anymore. & I have decided to cease playing table tennis. & I’m feeling a bit out of touch with table tennis these days. & I no longer want to play table tennis. \\
        \hline
        My opponent bowled me with the very first ball. I was wondering if it's appropriate for me to play table tennis. & Sadness & My opponent bowled me with the very first ball. I was wondering if it’s appropriate for me to play table tennis. & My opponent defeated me with the first ball bowled. I am considering whether table tennis is an appropriate sport for me. & My opponent totally crushed me with the first ball! I’m thinking about trying table tennis. Do you think I should give it a shot? & My opponent bowled me with the first ball. Is table tennis appropriate for me? \\
        \hline
        I just had a terrible week. & Sadness & I had a terrible week. & This week has been particularly challenging for me. & I just had a rough week, but I’m trying to stay positive. & I had a terrible week. \\
        \hline
        
        I fell down and hurt my arm. & Sadness & I had an unfortunate fall and sustained an arm injury. & I sustained an injury to my arm after falling. & I took a tumble and scraped my arm. & I fell and hurt my arm. \\
        \hline
        I’m rather disappointed at not being able to see my best friend off. & Sadness & I’m quite disappointed that I won’t be able to see my best friend off. & I am rather disappointed that I was unable to bid farewell to my closest friend. & I’m a bit bummed that I can’t make it to say goodbye to my bestie. & I’m disappointed I couldn’t see my best friend off. \\
        \hline
        I fell on the way to school, and your bike got scratched. I'm really sorry. & Sadness & I’m truly sorry to inform you that I accidentally fell on my way to school and scratched your bike. & While en route to school, I unfortunately fell and caused your bicycle to sustain a scratch. I sincerely apologize for any inconvenience this may have caused. & Oops! I had a bit of a mishap on my way to school and unfortunately, your bike got a little scratched. I’m so sorry about that! & I accidentally scratched your bike on my way to school. I’m really sorry. \\
        \hline
        My girlfriend dumped me. & Sadness & My girlfriend ended our relationship. & My romantic partner ended our relationship. & My girlfriend and I had a breakup. & My girlfriend broke up with me. \\
        \hline
        Yes, I know, sir. I'm sorry. & Sadness & Yes, I understand, sir. I apologize. & Indeed, I understand your concern, sir. I apologize for any inconvenience this may have caused. & Yes, I know, sir. Sorry about that! & Yes, sir, I know. I’m sorry. \\
        \hline
        I didn't sleep a wink last night. & Sadness & I didn’t get a single wink of sleep last night. & I experienced a complete lack of sleep throughout the night. & I didn’t catch a single wink last night. & I didn’t sleep last night. \\
        \hline
        Well, I am under a lot of pressure. My boss is very pushy. He assigned me three projects. Now the deadlines are near, and I still have not finished all of my projects. & Sadness & Well, I’m under a lot of pressure. My boss is very pushy and has assigned me three projects. Now that the deadlines are approaching, I still haven’t finished all of them. & Currently, I am experiencing significant pressure due to the demanding nature of my boss. He has assigned me three projects, and as the deadlines approach, I find myself struggling to complete all of them. & I’m feeling a bit overwhelmed right now. My boss is super demanding and has given me three projects to work on, with tight deadlines looming. I’m doing my best, but I’m not sure if I’ll be able to finish everything on time. & I’m under immense pressure. My pushy boss assigned me three projects with nearing deadlines, and I haven’t finished any of them. \\
        \hline
        I am not in the mood. & Sadness & I’m not feeling up to it. & I am currently experiencing a lack of enthusiasm. & I am not in the mood for anything right now. & I’m not in the mood. \\
        \hline
        I was told my girlfriend was speaking ill of me. It's a real letdown. & Sadness & I was informed that my girlfriend was spreading rumors about me. It’s a significant disappointment. & I was informed that my girlfriend has been disparaging about me. This news is deeply disappointing. & I heard my girlfriend was talking negatively about me. It’s a bit disappointing. & I was told my girlfriend was speaking badly of me. It’s disappointing. \\
        \hline
        But she did and it made me disappointed. & Sadness & She did it, and it disappointed me. & Despite her efforts, I was left feeling disappointed. & But she did and it left me feeling a bit let down. & She did, and I was disappointed. \\
        \hline
        But she means a lot to me. & Sadness & She holds a great deal of significance for me. & She holds significant importance in my life. & She means the world to me! & She means a lot to me. \\
        \hline
        Oh, I just can't forget it. & Sadness & Oh, I can’t seem to forget it. & The memory of that incident remains vivid in my mind. & Oh, that memory is still so vivid! & Oh, I can’t forget it. \\
        \hline
        I failed my oral test. & Sadness & I failed my oral exam. & I regrettably failed my oral examination. & I failed my oral test. & I failed my oral exam. \\
        \hline
        That’s weird. & Sadness & That’s strange. & That is peculiar. & That’s quite peculiar. & That’s weird. \\
        \hline
        NYU was my first choice. & Sadness & New York University was my first choice of college. & New York University was my preferred choice. & NYU was my top pick! & NYU was my top choice. \\
        \hline
        I’m sorry that you didn’t get in. & Sadness & I’m sorry you didn’t get in. & I regret to inform you that you were not selected for admission. & I’m sorry that you didn’t get in. & I’m sorry you didn’t get in. \\
        \hline
        Bummer. I guess we're both out of luck. & Sadness & It’s unfortunate. I suppose we’re both in a losing position. & Regrettably, it appears that our endeavors have not yielded favorable outcomes. & Oh no, it seems like we’re both out of luck this time. & Bummer. We’re both unlucky. \\
        \hline
        Oh, I'm sorry. & Sadness & Oh, I’m sorry. & My apologies for any inconvenience this may have caused. & Oh, I’m so sorry! & Oh, I’m sorry. \\
        \hline
        I'm so sorry, I… I didn't realize it was that loud. & Sadness & I’m so sorry; I didn’t realize it was that loud. & I sincerely apologize for not realizing the extent of the noise. & Oh no, I didn’t realize it was that loud! & I’m sorry, I didn’t realize it was that loud. \\
        \hline
        I am not in the mood. & Sadness & I’m not feeling up to it. & I am currently experiencing a lack of enthusiasm. & I am not in the mood for anything right now. & I’m not in the mood. \\
        \hline
        But she did and it made me disappointed. & Sadness & She did it, and it disappointed me. & Despite her efforts, I was left feeling disappointed. & But she did and it left me feeling a bit let down. & She did, and I was disappointed. \\
        \hline
        Sorry, I am not sure, can I take a message? & Sadness & I’m not sure, but can I take a message? & Excuse me, may I take a message? & Sorry, I’m not sure. Can I take a message? & Can I take a message? \\
        \hline
    \end{tabular}
\end{table*}

\begin{table*}[h]
\vspace{-60px}
    \centering
    \tiny
    \renewcommand{\arraystretch}{1.2}
    \setlength{\tabcolsep}{2pt}
    \caption{Evaluation of dataset 2 Dialog: Neutral}
    \label{tab:dialog_neutral}
    \begin{tabular}{|p{3cm}|p{1cm}|p{3cm}|p{3cm}|p{3cm}|p{3cm}|}
        \hline
        \textbf{Original Text} & \textbf{Type} & \textbf{Apple Rewrite} & \textbf{Apple Professional} & \textbf{Apple Friendly} & \textbf{Apple Concise} \\
        \hline
        Say, Jim, how about going for a few beers after dinner? & Neutral & Hey Jim, how about we grab a few beers after dinner? & Hello Jim, would you be interested in joining me for a few beers after dinner? & Hey Jim, how about we grab a few cold ones after dinner? & Say, Jim, how about we go for a few beers after dinner? \\
        \hline
        You know that is tempting but is really not good for our fitness. & Neutral & You know that’s tempting, but it’s really not good for our fitness. & Although it may be tempting, it is not beneficial for our fitness. & You know that is tempting but is really not good for our fitness. & You know that’s tempting, but it’s not good for our fitness. \\
        \hline
        What do you mean? It will help us to relax. & Neutral & What do you mean by that? It will help us relax. & Could you please clarify? This will assist us in relaxing. & What do you mean? It will help us unwind and enjoy some quality time together. & What do you mean? It’ll help us relax. \\
        \hline
        Do you really think so? I don't. It will just make us fat and act silly. Remember last time? & Neutral & Do you genuinely believe that? I don’t. It will only make us overweight and behave foolishly. Do you recall the last time? & Do you genuinely believe that? I do not. It will only lead to obesity and foolishness. Recall the previous incident? & I’m not sure about that. I think it might be a bit too much for us. Remember when we tried that fad diet last time? & Do you really think so? I don’t. It’ll make us fat and silly. Remember last time? \\
        \hline
        I guess you are right. But what shall we do? I don't feel like sitting at home. & Neutral & I suppose you’re correct. However, what should we do? I’m not in the mood to stay home. & Indeed, you are correct. However, I am not inclined to remain indoors. & You’re absolutely right! But what are we going to do about it? I’m not really in the mood to stay at home. & I suppose you’re right. But what should we do? I don’t want to stay home. \\
        \hline
        I suggest a walk over to the gym where we can play singsong and meet some of our friends. & Neutral & I propose we take a stroll over to the gym where we can sing along to some tunes and catch up with a few of our friends. & I propose that we visit the gymnasium for a leisurely stroll, where we can engage in a sing-along and socialize with our acquaintances. & How about a fun walk to the gym? We can sing some silly songs, meet up with our pals, and have a blast! & Let’s go to the gym to play singsong and meet friends. \\
        \hline
        Can you do push-ups? & Neutral & Are you able to perform push-ups? & Are you capable of performing push-ups? & Can you do some push-ups for me? & Can you do push-ups? \\
        \hline
        Of course I can. It's a piece of cake! Believe it or not, I can do 30 push-ups a minute. & Neutral & Of course, I can do that. It’s a piece of cake! Believe it or not, I can do 30 push-ups in a minute. & Certainly, I am able to assist you with this task. It is not an arduous endeavor, as you may believe. In fact, I possess the ability to perform 30 push-ups per minute. & Absolutely! It’s a breeze! You won’t believe it, but I can do an impressive 30 push-ups in a minute. & Of course, I can. It’s easy! Believe it or not, I can do 30 push-ups a minute. \\
        \hline
        You mean 30 push-ups? & Neutral & Are you referring to 30 push-ups? & Are you referring to 30 push-ups? & You mean 30 push-ups? That’s a lot! & Do you mean 30 push-ups? \\
        \hline
        It's easy. If you do exercise everyday, you can make it, too. & Neutral & It’s easy to make it a habit. If you exercise every day, you can do it too. & Regular exercise is a simple yet effective way to achieve your fitness goals. & It’s totally doable! If you can stick to exercising every day, you’ll be on your way to achieving your fitness goals. & It’s easy. If you exercise daily, you can make it too. \\
                \hline
        Can you study with the radio on? & Neutral & Can you study with the radio playing? & May I study with the radio on? & Can you study with the radio on? & Can you study with the radio on? \\
        \hline
        No, I listen to background music. & Neutral & No, I prefer listening to background music. & No, I listen to background music. & Nope, I’m jamming to some background tunes. & I listen to background music. \\
        \hline
        What is the difference? & Neutral & What is the difference between them? & Could you please clarify the distinction between the two? & What’s the difference between these two things? & What’s the difference? \\
        \hline
        The radio has too many commercials. & Neutral & The radio is bombarded with too many commercials. & The radio broadcasts an excessive number of commercials. & The radio is bombarded with commercials. & The radio has too many commercials. \\
        \hline
        That’s true, but then you have to buy a record player. & Neutral & That’s true, but then you’ll need to purchase a record player. & Indeed, that is true, but then you will need to purchase a record player. & That’s true, but then you’ll need to buy a record player. & That’s true, but then you’ll need to buy a record player. \\
        \hline
        Are you all right? & Neutral & Are you all doing well? & Are you all in good health? & Are you all doing swell? & Are you all well? \\
        \hline
        I will be all right soon. I was terrified when I watched them fall from the wire. & Neutral & I’ll be fine soon. I was terrified when I watched them fall from the wire. & I will be fine soon. I was terrified when I witnessed them fall from the wire. & I’ll be back to my old self in no time! I was absolutely terrified when I saw them fall from the wire. & I’ll be fine soon. I was terrified when I saw them fall from the wire. \\
        \hline
        Don’t worry. He is an acrobat. & Neutral & Don’t worry; it’s an acrobat. & Rest assured, the individual in question is an acrobat. & Don’t worry, he’s a circus star! & Don’t worry, it’s an acrobat. \\
        \hline
        Hey John, nice skates. Are they new? & Neutral & Hey John, those are nice skates. Are they new? & Hello John, your skates appear to be new. Could you please confirm? & Hey John, those skates look awesome! Are they brand new? & Hey John, nice skates. Are they new? \\
        \hline
        Yeah, I just got them. I started playing ice hockey in a community league. So, I finally got myself new skates. & Neutral & Yeah, I just got myself new skates. I started playing ice hockey in a community league, so I finally got myself some new ones. & Indeed, I recently acquired new ice hockey skates. I commenced playing ice hockey in a community league, and this acquisition was a long-awaited milestone. & I just got my hands on some new skates! I’ve been playing ice hockey in a community league for a while now, and I’m so excited to have these new ones. & Yeah, I just got new skates. I started playing ice hockey in a community league. \\
        \hline
        What position do you play? & Neutral & What position do you play on the field? & Could you please tell me what position you play on the team? & What position do you play? & What position do you play? \\
        \hline
        I’m a defender. It’s a lot of fun. You don’t have to be able to skate as fast on defense. & Neutral & I’m a defender, and it’s a lot of fun. You don’t necessarily need to be as fast on defense. & As a defender, the role offers a great deal of enjoyment. While skating speed is not a prerequisite for this position, it is advantageous. & I’m a defender, and it’s super fun! You don’t have to be the fastest skater on defense. & I’m a defender. It’s fun. You don’t need to skate fast on defense. \\
        \hline
        Yeah, you’re a pretty big guy. I play goalie, myself. & Neutral & Yeah, you’re quite a big guy. I play goalie myself. & Indeed, you appear to be a substantial individual. I, myself, am a goalkeeper. & Oh, you’re a big guy! I play goalie too, by the way. & Yeah, you’re a pretty big guy. I play goalie too. \\
        \hline
        All right, see you later. & Neutral & Alright, see you later. & Goodbye, and see you later. & Catch ya later! & See you later. \\
        \hline
        Hey Lydia, what are you reading? & Neutral & Hey Lydia, what are you reading? & Hello Lydia, what are you currently reading? & Hey Lydia, what’s the scoop on what you’re reading? & Hey Lydia, what are you reading? \\
        \hline
        What are you talking about? Let me see that... What are horoscopes? & Neutral & What are you referring to? Let me take a look… What are horoscopes? & Could you please clarify what you are referring to? Allow me to review the information. What are horoscopes? & What are you referring to? Let me take a look… What are horoscopes? & What are you referring to? Let me see… What are horoscopes? \\
        \hline
        It’s a prediction of your month, based on your zodiac sign. You have a different sign for the month and date you were born in. I was born on April 15th, so I’m an Aries. When were you born? & Neutral & It’s a prediction of your month based on your zodiac sign. You have a different sign for the month and date you were born. For instance, I was born on April 15th, so I’m an Aries. When were you born? & This is a prediction of your month, based on your zodiac sign. You have a different sign for the month and date you were born in. I was born on April 15th, so I am an Aries. When were you born? & I’ve got a prediction for you based on your zodiac sign. You have a different sign for the month and date you were born in. I was born on April 15th, so I’m an Aries. When were you born? & It’s a prediction of your month based on your zodiac sign. You have a different sign for the month and date you were born. I was born on April 15th, so I’m an Aries. When were you born? \\
        \hline
        That’s bogus. I don't feel any stress at work, and my love life is practically nonexistent. This zodiac stuff is all a bunch of nonsense. & Neutral & That’s nonsense. I don’t feel any stress at work, and my love life is practically nonexistent. This zodiac stuff is all a bunch of hooey. & The notion that astrology can predict stress levels and romantic relationships is unfounded. I do not experience significant stress at work, and my romantic life is relatively non-existent. Therefore, I find astrology to be a pseudoscience. & I’m not really convinced about the whole zodiac thing. I don’t feel stressed at work and my love life is pretty low-key. It seems like a lot of hooey to me. & That’s nonsense. I don’t feel stressed at work, and my love life is non-existent. \\
        \hline
        No, it’s not, your astrology sign can tell you a lot about your personality. See? It says that an Aries is energetic and loves to socialize. & Neutral & No, it’s not. Your astrology sign can reveal a lot about your personality. See? It says that an Aries is energetic and enjoys socializing. & No, it is not. Your astrology sign can provide valuable insights into your personality traits. For instance, it indicates that an Aries is characterized by their energy levels and their inclination towards social interactions. & Nope, it’s not true. Your astrology sign can reveal a lot about your personality. Check it out! It says that an Aries is energetic and loves to socialize. & No, it’s not. Your astrology sign can reveal a lot about your personality. For instance, an Aries is energetic and loves socializing. \\
        \hline
        Well, you certainly match those criteria, but they’re so broad they could apply to anyone. What does it say about me? & Neutral & Well, you certainly meet those criteria, but they’re so broad that they could apply to anyone. What does that say about me? & Although you certainly meet those criteria, they are so broad that they could apply to anyone. What does this say about me? & You certainly meet those criteria, but they’re so broad that they could apply to anyone. What does that say about me? & You certainly meet those criteria, but they’re so broad that they could apply to anyone. What does that say about me? \\
        \hline
        Okay. How much do you want me to get? A gallon? & Neutral & Okay, how much would you like me to get? A gallon? & Could you please specify the quantity you would like me to purchase? A gallon would be suitable. & How much would you like me to get? A gallon? & How much do you want me to get? A gallon? \\
        \hline
        No. I think a half gallon will be enough. & Neutral & No, I think a half gallon will be sufficient. & No, I believe a half-gallon will suffice. & Nope, I think a half gallon will be just right. & I think a half gallon will suffice. \\
        \hline
        
    \end{tabular}
\end{table*}

\end{document}